# Random Worlds and Maximum Entropy

**Adam J. Grove**                                         GROVE@RESEARCH.NJ.NEC.COM
*NEC Research Institute, 4 Independence Way*
*Princeton, NJ 08540*

**Joseph Y. Halpern**                                       HALPERN@ALMADEN.IBM.COM
*IBM Almaden Research Center, 650 Harry Rd.*
*San Jose, CA 95120*

**Daphne Koller**                                            DAPHNE@CS.BERKELEY.EDU
*Computer Science Division, University of California*
*Berkeley, CA 94720*

## Abstract

Given a knowledge base *KB* containing first-order and statistical facts, we consider a principled method, called the *random-worlds method*, for computing a degree of belief that some formula $\varphi$ holds given *KB*. If we are reasoning about a world or system consisting of $N$ individuals, then we can consider all possible worlds, or first-order models, with domain $\{1, \ldots, N\}$ that satisfy *KB*, and compute the fraction of them in which $\varphi$ is true. We define the degree of belief to be the asymptotic value of this fraction as $N$ grows large. We show that when the vocabulary underlying $\varphi$ and *KB* uses constants and unary predicates only, we can naturally associate an *entropy* with each world. As $N$ grows larger, there are many more worlds with higher entropy. Therefore, we can use a *maximum-entropy* computation to compute the degree of belief. This result is in a similar spirit to previous work in physics and artificial intelligence, but is far more general. Of equal interest to the result itself are the limitations on its scope. Most importantly, the restriction to unary predicates seems necessary. Although the random-worlds method makes sense in general, the connection to maximum entropy seems to disappear in the non-unary case. These observations suggest unexpected limitations to the applicability of maximum-entropy methods.

## 1. Introduction

Consider an agent (or expert system) with some information about a particular subject, such as internal medicine. Some facts, such as "all patients with hepatitis exhibit jaundice", can be naturally expressed in a standard first-order logic, while others, such as "80% of patients that exhibit jaundice have hepatitis", are statistical. Suppose the agent wants to use this information to make decisions. For example, a doctor might need to decide whether to administer antibiotics to a particular patient Eric. To apply standard tools of decision theory (see (Luce & Raiffa, 1957) for an introduction), the agent must assign probabilities, or *degrees of belief*, to various events. For example, the doctor may need to assign a degree of belief to an event such as "Eric has hepatitis". We would therefore like techniques for computing degrees of belief in a principled manner, using all the data at hand. In this paper we investigate the properties of one particular formalism for doing this.

The method we consider, which we call the *random-worlds method*, has origins that go back to Bernoulli and Laplace (1820). It is essentially an application of what has been called the *principle of indifference* (Keynes, 1921). The basic idea is quite straightforward. Suppose we are interested in attaching a degree of belief to a formula $\varphi$ given a knowledge base $KB$. One useful way of assigning semantics to degrees of belief formulas is to use a probability distribution over a set of *possible worlds* (Halpern, 1990). More concretely, suppose for now that we are reasoning about $N$ individuals, $1, \ldots, N$. A *world* is a complete description of which individuals have each of the properties of interest. Formally, a world is just a model, or interpretation, over our first-order language. For example, if our language consists of the unary predicates *Hepatitis*, *Jaundice*, *Child*, and *BlueEyed*, the binary predicate *Infected-By*, and the constant *Eric*, then a world describes which subset of the $N$ individuals satisfies each of the unary predicates, which set of pairs is in the *Infected-By* relation, and which of the $N$ individuals is *Eric*. Given a prior probability distribution over the set of possible worlds, the agent can obtain a degree of belief in $\varphi$ given $KB$ by conditioning on $KB$ to obtain a posterior distribution, and then computing the probability of $\varphi$ according to this new distribution. The random-worlds method uses the principle of indifference to choose a particular prior distribution over the set of worlds: all the worlds are taken to be equally likely. It is easy to see that the degree of belief in $\varphi$ given $KB$ is then precisely the fraction of worlds satisfying $KB$ that also satisfy $\varphi$.

The approach so far described applies whenever we actually know the precise domain size $N$; unfortunately this is fairly uncommon. In many cases, however, it is reasonable to believe that $N$ is "large". We are thus particularly interested in the asymptotic behavior of this fraction; that is, we take our degree of belief to be the asymptotic value of this fraction as $N$ grows large.

For example, suppose we want to reason about a domain of hospital patients, and $KB$ is the conjunction of the following four formulas:

- $\forall x (Hepatitis(x) \Rightarrow Jaundice(x))$ ("all patients with hepatitis exhibit jaundice"),

- $\|Hepatitis(x)|Jaundice(x)\|_x \approx 0.8$ ("approximately 80% of patients that exhibit jaundice have hepatitis"; we explain this formalism and the reason we say "approximately 80%" rather than "exactly 80%" in Section 2),

- $\|BlueEyed(x)\|_x \approx 0.25$ ("approximately 25% of patients have blue eyes"),

- $Jaundice(Eric) \wedge Child(Eric)$ ("Eric is a child who exhibits jaundice").

Let $\varphi$ be $Hepatitis(Eric)$; that is, we want to ascribe a degree of belief to the statement "Eric has hepatitis". Suppose the domain has size $N$. Then we want to consider all worlds with domain $\{1, \ldots, N\}$ such that the set of individuals satisfying *Hepatitis* is a subset of those satisfying *Jaundice*, approximately 80% of the individuals satisfying *Jaundice* also satisfy *Hepatitis*, approximately 25% of the individuals satisfy *BlueEyed*, and (the interpretation of) *Eric* is an individual satisfying *Jaundice* and *Child*. It is straightforward to show that, as expected, *Hepatitis(Eric)* holds in approximately 80% of these structures. Moreover, as $N$ gets large, the fraction of structures in which *Hepatitis(Eric)* holds converges to exactly 0.8.

Since 80% of the patients that exhibit jaundice have hepatitis and Eric exhibits jaundice, a degree of belief of 0.8 that Eric has hepatitis seems justifiable. Note that, in this example, the information that Eric is a child is essentially treated as irrelevant. We would get the same answer if we did not have the information *Child(Eric)*. It can also be shown that the degree of belief in *BlueEyed*(*Eric*) converges to 0.25 as $N$ gets large. Furthermore, the degree of belief of *BlueEyed*(*Eric*) $\wedge$ *Jaundice*(*Eric*) converges to 0.2, the product of 0.8 and 0.25. As we shall see, this is because the random-worlds method treats *BlueEyed* and *Jaundice* as being independent, which is reasonable because there is no evidence to the contrary. (It would surely be strange to postulate that two properties were correlated unless there were reason to believe they were connected in some way.)

Thus, at least in this example, the random-worlds method gives answers that follow from the heuristic assumptions made in many standard AI systems (Pearl, 1989; Pollock, 1984; Spiegelhalter, 1986). Are such intuitive results typical? When do we get convergence? And when we do, is there a practical way to compute degrees of belief?

The answer to the first question is yes, as we discuss in detail in (Bacchus, Grove, Halpern, & Koller, 1994). In that paper, we show that the random-worlds method is remarkably successful at satisfying the desiderata of both nonmonotonic (default) reasoning (Ginsberg, 1987) and reference class reasoning (Kyburg, 1983). The results of (Bacchus et al., 1994) show that the behavior we saw in the example above holds quite generally, as do many other properties we would hope to have satisfied. Thus, in this paper we do not spend time justifying the random-worlds approach, nor do we discuss its strengths and weaknesses; the reader is referred to (Bacchus et al., 1994) for such discussion and for an examination of previous work in the spirit of random worlds (most notably (Carnap, 1950, 1952) and subsequent work). Rather, we focus on the latter two questions asked above. These questions may seem quite familiar to readers aware of the work on asymptotic probabilities for various logics. For example, in the context of first-order formulas, it is well-known that a formula with no constant or function symbols has an asymptotic probability of either 0 or 1 (Fagin, 1976; Glebskiǐ, Kogan, Liogon'kiǐ, & Talanov, 1969). Furthermore, we can decide which (Grandjean, 1983). However, the 0-1 law fails if the language includes constants or if we look at conditional probabilities (Fagin, 1976), and we need both these features in order to reason about degrees of belief for formulas involving particular individuals, conditioned on what is known.

In two companion papers (Grove, Halpern, & Koller, 1993a, 1993b), we consider the question of what happens in the pure first-order case (where there is no statistical information) in greater detail. We show that as long as there is at least one binary predicate symbol in the language, then not only do we not get asymptotic conditional probabilities in general (as was already shown by Fagin (1976)), but almost all the questions one might want to ask (such as deciding whether the limiting probability exists) are highly undecidable. However, if we restrict to a vocabulary with only unary predicate symbols and constants, then as long as the formula on which we are conditioning is satisfiable in arbitrarily large models (a question which is decidable in the unary case), the asymptotic conditional probability exists and can be computed effectively.

In this paper, we consider the much more useful case where the knowledge base has statistical as well as first-order information. In light of the results of (Grove et al., 1993a, 1993b), for most of the paper we restrict attention to the case when the knowledge base is expressed in a unary language. Our major result involves showing that asymptotic conditional probabilities can often be computed using *the principle of maximum entropy* (Jaynes, 1957; Shannon & Weaver, 1949).

To understand the use of maximum entropy, suppose the vocabulary consists of the unary predicate symbols $P_1, \ldots, P_k$. We can consider the $2^k$ *atoms* that can be formed from these predicate symbols, namely, the formulas of the form $P_1' \wedge \ldots \wedge P_k'$, where each $P_i'$ is either $P_i$ or $\neg P_i$. We can view the knowledge base as placing constraints on the proportion of domain elements satisfying each atom. For example, the constraint $\|P_1(x)|P_2(x)\|_x = 1/2$ says that the proportion of the domain satisfying some atom that contains $P_2$ as a conjunct is twice the proportion satisfying atoms that contain both $P_1$ and $P_2$ as conjuncts. Given a model of $KB$, we can define the entropy of this model as the entropy of the vector denoting the proportions of the different atoms. We show that, as $N$ grows large, there are many more models with high entropy than with lower entropy. Therefore, models with high entropy dominate. We use this *concentration phenomenon* to show that our degree of belief in $\varphi$ given $KB$ according to the random-worlds method is closely related to the assignment of proportions to atoms that has maximum entropy among all assignments consistent with the constraints imposed by $KB$.

The concentration phenomenon relating entropy to the random-worlds method is well-known (Jaynes, 1982, 1983). In physics, the "worlds" are the possible configurations of a system typically consisting of many particles or molecules, and the mutually exclusive properties (our atoms) can be, for example, quantum states. The corresponding entropy measure is at the heart of statistical mechanics and thermodynamics. There are subtle but important differences between our viewpoint and that of the physicists. The main one lies in our choice of language. We want to express some intelligent agent's knowledge, which is why we take first-order logic as our starting point. The most specific difference concerns constant symbols. We need these because the most interesting questions for us arise when we have some knowledge about—and wish to assign degrees of belief to statements concerning—a particular individual. The parallel in physics would address properties of a single particle, which is generally considered to be well outside the scope of statistical mechanics.

Another work that examines the connection between random worlds and entropy from our point of view—computing degrees of belief for formulas in a particular logic—is that of Paris and Vencovska (1989). They restrict the knowledge base to consist of a conjunction of constraints that (in our notation) have the form $\|\alpha(x)|\beta(x)\|_x \approx r$ and $\|\alpha(x)\|_x \approx r$, where $\beta$ and $\alpha$ are quantifier-free formulas involving unary predicates only, with no constant symbols. Not only is most of the expressive power of first-order logic not available in their approach, but the statistical information that can be expressed is quite limited. For example, it is not possible to make general assertions about statistical independence. Paris and Vencovska show that the degree of belief can be computed using maximum entropy for their language. Shastri (1989) has also shown such a result, of nearly equivalent scope. But, as we have already suggested, we believe that it is important to look at a far richer language. Our language allows arbitrary first-order assertions, full Boolean logic, arbitrary polynomial combinations of statistical expressions, and more; these are all features that are actually useful to knowledge-representation practitioners. Furthermore, the random-worlds method makes perfect sense in this rich setting. The goal of this paper is to discover whether the connection to maximum entropy also holds. We show that maximum entropy continues to be widely useful, covering many problems that are far outside the scope of (Paris & Vencovska, 1989; Shastri, 1989).

On the other hand, it turns out that we cannot make this connection for our entire language. For one thing, as we hinted earlier, there are problems if we try to condition on a knowledge base that includes non-unary predicates; we suspect that maximum entropy has no role whatsoever in this case. In addition, we show that there are subtleties that arise involving the interaction between statistical information and first-order quantification. We feel that an important contribution of this paper lies in pointing out some limitations of maximum-entropy methods.

The rest of this paper is organized as follows. In the next section, we discuss our formal framework (essentially, that of (Bacchus, 1990; Halpern, 1990)). We discuss the syntax and semantics of statistical assertions, issues involving "approximately equals", and define the random-worlds method formally. In Section 3 we state the basic results that connect maximum entropy to random-worlds, and in Section 4 we discuss how to use these results as effective computational procedures. In Section 5 we return to the issue of unary versus non-unary predicates, and the question of how widely applicable the principle of maximum entropy is. We conclude in Section 6 with some discussion.

## 2. Technical preliminaries

In this section, we give the formal definition of our language and the random-worlds method. The material is largely taken from (Bacchus et al., 1994).

### 2.1 The language

We are interested in a formal logical language that allows us to express both statistical information and first-order information. We therefore define a statistical language $\mathcal{L}^{\approx}$, which is a variant of a language designed by Bacchus (1990). For the remainder of the paper, let $\Phi$ be a finite first-order vocabulary, consisting of predicate and constant symbols, and let $\mathcal{X}$ be a set of variables.[1]

Our statistical language augments standard first-order logic with a form of statistical quantifier. For a formula $\psi(x)$, the term $||\psi(x)||_x$ is a *proportion expression*. It will be interpreted as a rational number between 0 and 1, that represents the proportion of domain elements satisfying $\psi(x)$. We actually allow an arbitrary set of variables in the subscript and in the formula $\psi$. Thus, for example, $||Child(x, y)||_x$ describes, for a fixed $y$, the proportion of domain elements that are children of $y$; $||Child(x, y)||_y$ describes, for a fixed $x$, the proportion of domain elements whose child is $x$; and $||Child(x, y)||_{x,y}$ describes the proportion of pairs of domain elements that are in the child relation.[2]

We also allow proportion expressions of the form $||\psi(x)|\theta(x)||_x$, which we call *conditional proportion expressions*. Such an expression is intended to denote the proportion of domain elements satisfying $\psi$ from among those elements satisfying $\theta$. Finally, any rational number is also considered to be a proportion expression, and the set of proportion expressions is closed under addition and multiplication.

---

1. For simplicity, we assume that $\Phi$ does not contain function symbols, since these can be defined in terms of predicates.

2. Strictly speaking, these proportion expression should be written with sets of variables in the subscript, as in $||Child(x, y)||_{\{x,y\}}$. However, when the interpretation is clear, we often abuse notation and drop the set delimiters.

One important difference between our syntax and that of (Bacchus, 1990) is the use of *approximate equality* to compare proportion expressions. There are both philosophical and practical reasons why exact comparisons can be inappropriate. Consider a statement such as "80% of patients with jaundice have hepatitis". If this statement appears in a knowledge base, it is almost certainly there as a summary of a large pool of data. So it would be wrong to interpret the value too literally, to mean that *exactly* 80% of all patients with jaundice have hepatitis. Furthermore, this interpretation would imply (among other things) that the number of jaundiced patients is a multiple of five! This is unlikely to be something we intend. We therefore use the approach described in (Bacchus et al., 1994; Koller & Halpern, 1992), and compare proportion expressions using (instead of $=$ and $\leq$) one of an infinite family of connectives $\approx_i$ and $\preceq_i$, for $i = 1, 2, 3 \ldots$ ("$i$-approximately equal" or "$i$-approximately less than or equal"). For example, we can express the statement "80% of jaundiced patients have hepatitis" by the *proportion formula* $\|Hep(x)|Jaun(x)\|_x \approx_1 0.8$. The intuition behind the semantics of approximate equality is that each comparison should be interpreted using some small tolerance factor to account for measurement error, sample variations, and so on. The appropriate tolerance will differ for various pieces of information, so our logic allows different subscripts on the "approximately equals" connectives. A formula such as $\|Fly(x)|Bird(x)\|_x \approx_1 1 \wedge \|Fly(x)|Bat(x)\|_x \approx_2 1$ says that both $\|Fly(x)|Bird(x)\|_x$ and $\|Fly(x)|Bat(x)\|_x$ are approximately 1, but the notion of "approximately" may be different in each case. The actual choice of subscript for $\approx$ is unimportant. However, it is important to use different subscripts for different approximate comparisons unless the tolerances for the different measurements are known to be the same.

We can now give a recursive definition of the language $\mathcal{L}^{\approx}$.

**Definition 2.1:** The set of *terms* in $\mathcal{L}^{\approx}$ is $\mathcal{X} \cup \mathcal{C}$ where $\mathcal{C}$ is the set of constant symbols in $\Phi$. The set of *proportion expressions* is the least set that

(a) contains the rational numbers,

(b) contains *proportion terms* of the form $\|\psi\|_X$ and $\|\psi|\theta\|_X$ for formulas $\psi, \theta \in \mathcal{L}^{\approx}$ and a finite set of variables $X \subseteq \mathcal{X}$, and

(c) is closed under addition and multiplication.

The set of formulas in $\mathcal{L}^{\approx}$ is the least set that

(a) contains *atomic formulas* of the form $R(t_1, \ldots, t_r)$, where $R$ is a predicate symbol in $\Phi \cup \{=\}$ of arity $r$ and $t_1, \ldots, t_r$ are terms,

(b) contains *proportion formulas* of the form $\zeta \approx_i \zeta'$ and $\zeta \preceq_i \zeta'$, where $\zeta$ and $\zeta'$ are proportion expressions and $i$ is a natural number, and

(c) is closed under conjunction, negation, and first-order quantification. ∎

Note that $\mathcal{L}^{\approx}$ allows the use of equality when comparing terms, but not when comparing proportion expressions.

This definition allows arbitrary nesting of quantifiers and proportion expressions. As observed in (Bacchus, 1990), the subscript $x$ in a proportion expressions binds the variable $x$ in the expression; indeed, we can view $\|\cdot\|_x$ as a new type of quantification.

We now need to define the semantics of the logic. As we shall see below, most of the definitions are fairly straightforward. The two features that cause problems are approximate comparisons and conditional proportion expressions. We interpret the approximate connective $\zeta \approx_i \zeta'$ to mean that $\zeta$ is very close to $\zeta'$. More precisely, it is within some very small tolerance factor. We formalize this using a *tolerance vector* $\vec{\tau} = \langle \tau_1, \tau_2, \ldots \rangle$, $\tau_i > 0$. Intuitively $\zeta \approx_i \zeta'$ if the values of $\zeta$ and $\zeta'$ are within $\tau_i$ of each other. Of course, one problem with this is that we generally will not know the value of $\tau_i$. We postpone discussion of this issue until the next section.

Another difficulty arises when interpreting conditional proportion expressions. The problem is that $\|\psi|\theta\|_X$ cannot be defined as a conditional probability when there are no assignments to the variables in $X$ that would satisfy $\theta$, because we cannot divide by zero. When standard equality is used rather than approximate equality this problem is easily overcome, simply by avoiding conditional probabilities in the semantics altogether. Following (Halpern, 1990), we can eliminate conditional proportion expressions altogether by viewing a statement such as $\|\psi|\theta\|_X = \alpha$ as an abbreviation for $\|\psi \wedge \theta\|_X = \alpha\|\theta\|_X$. Thus, we never actually form quotients of probabilities. This approach agrees completely with the standard interpretation of conditionals so long as $\|\theta\|_X \neq 0$. If $\|\theta\|_X = 0$, it enforces the convention that formulas such as $\|\psi|\theta\|_X = \alpha$ or $\|\psi|\theta\|_X \leq \alpha$ are true for any $\alpha$. (Note that we do not really care much what happens in such cases, so long as it is consistent and well-defined. This convention represents one reasonable choice.)

We used the same approach in an earlier version of this paper (Grove, Halpern, & Koller, 1992) in the context of a language that uses approximate equality. Unfortunately, as the following example shows, this has problems. Unlike the case for true equality, if we multiply by $\|\theta\|_X$ to clear all quotients, we do not obtain an equivalent formula even if $\|\theta\|_X$ is nonzero.

**Example 2.2:** First consider the knowledge base $KB = (\|Fly(x)|Penguin(x)\|_x \approx_1 0)$. This says that the number of flying penguins forms a tiny proportion of all penguins. However, if we interpret conditional proportions as above and multiply out, we obtain the knowledge base $KB' = \|Fly(x) \wedge Penguin(x)\|_x \approx_1 0 \cdot \|Penguin(x)\|_x$, which is equivalent to $\|Fly(x) \wedge Penguin(x)\|_x \approx_1 0$. $KB'$ just says that the number of flying penguins is small, and has lost the (possibly important) information that the number of flying penguins is small *relative to the number of penguins*. It is quite consistent with $KB'$ that all penguins fly (provided the total number of penguins is small); this is not consistent with $KB$. Clearly, the process of multiplying out across an approximate connective does not preserve the intended interpretation of the formulas. ∎

This example demonstrates an undesirable interaction between the semantics we have chosen for approximate equality and the process of multiplying-out to eliminate conditional proportions. We expect $\|\psi|\theta\|_X \approx_1 \alpha$ to mean that $\|\psi|\theta\|_X$ is within some tolerance $\tau_1$ of $\alpha$. Assuming $\|\theta\|_X > 0$, this is the same as saying that $\|\psi \wedge \theta\|_X$ is within $\tau_1 \|\theta\|_X$ of $\alpha \|\theta\|_X$. On the other hand, the expression that results by multiplying out is $\|\psi \wedge \theta\|_X \approx_1 \alpha\|\theta\|_X$. This says that $\|\psi \wedge \theta\|_X$ is within $\tau_1$ (not $\tau_1 \|\theta\|_X$!) of $\alpha\|\theta\|_X$. As we saw above, the difference between the two interpretations can be significant.

Because of this problem, we cannot treat conditional proportions as abbreviations and instead have added them as primitive expressions in the language. Of course, we now have to give them a semantics that avoids the problem illustrated by Example 2.2. We would like to maintain the conventions used when we had equality in the language. Namely, in worlds where $\|\theta(x)\|_x \neq 0$, we want $\|\psi(x)|\theta(x)\|_x$ to denote the fraction of elements satisfying $\theta(x)$ that also satisfy $\psi(x)$. In worlds where $\|\theta(x)\|_x = 0$, we want formulas of the form $\|\psi(x)|\theta(x)\|_x \approx_i \alpha$ or $\|\psi(x)|\theta(x)\|_x \preceq_i \alpha$ to be true. There are a number of ways of accomplishing this. The way we take is perhaps not the simplest, but it introduces machinery that will be helpful later. The basic idea is to make the interpretation of $\approx$ more explicit, so that we can eliminate conditional proportions by multiplication and keep track of all the consequences of doing so.

We give semantics to the language $\mathcal{L}^{\approx}$ by providing a translation from formulas in $\mathcal{L}^{\approx}$ to formulas in a language $\mathcal{L}^{=}$ whose semantics is more easily described. The language $\mathcal{L}^{=}$ is essentially the language of (Halpern, 1990), that uses true equality rather than approximate equality when comparing proportion expressions. More precisely, the definition of $\mathcal{L}^{=}$ is identical to the definition of $\mathcal{L}^{\approx}$ given in Definition 2.1, except that:

- we use $=$ and $\leq$ instead of $\approx_i$ and $\preceq_i$,

- we allow the set of proportion expressions to include arbitrary real numbers (not just rational numbers),

- we do not allow conditional proportion expressions,

- we assume that $\mathcal{L}^{=}$ has a special family of variables $\varepsilon_i$, for $i = 1, 2, \ldots$, interpreted over the reals.

The variable $\varepsilon_i$ is used to explicitly interpret the approximate equality connectives $\approx_i$ and $\preceq_i$. Once this is done, we can safely multiply out the conditionals, as described above. More precisely, every formula $\chi \in \mathcal{L}^{\approx}$ can be associated with a formula $\chi^{\star} \in \mathcal{L}^{=}$ as follows:

- every proportion formula $\zeta \preceq_i \zeta'$ in $\chi$ is (recursively) replaced by $\zeta - \zeta' \leq \varepsilon_i$,

- every proportion formula $\zeta \approx_i \zeta'$ in $\chi$ is (recursively) replaced by the conjunction $(\zeta - \zeta' \leq \varepsilon_i) \wedge (\zeta' - \zeta \leq \varepsilon_i)$,

- finally, conditional proportion expressions are eliminated by multiplying out.

This translation allows us to embed $\mathcal{L}^{\approx}$ into $\mathcal{L}^{=}$. Thus, for the remainder of the paper, we regard $\mathcal{L}^{\approx}$ as a sublanguage of $\mathcal{L}^{=}$. This embedding avoids the problem encountered in Example 2.2, because when we multiply to clear conditional proportions the tolerances are explicit, and so are also multipled as appropriate.

The semantics for $\mathcal{L}^{=}$ is quite straightforward, and is similar to that in (Halpern, 1990). We give semantics to $\mathcal{L}^{=}$ in terms of *worlds*, or finite first-order models. For any natural number $N$, let $\mathcal{W}_N$ consist of all worlds with domain $\{1, \ldots, N\}$. Thus, in $\mathcal{W}_N$, we have one world for each possible interpretation of the symbols in $\Phi$ over the domain $\{1, \ldots, N\}$. Let $\mathcal{W}^{\star}$ denote $\cup_N \mathcal{W}_N$.

Now, consider some world $W \in \mathcal{W}^{\star}$ over the domain $D = \{1, \ldots, N\}$, some valuation $V : \mathcal{X} \to D$ for the variables in $\mathcal{X}$, and some tolerance vector $\vec{\tau}$. We simultaneously assign to each proportion expression $\zeta$ a real number $[\zeta]_{(W,V,\vec{\tau})}$ and to each formula $\xi$ a truth value with respect to $(W, V, \vec{\tau})$. Most of the clauses of the definition are completely standard, so we omit them here. In particular, variables are interpreted using $V$, the tolerance variables $\varepsilon_i$ are interpreted using the tolerances $\tau_i$, the predicates and constants are interpreted using $W$, the Boolean connectives and the first-order quantifiers are defined in the standard fashion, and when interpreting proportion expressions, the real numbers, addition, multiplication, and $\leq$ are given their standard meaning. It remains to interpret proportion terms. Recall that we eliminate conditional proportion terms by multiplying out, so that we need to deal only with unconditional proportion terms. If $\zeta$ is the proportion expression $||\psi||_{x_{i_1}, \ldots, x_{i_k}}$ (for $i_1 < i_2 < \ldots < i_k$), then

$$[\zeta]_{(W, V, \vec{\tau})} = \frac{1}{|D^k|} \left| \left\{ (d_1, \ldots, d_k) \in D^k \; : \; (W, V[x_{i_1}/d_1, \ldots, x_{i_k}/d_k], \vec{\tau}) \models \psi \right\} \right|.$$

Thus, if $|D| = N$, the proportion expression $||\psi||_{x_{i_1}, \ldots, x_{i_k}}$ denotes the fraction of the $N^k$ $k$-tuples in $D^k$ that satisfy $\psi$. For example, $[|| Child(x, y)||_x]_{(W, V, \vec{\tau})}$ is the fraction of domain elements $d$ that are children of $V(y)$.

Using our embedding of $\mathcal{L}^\approx$ into $\mathcal{L}^=$, we now have semantics for $\mathcal{L}^\approx$. For $\chi \in \mathcal{L}^\approx$, we say that $(W, V, \vec{\tau}) \models \chi$ iff $(W, V, \vec{\tau}) \models \chi^\bullet$. It is sometimes useful in our future results to incorporate particular values for the tolerances into the formula $\chi^\bullet$. Thus, let $\chi[\vec{\tau}]$ represent the formula that results from $\chi^\bullet$ if each variable $\varepsilon_i$ is replaced with its value according to $\vec{\tau}$, that is, $\tau_i$.[3]

Typically we are interested in closed sentences, that is, formulas with no free variables. In that case, it is not hard to show that the valuation plays no role. Thus, if $\chi$ is closed, we write $(W, \vec{\tau}) \models \chi$ rather than $(W, V, \vec{\tau}) \models \chi$. Finally, if $KB$ and $\chi$ are closed formulas, we write $KB \models \chi$ if $(W, \vec{\tau}) \models KB$ implies $(W, \vec{\tau}) \models \chi$.

## 2.2 Degrees of belief

As we explained in the introduction, we give semantics to degrees of belief by considering all worlds of size $N$ to be equally likely, conditioning on $KB$, and then checking the probability of $\varphi$ over the resulting probability distribution. In the previous section, we defined what it means for a sentence $\chi$ to be satisfied in a world of size $N$ using a tolerance vector $\vec{\tau}$. Given $N$ and $\vec{\tau}$, we define $\#worlds_N^{\vec{\tau}}(\chi)$ to be the number of worlds in $\mathcal{W}_N$ such that $(W, \vec{\tau}) \models \chi$. Since we are taking all worlds to be equally likely, the degree of belief in $\varphi$ given $KB$ with respect to $\mathcal{W}_N$ and $\vec{\tau}$ is

$$\Pr_N^{\vec{\tau}}(\varphi|KB) = \frac{\#worlds_N^{\vec{\tau}}(\varphi \wedge KB)}{\#worlds_N^{\vec{\tau}}(KB)}.$$

If $\#worlds_N^{\vec{\tau}}(KB) = 0$, this degree of belief is not well-defined.

The careful reader may have noticed a potential problem with this definition. Strictly speaking, we should write $\mathcal{W}_N(\Phi)$ rather than $\mathcal{W}_N$, since the set of worlds under consideration clearly depends on the vocabulary. Hence, the number of worlds in $\mathcal{W}_N$ also depends on the vocabulary. Thus, both $\#worlds_N^{\vec{\tau}}(\varphi)$ and $\#worlds_N^{\vec{\tau}}(\varphi \wedge KB)$ depend on the choice of $\Phi$. Fortunately, this dependence "cancels out": If $\Phi' \supset \Phi$, then there is a constant $c$ such that for all formulas $\chi$ over the vocabulary $\Phi$, $\#[\Phi'] worlds_N^{\vec{\tau}}(\chi) = c \#[\Phi] worlds_N^{\vec{\tau}}(\chi)$. This result, from which it follows that the degree of belief $\Pr_N^{\vec{\tau}}(\varphi|KB)$ is independent of our choice of vocabulary, is proved in (Grove et al., 1993b).

---

3. Note that some of the tolerances $\tau_i$ may be irrational; it is for this reason that we allowed irrational numbers in proportion expressions in $\mathcal{L}^=$.

Typically, we know neither $N$ nor $\vec{\tau}$ exactly. All we know is that $N$ is "large" and that $\vec{\tau}$ is "small". Thus, we would like to take our *degree of belief* in $\varphi$ given $KB$ to be $\lim_{\vec{\tau} \to \vec{0}} \lim_{N \to \infty} \Pr_N^{\vec{\tau}}(\varphi|KB)$. Notice that the order of the two limits over $\vec{\tau}$ and $N$ is important. If the limit $\lim_{\vec{\tau} \to \vec{0}}$ appeared last, then we would gain nothing by using approximate equality, since the result would be equivalent to treating approximate equality as exact equality.

This definition, however, is not sufficient; the limit may not exist. We observed above that $\Pr_N^{\vec{\tau}}(\varphi|KB)$ is not always well-defined. In particular, it may be the case that for certain values of $\vec{\tau}$, $\Pr_N^{\vec{\tau}}(\varphi|KB)$ is not well-defined for arbitrarily large $N$. In order to deal with this problem of well-definedness, we define $KB$ to be *eventually consistent* if for all sufficiently small $\vec{\tau}$ and sufficiently large $N$, $\#worlds_N^{\vec{\tau}}(KB) > 0$. Among other things, eventual consistency implies that the $KB$ is satisfiable in finite domains of arbitrarily large size. For example, a $KB$ stating that "there are exactly 7 domain elements" is not eventually consistent. For the remainder of the paper, we assume that all knowledge bases are eventually consistent. In practice, we expect eventual consistency to be no harder to check than consistency. We do not expect a knowledge base to place bounds on the domain size, except when the bound is readily apparent. For those unsatisfied with this intuition, it is also possible to find formal conditions ensuring eventual consistency. For instance, it is possible to show that the following conditions are sufficient to guarantee that $KB$ is eventually consistent: (a) $KB$ does not use any non-unary predicates, including equality between terms and (b) $KB$ is consistent for *some* domain size when all approximate comparisons are replaced by exact comparisons. Since we concentrate on unary languages in this paper, this result covers most cases of interest.

Even if $KB$ is eventually consistent, the limit may not exist. For example, it may be the case that $\Pr_N^{\vec{\tau}}(\varphi|KB)$ oscillates between $\alpha + \tau_i$ and $\alpha - \tau_i$ for some $i$ as $N$ gets large. In this case, for any particular $\vec{\tau}$, the limit as $N$ grows will not exist. However, it seems as if the limit as $\vec{\tau}$ grows small should, in this case, be $\alpha$, since the oscillations about $\alpha$ go to 0. We avoid such problems by considering the *lim sup* and *lim inf*, rather than the limit. For any set $S \subset \mathbb{R}$, the infimum of $S$, $\inf S$, is the greatest lower bound of $S$. The *lim inf* of a sequence is the limit of the infimums; that is,

$$\liminf_{N \to \infty} a_N = \lim_{N \to \infty} \inf\{a_i : i > N\}.$$

The lim inf exists for any sequence bounded from below, even if the limit does not. The *lim sup* is defined analogously, where $\sup S$ denotes the least upper bound of $S$. If $\lim_{N \to \infty} a_N$ does exist, then $\lim_{N \to \infty} a_N = \liminf_{N \to \infty} a_N = \limsup_{N \to \infty} a_N$. Since, for any $\vec{\tau}$, the sequence $\Pr_N^{\vec{\tau}}(\varphi|KB)$ is always bounded from above and below, the lim sup and lim inf always exist. Thus, we do not have to worry about the problem of nonexistence for particular values of $\vec{\tau}$. We can now present the final form of our definition.

**Definition 2.3:** If

$$\lim_{\vec{\tau} \to \vec{0}} \liminf_{N \to \infty} \Pr_N^{\vec{\tau}}(\varphi|KB) \quad \text{and} \quad \lim_{\vec{\tau} \to \vec{0}} \limsup_{N \to \infty} \Pr_N^{\vec{\tau}}(\varphi|KB)$$

both exist and are equal, then the *degree of belief in $\varphi$ given KB*, written $\Pr_\infty(\varphi|KB)$, is defined as the common limit; otherwise $\Pr_\infty(\varphi|KB)$ does not exist.

We close this section with a few remarks on our definition. First note that, even using this definition, there are many cases where the degree of belief does not exist. However, as some of our later examples show, in many situations the nonexistence of a degree of belief can be understood intuitively (for instance, see Example 4.3 and the subsequent discussion). We could, alternatively, have taken the degree of belief to be the interval defined by $\lim_{\vec{\tau}\to\vec{0}} \liminf_{N\to\infty} \Pr_N^{\vec{\tau}}(\varphi|KB)$ and $\lim_{\vec{\tau}\to\vec{0}} \limsup_{N\to\infty} \Pr_N^{\vec{\tau}}(\varphi|KB)$, provided each of them exist. This would have been a perfectly reasonable choice; most of the results we state would go through with very little change if we had taken this definition. Our definition simplifies the exposition slightly.

Finally, we remark that it may seem unreasonable to take limits if we know the domain size or have a bound on the domain size. Clearly, if we know $N$ and $\vec{\tau}$, then it seems more reasonable to use $\Pr_N^{\vec{\tau}}$ rather than $\Pr_\infty$ as our degree of belief. Indeed, as shown in (Bacchus et al., 1994), many of the important properties that hold for the degree of belief defined by $\Pr_\infty$ hold for $\Pr_N^{\vec{\tau}}$, for all choices of $N$ and $\vec{\tau}$. The connection to maximum entropy that we make in this paper holds only at the limit, but because (as our proofs show) the convergence is rapid, the degree of belief $\Pr_\infty(\varphi|KB)$ is typically a very good approximation to $\Pr_N^{\vec{\tau}}(\varphi|KB)$, even for moderately large $N$ and moderately small $\vec{\tau}$.

## 3. Degrees of belief and entropy

### 3.1 Introduction to maximum entropy

The idea of maximizing *entropy* has played an important role in many fields, including the study of probabilistic models for inferring degrees of belief (Jaynes, 1957; Shannon & Weaver, 1949). In the simplest setting, we can view entropy as a real-valued function on finite probability spaces. If $\Omega$ is a finite set and $\mu$ is a probability measure on $\Omega$, the entropy $H(\mu)$ is defined to be $-\sum_{\omega\in\Omega} \mu(\omega) \ln \mu(\omega)$ (we take $0 \ln 0 = 0$).

One standard application of entropy is the following. Suppose we know the space $\Omega$, but have only partial information about $\mu$, expressed in the form of constraints. For example, we might have a constraint such as $\mu(\omega_1) + \mu(\omega_2) \geq 1/3$. Although there may be many measures $\mu$ that are consistent with what we know, the *principle of maximum entropy* suggests that we adopt that $\mu^*$ which has the largest entropy among all the consistent possibilities. Using the appropriate definitions, it can be shown that there is a sense in which this $\mu^*$ incorporates the "least" additional information (Shannon & Weaver, 1949). For example, if we have no constraints on $\mu$, then $\mu^*$ will be the measure that assigns equal probability to all elements of $\Omega$. Roughly speaking, $\mu^*$ assigns probabilities as equally as possible given the constraints.

### 3.2 From formulas to constraints

Like maximum entropy, the random-worlds method is also used to determine degrees of belief (i.e., probabilities) relative to a knowledge base. Aside from this, is there any connection between the two ideas? Of course, there is the rather trivial observation that random-worlds considers a uniform probability distribution (over the set of worlds satisfying $KB$), and it is well-known that the uniform distribution over any set has the highest possible entropy. But in this section we show another, entirely different and much deeper, connection between random-worlds and the principle of maximum entropy. This connection holds *provided that we restrict the knowledge base so that it uses only unary predicates and constants.* In this case we can consider probability distributions, and in particular the maximum-entropy distribution, over the set of atoms. Atoms are of course very different from possible worlds; for instance, there are only finitely many of them (independent of the domain size $N$). Furthermore, the maximum-entropy distributions we consider will typically not be uniform. Nevertheless, maximum entropy in this new space can tell us a lot about the degrees of belief defined by random worlds. In particular, this connection will allow us to use maximum entropy as a tool for computing degrees of belief. We believe that the restriction to unary predicates is necessary for the connection we are about to make. Indeed, as long as the knowledge base makes use of a binary predicate symbol (or unary function symbol), we suspect that there is no useful connection between the two approaches at all; see Section 5 for some discussion.

Let $\mathcal{L}_1^{\approx}$ be the sublanguage of $\mathcal{L}^{\approx}$ where only unary predicate symbols and constant symbols appear in formulas; in particular, we assume that equality between terms does not occur in formulas in $\mathcal{L}_1^{\approx}$.[4] (Recall that in $\mathcal{L}^{\approx}$, we allow equality between terms, but disallow equality between proportion expressions.) Let $\mathcal{L}_1^{=}$ be the corresponding sublanguage of $\mathcal{L}^{=}$. In this subsection, we show that the expressive power of a knowledge base $KB$ in the language $\mathcal{L}_1^{\approx}$ is quite limited. In fact, such a $KB$ can essentially only place constraints on the proportions of the atoms. If we then think of these as constraints on the "probabilities of the atoms", then we have the ingredients necessary to apply maximum entropy. In Section 3.3 we show that there is a strong connection between the maximum-entropy distribution found this way and the degree of belief generated by random-worlds method.

To see what constraints a formula places on the probabilities of atoms, it is useful to convert the formula to a certain canonical form. As a first step to doing this, we formalize the definition of atom given in the introduction. Let $\mathcal{P} = \{P_1, \ldots, P_k\}$ consist of the unary predicate symbols in the vocabulary $\Phi$.

**Definition 3.1:** An *atom (over $\mathcal{P}$)* is conjunction of the form $P_1'(x) \wedge \ldots \wedge P_k'(x)$, where each $P_i'$ is either $P_i$ or $\neg P_i$. Since the variable $x$ is irrelevant to our concerns, we typically suppress it and describe an atom as a conjunction of the form $P_1' \wedge \ldots \wedge P_k'$. ∎

Note that there are $2^{|\mathcal{P}|} = 2^k$ atoms over $\mathcal{P}$ and that they are mutually exclusive and exhaustive. Throughout this paper, we use $K$ to denote $2^k$ and $A_1, \ldots, A_K$ to denote the atoms over $\mathcal{P}$, listed in some fixed order.

**Example 3.2:** There are $K = 4$ atoms over $\mathcal{P} = \{P_1, P_2\}$: $A_1 = P_1 \wedge P_2$, $A_2 = P_1 \wedge \neg P_2$, $A_3 = \neg P_1 \wedge P_2$, $A_4 = \neg P_1 \wedge \neg P_2$. ∎

The *atomic proportion terms* $||A_1(x)||_x, \ldots, ||A_K(x)||_x$ will play a significant role in our technical development. It turns out that $\mathcal{L}_1^{\approx}$ is a rather weak language: a formula $KB \in \mathcal{L}_1^{\approx}$ does little more than constrain the proportion of the atoms. In other words, for any such $KB$ we can find an equivalent formula in which the only proportion expressions are these unconditional proportions of atoms. The more complex syntactic machinery in $\mathcal{L}_1^{\approx}$—proportions over tuples, first-order quantification, nested proportions, and conditional proportions—does not add expressive power. (It does add convenience, however; knowledge can often be expressed far more succinctly if the full power of the language is used.)

---

4. We remark that many of our results can be extended to the case where the $KB$ mentions equality, but the extra complexity obscures many of the essential ideas.

Given any $KB$, the first step towards applying maximum entropy is to use $\mathcal{L}_1^{\approx}$'s lack of expressivity and replace all proportion terms by atomic proportion terms. It is also useful to make various other simplifications to $KB$ that will help us in Section 4. We combine these steps and require that $KB$ be transformed into a special *canonical form* which we now describe.

**Definition 3.3:** An *atomic term* $t$ *over* $\mathcal{P}$ is a polynomial over terms of the form $\|A(x)\|_x$, where $A$ is an atom over $\mathcal{P}$. Such an atomic term $t$ is *positive* if every coefficient of the polynomial $t$ is positive. ∎

**Definition 3.4:** A (closed) sentence $\chi \in \mathcal{L}_1^{=}$ is in *canonical form* if it is a disjunction of conjunctions, where each conjunct is one of the following:

- $t' = 0$, $(t' > 0 \wedge t \leq t'\varepsilon_i)$, or $(t' > 0 \wedge \neg(t \leq t'\varepsilon_i))$, where $t$ and $t'$ are atomic terms and $t'$ is positive,

- $\exists x \, A_i(x)$ or $\neg\exists x \, A_i(x)$ some atom $A_i$, or

- $A_i(c)$ for some atom $A_i$ and some constant $c$.

Furthermore, a disjunct cannot contain both $A_i(c)$ and $A_j(c)$ for $i \neq j$ as conjuncts, nor can it contain both $A_i(c)$ and $\neg\exists x \, A_i(x)$. (Note that these last conditions are simply minimal consistency requirements.) ∎

**Theorem 3.5:** *Every formula in $\mathcal{L}_1^{=}$ is equivalent to a formula in canonical form. Moreover, there is an effective procedure that, given a formula $\xi \in \mathcal{L}_1^{=}$, constructs an equivalent formula $\widehat{\xi}$ in canonical form.*

The proof of this theorem, and of all theorems in this paper, can be found in the appendix.

We remark that the length of the formula $\widehat{\xi}$ is typically exponential in the length of $\xi$. Such a blowup seems inherent in any scheme defined in terms of atoms.

Theorem 3.5 is a generalization of Claim 5.7.1 in (Halpern, 1990). It, in turn, is a generalization of a well-known result which says that any first-order formula with only unary predicates is equivalent to one with only depth-one quantifier nesting. Roughly speaking, this is because for a quantified formula such as $\exists x \, \xi'$, subformulas talking about a variable $y$ other than $x$ can be moved outside the scope of the quantifier. This is possible because no literal subformula can talk about $x$ and $y$ together. Our proof uses the same idea and extends it to proportion statements. In particular, it shows that for any $\xi \in \mathcal{L}_1^{\approx}$ there is an equivalent $\hat{\xi}$ which has no nested quantifiers or nested proportions.

Notice, however, that such a result does not hold once we allow even a single binary predicate in the language. For example, the formula $\forall y \, \exists x \, R(x, y)$ clearly needs nested quantification because $R(x, y)$ talks about both $x$ and $y$ and so must remain within the scope of both quantifiers. With binary predicates, each additional depth of nesting really does add expressive power. This shows that there can be no "canonical form" theorem quite like Theorem 3.5 for richer languages. This issue is one of the main reasons why we restrict the $KB$ to a unary language in this paper. (See Section 5 for further discussion.)

Given any formula in canonical form we can immediately derive from it, in a syntactic manner, a set of constraints on the possible proportions of atoms.

**Definition 3.6:** Let $KB$ be in canonical form. We construct a formula $\Gamma(KB)$ in the language of real closed fields (i.e., over the vocabulary $\{0, 1, +, \times\}$) as follows, where $u_1, \ldots, u_K$ are fresh variables (distinct from the tolerance variables $\varepsilon_j$):

- we replace each occurrence of the formula $A_i(c)$ by $u_i > 0$,

- we replace each occurrence of $\exists x \, A_i(x)$ by $u_i > 0$ and replace each occurrence of $\neg \exists x \, A_i(x)$ by $u_i = 0$,

- we replace each occurrence of $\|A_i(x)\|_x$ by $u_i$. ∎

Notice that $\Gamma(KB)$ has two types of variables: the new variables $u_i$ that we just introduced, and the tolerance variables $\varepsilon_i$. In order to eliminate the dependence on the latter, we often consider the formula $\Gamma(KB[\vec{\tau}])$ for some tolerance vector $\vec{\tau}$.

**Definition 3.7:** Given a formula $\gamma$ over the variables $u_1, \ldots, u_K$, let $Sol[\gamma]$ be the set of vectors in $\Delta^K = \{\vec{u} \in [0, 1]^K : \sum_i^K u_i = 1\}$ satisfying $\gamma$. Formally, if $(a_1, \ldots, a_K) \in \Delta^K$, then $(a_1, \ldots, a_K) \in Sol[\gamma]$ iff $(\mathbb{R}, V) \models \gamma$, where $V$ is a valuation such that $V(u_i) = a_i$. ∎

**Definition 3.8:** The *solution space of KB given* $\vec{\tau}$, denoted $S^{\vec{\tau}}[KB]$, is defined to be the closure of $Sol[\Gamma(KB[\vec{\tau}])]$.[5] ∎

If $KB$ is not in canonical form, we define $\Gamma(KB)$ and $S^{\vec{\tau}}[KB]$ to be $\Gamma(\widehat{KB})$ and $S^{\vec{\tau}}[\widehat{KB}]$, respectively, where $\widehat{KB}$ is the formula in canonical form equivalent to $KB$ obtained by the procedure appearing in the proof of Theorem 3.5.

**Example 3.9:** Let $\mathcal{P}$ be $\{P_1, P_2\}$, with the atoms ordered as in Example 3.2. Consider

$$KB = \forall x \, P_1(x) \wedge 3 \|P_1(x) \wedge P_2(x)\|_x \preceq_i 1.$$

The canonical formula $\widehat{KB}$ equivalent to $KB$ is:[6]

$$\neg \exists x \, A_3(x) \wedge \neg \exists x \, A_4(x) \wedge 3 \|A_1(x)\|_x - 1 \leq \varepsilon_i.$$

As expected, $\widehat{KB}$ constrains both $\|A_3(x)\|_x$ and $\|A_4(x)\|_x$ (i.e., $u_3$ and $u_4$) to be 0. We also see that $\|A_1(x)\|_x$ (i.e., $u_1$) is (approximately) at most $1/3$. Therefore:

$$S^{\vec{\tau}}[KB] = \left\{ (u_1, \ldots, u_4) \in \Delta^4 : u_1 \leq 1/3 + \tau_i/3, u_3 = u_4 = 0 \right\}. \quad ∎$$

---

5. Recall that the *closure* of a set $X \subseteq \mathbb{R}^K$ consists of all $K$-tuples that are the limit of a sequence of $K$-tuples in $X$.

6. Note that here we are viewing $KB$ as a formula in $\mathcal{L}^=$, under the translation defined earlier; we do this throughout the paper without further comment.

### 3.3 The concentration phenomenon

With every world $W \in \mathcal{W}^*$, we can associate a particular tuple $(u_1, \ldots, u_K)$, where $u_i$ is the fraction of the domain satisfying atom $A_i$ in $W$:

**Definition 3.10:** Given a world $W \in \mathcal{W}^*$, we define $\pi(W) \in \Delta^K$ to be

$$(||A_1(x)||_x, ||A_2(x)||_x, \ldots, ||A_K(x)||_x)$$

where the values of the proportions are interpreted over $W$. We say that the vector $\pi(W)$ is the *point* associated with $W$. ∎

We define the *entropy* of any model $W$ to be the entropy of $\pi(W)$; that is, if $\pi(W) = (u_1, \ldots, u_K)$, then the entropy of $W$ is $H(u_1, \ldots, u_K)$. As we are about to show, the entropy of $\vec{u}$ turns out to be a very good asymptotic indicator of how many worlds $W$ there are such that $\pi(W) = \vec{u}$. In fact, there are so many more worlds near points of high entropy that we can ignore all the other points when computing degrees of belief. This *concentration* phenomenon, as Jaynes (1982) has called it, is essentially the content of the next lemma and justifies our interest in the maximum-entropy point(s) of $S^{\vec{r}}[KB]$.

For any $\mathcal{S} \subseteq \Delta^K$ let $\#worlds_N^{\vec{r}}[\mathcal{S}](KB)$ denote the number of worlds $W$ of size $N$ such that $(W, \vec{r}) \models KB$ and such that $\pi(W) \in \mathcal{S}$; for any $\vec{u} \in \Delta^K$ let $\#worlds_N^{\vec{r}}[\vec{u}](KB)$ abbreviate $\#worlds_N^{\vec{r}}[\{\vec{u}\}](KB)$. Of course $\#worlds_N^{\vec{r}}[\vec{u}](KB)$ is necessarily zero unless all components of $\vec{u}$ are multiples of $1/N$. However, if there are any models associated with $\vec{u}$ at all, we can estimate their number quite accurately using the entropy function:

**Lemma 3.11:** *There exist some function $h : \mathbb{N} \to \mathbb{N}$ and two strictly positive polynomial functions $f, g : \mathbb{N} \to \mathbb{R}$ such that, for $KB \in \mathcal{L}_1^{\approx}$ and $\vec{u} \in \Delta^K$, if $\#worlds_N^{\vec{r}}[\vec{u}](KB) \neq 0$, then*

$$(h(N)/f(N))e^{NH(\vec{u})} \leq \#worlds_N^{\vec{r}}[\vec{u}](KB) \leq h(N)g(N)e^{NH(\vec{u})}.$$

Of course, it follows from the lemma that tuples whose entropy is near maximum have overwhelmingly more worlds associated with them than tuples whose entropy is further from maximum. This is essentially the concentration phenomenon.

Lemma 3.11 is actually fairly easy to prove. The following simple example illustrates the main idea.

**Example 3.12:** Suppose $\Phi = \{P\}$ and $KB = true$. We have

$$\Delta^K = \Delta^2 = \{(u_1, 1 - u_1) : \quad 0 \leq u_1 \leq 1\},$$

where the atoms are $A_1 = P$ and $A_2 = \neg P$. For any $N$, partition the worlds in $\mathcal{W}_N$ according to the point to which they correspond. For example, the graph in Figure 1 shows us the partition of $\mathcal{W}_4$. In general, consider some point $\vec{u} = (r/N, (N-r)/N)$. The number of worlds corresponding to $\vec{u}$ is simply the number of ways of choosing the denotation of $P$. We need to choose which $r$ elements satisfy $P$; hence, the number of such worlds is $\binom{N}{r} = \frac{N!}{r!(N-r)!}$. Figure 2 shows the qualitative behavior of this function for large values of $N$. It is easy to see the asymptotic concentration around $\vec{u} = (0.5, 0.5)$.

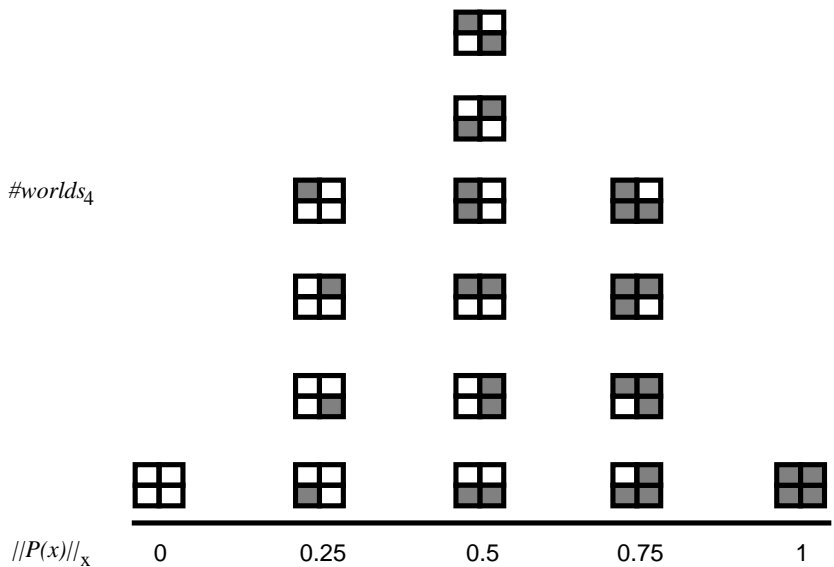

Figure 1: Partition of $\mathcal{W}_4$ according to $\pi(W)$.

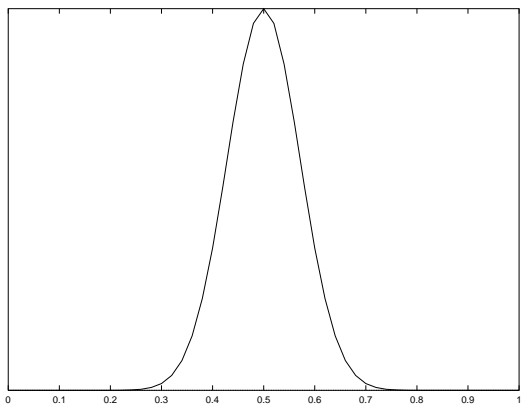

Figure 2: Concentration phenomenon for worlds of size $N$.

We can estimate the factorials appearing in this expression using Stirling's approximation, which asserts that the factorial $m!$ is approximately $m^m = e^{m \ln m}$. So, after substituting for the three factorials, we can estimate $\binom{N}{r}$ as $e^{N \log N - (r \log r + (N-r) \log(N-r))}$, which reduces to $e^{NH(\vec{u})}$. The entropy term in the general case arises from the use of Stirling's approximation in an analogous way. (A more careful estimate is done in the proof of Lemma 3.11 in the appendix.) ∎

Because of the exponential dependence on $N$ times the entropy, the number of worlds associated with points of high entropy swamp all other worlds as $N$ grows large. This concentration phenomenon, well-known in the field of statistical physics, forms the basis for our main result in this section. It asserts that it is possible to compute degrees of belief according to random worlds while ignoring all but those worlds whose entropy is near maximum. The next theorem essentially formalizes this phenomenon.

**Theorem 3.13:** *For all sufficiently small $\vec{\tau}$, the following is true. Let $\mathcal{Q}$ be the points with greatest entropy in $S^{\vec{\tau}}[KB]$ and let $\mathcal{O} \subseteq \mathbb{R}^K$ be any open set containing $\mathcal{Q}$. Then for all $\theta \in \mathcal{L}^{\approx}$ and for $\lim^{\star} \in \{\limsup, \liminf\}$ we have*

$$\lim_{N \to \infty}{}^{\star} \Pr_N^{\vec{\tau}}(\theta | KB) = \lim_{N \to \infty}{}^{\star} \frac{\# worlds_N^{\vec{\tau}}[\mathcal{O}](\theta \wedge KB)}{\# worlds_N^{\vec{\tau}}[\mathcal{O}](KB)}.$$

We remark that this is quite a difficult theorem. We have discussed why Lemma 3.11 lets us look at *models of KB* whose entropy is (near) maximum. But the theorem tells us to look at the maximum-entropy points of $S^{\vec{\tau}}[KB]$, which we defined using a (so far unmotivated) syntactic procedure applied to $KB$. It seems reasonable to expect that $S^{\vec{\tau}}[KB]$ should tell us *something* about models of $KB$. But making this connection precise, and in particular showing how the maximum-entropy points of $S^{\vec{\tau}}[KB]$ relate to models of $KB$ with near-maximum entropy, is difficult. However, we defer all details of the proof of that result to the appendix.

In general, Theorem 3.13 may seem to be of limited usefulness: knowing that we only have to look at worlds near the maximum-entropy point does not substantially reduce the number of worlds we need to consider. (Indeed, the whole point of the concentration phenomenon is that almost all worlds have high entropy.) Nevertheless, as the rest of this paper shows, this result can be quite useful when combined with the following two results. The first of these says that if all the worlds near the maximum-entropy points have a certain property, then we should have degree of belief 1 that this property is true.

**Corollary 3.14:** *For all sufficiently small $\vec{\tau}$, the following is true. Let $\mathcal{Q}$ be the points with greatest entropy in $S^{\vec{\tau}}[KB]$, let $\mathcal{O} \subseteq \mathbb{R}^K$ be an open set containing $\mathcal{Q}$, and let $\theta[\mathcal{O}] \in \mathcal{L}^{=}$ be an assertion that holds for every world $W$ such that $\pi(W) \in \mathcal{O}$. Then*

$$\Pr_\infty^{\vec{\tau}}(\theta[\mathcal{O}] | KB) = 1.$$

**Example 3.15:** For the knowledge base *true* in Example 3.12, it is easy to see that the maximum-entropy point is $(0.5, 0.5)$. Fix some arbitrary $\epsilon > 0$. Clearly, there is some open set $\mathcal{O}$ around this point such that the assertion $\theta = ||P(x)||_x \in [0.5 - \epsilon, 0.5 + \epsilon]$ holds for every world in $\mathcal{O}$. Therefore, we can conclude that

$$\Pr_\infty^{\vec{\tau}}(||P(x)||_x \in [0.5 - \epsilon, 0.5 + \epsilon] \,|\, true) = 1. \quad \blacksquare$$

As we show in (Bacchus et al., 1994), formulas $\theta$ with degree of belief 1 can essentially be treated just like other knowledge in $KB$. That is, the degrees of belief relative to $KB$ and $KB \wedge \theta$ will be identical (even if $KB$ and $KB \wedge \theta$ are not logically equivalent). More formally:

**Theorem 3.16:** (Bacchus et al., 1994) *If* $\Pr_{\infty}^{\vec{\tau}}(\theta|KB) = 1$ *and* $\lim^{*} \in \{\limsup, \liminf\}$, *then for any formula* $\varphi$:

$$\lim_{N \to \infty}{}^{*} \Pr_N^{\vec{\tau}}(\varphi|KB) = \lim_{N \to \infty}{}^{*} \Pr_N^{\vec{\tau}}(\varphi|KB \wedge \theta).$$

**Proof:** For completeness, we repeat the proof from (Bacchus et al., 1994) here. Basic probabilistic reasoning shows that, for any $N$ and $\vec{\tau}$:

$$\Pr_N^{\vec{\tau}}(\varphi|KB) = \Pr_N^{\vec{\tau}}(\varphi|KB \wedge \theta) \Pr_N^{\vec{\tau}}(\theta|KB) + \Pr_N^{\vec{\tau}}(\varphi|KB \wedge \neg\theta) \Pr_N^{\vec{\tau}}(\neg\theta|KB).$$

By assumption, $\Pr_N^{\vec{\tau}}(\theta|KB)$ tends to 1 when we take limits, so the first term tends to $\Pr_N^{\vec{\tau}}(\varphi|KB \wedge \theta)$. On the other hand, $\Pr_N^{\vec{\tau}}(\neg\theta|KB)$ has limit 0. Because $\Pr_N^{\vec{\tau}}(\varphi|KB \wedge \neg\theta)$ is bounded, we conclude that the second product also tends to 0. The result follows. ∎

As we shall see in the next section, the combination of Corollary 3.14 and Theorem 3.16 is quite powerful.

# 4. Computing degrees of belief

Although the concentration phenomenon is interesting, its application to actually computing degrees of belief may not be obvious. Since we know that almost all worlds will have high entropy, a direct application of Theorem 3.13 does not substantially reduce the number of worlds we must consider. Yet, as we show in this section, the concentration theorem can form the basis of a practical technique for computing degrees of belief in many cases. We begin in Section 4.1 by presenting the intuitions underlying this technique. In Section 4.2 we build on these intuitions by presenting results for a restricted class of formulas: those queries which are quantifier-free formulas over a unary language with a single constant symbol. In spite of this restriction, many of the issues arising in the general case can be seen here. Moreover, as we show in Section 4.3, this restricted sublanguage is rich enough to allow us to embed two well-known propositional approaches that make use of maximum entropy: Nilsson's *probabilistic logic* (Nilsson, 1986) and the maximum-entropy extension of $\epsilon$-semantics (Geffner & Pearl, 1990) due to Goldszmidt, Morris, Pearl (1990) (see also (Goldszmidt, Morris, & Pearl, 1993)). In Section 4.4, we consider whether the results for the restricted language can be extended. We show that they can, but several difficult and subtle issues arise.

## 4.1 The general strategy

Although the random-worlds method is defined by counting worlds, we can sometimes find more direct ways to calculate the degrees of belief it yields. In (Bacchus et al., 1994) we present a number of such techniques, most of which apply only in very special cases. One of the simplest and most intuitive is the following version of what philosophers have termed *direct inference* (Reichenbach, 1949). Suppose that all we know about an individual $c$ is some assertion $\psi(c)$; in other words, $KB$ has the form $\psi(c) \wedge KB'$, and the constant $c$ does not appear in $KB'$. Also suppose that $KB$, together with a particular tolerance $\vec{\tau}$, implies that $\|\varphi(x)|\psi(x)\|_x$ is in some interval $[\alpha, \beta]$. It seems reasonable to argue that $c$ is should be treated as a "typical" element satisfying $\psi(x)$, because by assumption $KB$ contains no information suggesting otherwise. Therefore, we might hope to use the statistics directly, and conclude that $\Pr_\infty^{\vec{\tau}}(\varphi(c)|KB) \in [\alpha, \beta]$. This is indeed the case, as the following theorem shows.

**Theorem 4.1:** (Bacchus et al., 1994) *Let $KB$ be a knowledge base of the form $\psi(\vec{c}) \wedge KB'$, and assume that for all sufficiently small tolerance vectors $\vec{\tau}$,*

$$KB[\vec{\tau}] \models \|\varphi(\vec{x})|\psi(\vec{x})\|_{\vec{x}} \in [\alpha, \beta].$$

*If no constant in $\vec{c}$ appears in $KB'$, in $\varphi(\vec{x})$, or in $\psi(\vec{x})$, then $\Pr_\infty(\varphi(\vec{c})|KB) \in [\alpha, \beta]$ (if the degree of belief exists at all).*

This result, in combination with the results of the previous section, provides us with a very powerful tool. Roughly speaking, we propose to use the following strategy: The basic concentration phenomenon says that most worlds are very similar in a certain sense. As shown in Corollary 3.14, we can use this to find some assertions that are "almost certainly" true (i.e., with degree of belief 1) even if they are not logically implied by $KB$. Theorem 3.16 then tells us that we can treat these new assertions as if they are in fact known with certainty. When these new assertions state statistical "knowledge", they can vastly increase our opportunities to apply direct inference. The following example illustrates this idea.

**Example 4.2:** Consider a very simple knowledge base over a vocabulary containing the single unary predicate $\{P\}$:
$$KB = (\|P(x)\|_x \preceq_1 0.3).$$
There are two atoms $A_1$ and $A_2$ over $\mathcal{P}$, with $A_1 = P$ and $A_2 = \neg P$. The solution space of this $KB$ given $\vec{\tau}$ is clearly

$$S^{\vec{\tau}}[KB] = \{(u_1, u_2) \in \Delta^2 \ : \ u_1 \leq 0.3 + \tau_1\}.$$

A straightforward computation shows that, for $\tau_1 < 0.2$, this has a unique maximum-entropy point $\vec{v} = (0.3 + \tau_1, 0.7 - \tau_1)$.

Now, consider the query $P(c)$. For all $\epsilon > 0$, let $\theta[\epsilon]$ be the formula $\|P(x)\|_x \in [(0.3 + \tau_1) - \epsilon, (0.3 + \tau_1) + \epsilon]$. This satisfies the condition of Corollary 3.14, so it follows that $\Pr_\infty^{\vec{\tau}}(\theta[\epsilon]|KB) = 1$. Using Theorem 3.16, we know that for $\lim^* \in \{\liminf, \limsup\}$,

$$\lim_{N \to \infty}^* \Pr_N^{\vec{\tau}}(P(c)|KB) = \lim_{N \to \infty}^* \Pr_N^{\vec{\tau}}(P(c)|KB \wedge \theta[\epsilon]).$$

But now we can use direct inference. (Note that here, our "knowledge" about $c$ is vacuous, i.e., "$true(c)$".) We conclude that, if there is any limit at all, then necessarily

$$\Pr_\infty^{\vec{\tau}}(P(c)|KB \wedge \theta[\epsilon]) \in [(0.3 + \tau_1) - \epsilon, (0.3 + \tau_1) + \epsilon].$$

So, for all $\epsilon > 0$,
$$\Pr_\infty^{\vec{\tau}}(P(c)|KB) \in [(0.3 + \tau_1) - \epsilon, (0.3 + \tau_1) + \epsilon].$$

Since this is true for all $\epsilon$, the only possible value for $\Pr_\infty^{\vec{\tau}}(P(c)|KB)$ is $0.3 + \tau_1$, which is the value of $u_1$ (i.e., $\|P(x)\|_x$) *at the maximum-entropy point. Note that it is also clear what happens as $\vec{\tau}$ tends to $\vec{0}$: $\Pr_\infty(P(c)|KB)$ is $0.3$.* ■

This example demonstrates the main steps of one possible strategy for computing degrees of belief. First the maximum-entropy points of the space $S^{\vec{\tau}}[KB]$ are computed as a function of $\vec{\tau}$. Then, these are used to compute $\mathrm{Pr}_\infty^{\vec{\tau}}(\varphi|KB)$, assuming the limit exists (if not, the lim sup and lim inf of $\mathrm{Pr}_N(\varphi|KB)$ are computed instead). Finally, we compute the limit of this probability as $\vec{\tau}$ goes to zero.

Unfortunately, this strategy has a serious potential problem. We clearly cannot compute $\mathrm{Pr}_\infty^{\vec{\tau}}(\varphi|KB)$ separately for each of the infinitely many tolerance vectors $\vec{\tau}$ and then take the limit as $\vec{\tau}$ goes to 0. We might hope to compute this probability as an explicit function of $\vec{\tau}$, and then compute the limit. For instance, in Example 4.2 $\mathrm{Pr}_\infty^{\vec{\tau}}(P(c)|KB)$ was found to be $0.3 + \tau_1$, and so it is easy to see what happens as $\tau_1 \to 0$. But there is no reason to believe that $\mathrm{Pr}_\infty^{\vec{\tau}}(\varphi|KB)$ is, in general, an easily characterizable function of $\vec{\tau}$. If it is not, then computing the limit as $\vec{\tau}$ goes to 0 can be difficult or impossible. We would like to find a way to avoid this explicit limiting process altogether. It turns out that this is indeed possible in some circumstances. The main requirement is that the maximum-entropy points of $S^{\vec{\tau}}[KB]$ converge to the maximum-entropy points of $S^{\vec{0}}[KB]$. (For future reference, notice that $S^{\vec{0}}[KB]$ is the closure of the solution space of the constraints obtained from $KB$ by replacing all occurrences of $\approx_i$ by $=$ and all occurrences of $\preceq_i$ by $\leq$.) In many such cases, we can compute $\mathrm{Pr}_\infty(\varphi|KB)$ directly in terms of the maximum-entropy points of $S^{\vec{0}}[KB]$, without taking limits at all.

As the following example shows, this type of continuity does not hold in general: the maximum-entropy points of $S^{\vec{\tau}}[KB]$ do not necessarily converge to those of $S^{\vec{0}}[KB]$.

**Example 4.3:** Consider the knowledge base

$$KB = (||P(x)||_x \approx_1 0.3 \vee ||P(x)||_x \approx_2 0.4) \wedge ||P(x)||_x \not\approx_3 0.4 \ .$$

It is easy to see that $S^{\vec{0}}[KB]$ is just $\{(0.3, 0.7)\}$: The point $(0.4, 0.6)$ is disallowed by the second conjunct. Now, consider $S^{\vec{\tau}}[KB]$ for $\vec{\tau} > \vec{0}$. If $\tau_2 \leq \tau_3$, then $S^{\vec{\tau}}[KB]$ indeed does not contain points where $u_1$ is near 0.4; the maximum-entropy point of this space is easily seen to be $0.3 + \tau_1$. However, if $\tau_2 > \tau_3$ then there will be points in $S^{\vec{\tau}}[KB]$ where $u_1$ is around 0.4; for instance, those where $0.4 + \tau_3 < u_1 \leq 0.4 + \tau_2$. Since these points have a higher entropy than the points in the vicinity of 0.3, the former will dominate. Thus, the set of maximum-entropy points of $S^{\vec{\tau}}[KB]$ does not converge to a single well-defined set. What it converges to (if anything) depends on how $\vec{\tau}$ goes to $\vec{0}$. This nonconvergence has consequences for degrees of belief. It is not hard to show $\mathrm{Pr}_\infty^{\vec{\tau}}(P(c)|KB)$ can be either $0.3 + \tau_1$ or $0.4 + \tau_2$, depending on the precise relationship between $\tau_1$, $\tau_2$, and $\tau_3$. It follows that $\mathrm{Pr}_\infty(P(c)|KB)$ does not exist. ■

We say that a degree of belief $\mathrm{Pr}_\infty(\varphi|KB)$ is not *robust* if the behavior of $\mathrm{Pr}_\infty^{\vec{\tau}}(\varphi|KB)$ (or of lim inf $\mathrm{Pr}_N^{\vec{\tau}}(\varphi|KB)$ and lim sup $\mathrm{Pr}_N^{\vec{\tau}}(\varphi|KB)$) as $\vec{\tau}$ goes to $\vec{0}$ depends on *how* $\vec{\tau}$ goes to $\vec{0}$. In other worlds, nonrobustness describes situations when $\mathrm{Pr}_\infty(\varphi|KB)$ does not exist because of sensitivity to the exact choice of tolerances. We shall see a number of other examples of nonrobustness in later sections.

It might seem that the notion of robustness is an artifact of our approach. In particular, it seems to depend on the fact that our language has the expressive power to say that the two tolerances represent a different degree of approximation, simply by using different subscripts ($\approx_2$ vs. $\approx_3$ in the example). In an approach to representing approximate equality that does not make these distinctions, we are bound to get the answer 0.3 in the example above, since then $||P(x)||_x \not\approx_3 0.4$ really would be the negation of $||P(x)||_x \approx_2 0.4$. We would argue that the answer 0.3 is not as reasonable as it might at first seem. Suppose one of the two different instances of 0.4 in the previous example had been slightly different; for example, suppose we had used 0.399 rather than 0.4 in the first of them. In this case, the second conjunct is essentially vacuous, and can be ignored. The maximum-entropy point in $S^{\vec{0}}[KB]$ is now 0.399, and we indeed derive a degree of belief of 0.399 in $P(c)$. Thus, arbitrarily small changes to the numbers in the original knowledge base can cause large changes in our degrees of belief. But these numbers are almost always the result of approximate observations; this is reflected by our decision to use approximate equality rather than equality when referring to them. It does not seem reasonable to base actions on a degree of belief that can change so drastically in the face of small changes in the measurement of data. Note that, if we know that the two instances of 0.4 do, in fact, denote exactly the same number, we can represent this by using the same approximate equality connective in both disjuncts. In this case, it is easy to see that we do get the answer 0.3.

A close look at the example shows that the nonrobustness arises because of the negated proportion expression $||P(x)||_x \not\approx_3 0.4$. Indeed, we can show that if we start with a $KB$ in canonical form that does not contain negated proportion expressions then, in a precise sense, the set of maximum-entropy points of $S^{\vec{\tau}}[KB]$ necessarily converges to the set of maximum-entropy points of $S^{\vec{0}}[KB]$. An argument can be made that we should eliminate negated proportion expressions from the language altogether. It is one thing to argue that sometimes we have statistical values whose accuracy we are unsure about, so that we want to make logical assertions less stringent than exact numerical equality. It is harder to think of cases in which the opposite is true, and all we know is that some statistic is "not even approximately" equal to some value. However, we do not eliminate negated proportion expressions from the language, since without them we would not be able to prove an analogue to Theorem 3.5. (They arise when we try to flatten nested proportion expressions, for example.) Instead, we have identified a weaker condition that is sufficient to prevent problems such as that seen in Example 4.3. *Essential positivity* simply tests that negations are not interacting with the maximum-entropy computation in a harmful way.

**Definition 4.4:** Let $\Gamma^{\leq}(KB[\vec{0}])$ be the result of replacing each strict inequality in $\Gamma(KB[\vec{0}])$ with its weakened version. More formally, we replace each subformula of the form $t < 0$ with $t \leq 0$, and each subformula of the form $t > 0$ with $t \geq 0$. (Recall that these are the only constraints possible in $\Gamma(KB[\vec{0}])$, since all tolerance variables $\varepsilon_i$ are assigned 0.) Let $S^{\leq \vec{0}}[KB]$ be $\overline{Sol[\Gamma^{\leq}(KB[\vec{0}])]}$, where we use $\overline{X}$ to denote the closure of $X$. We say that $KB$ is *essentially positive* if the sets $S^{\leq \vec{0}}[KB]$ and $S^{\vec{0}}[KB]$ have the same maximum-entropy points. ∎

**Example 4.5:** Consider again the knowledge base $KB$ from Example 4.3. The constraint formula $\Gamma(KB[\vec{0}])$ is (after simplification):

$$(u_1 = 0.3 \lor u_1 = 0.4) \land (u_1 < 0.4 \lor u_1 > 0.4).$$

Its "weakened" version is $\Gamma^{\leq}(KB[\vec{0}])$:

$$(u_1 = 0.3 \lor u_1 = 0.4) \land (u_1 \leq 0.4 \lor u_1 \geq 0.4),$$

which is clearly equivalent to $u_1 = 0.3 \lor u_1 = 0.4$. Thus, $S^{\vec{0}}[KB] = \{(u_1, u_2) \in \Delta^2 : u_1 \leq 0.3\}$ whereas $S^{\leq \vec{0}}[KB] = S^{\vec{0}}[KB] \cup \{(0.4, 0.6)\}$. Since the two spaces have different maximum-entropy points, the knowledge base $KB$ is not essentially positive. ∎

As the following result shows, essential positivity suffices to guarantee that the maximum-entropy points of $S^{\vec{\tau}}[KB]$ converge to those of $S^{\vec{0}}[KB]$.

**Proposition 4.6:** *Assume that $KB$ is essentially positive and let $\mathcal{Q}$ be the set of maximum-entropy points of $S^{\vec{0}}[KB]$ (and thus also of $S^{\leq \vec{0}}[KB]$). Then for all $\epsilon > 0$ and all sufficiently small tolerance vectors $\vec{\tau}$ (where "sufficiently small" may depend on $\epsilon$), every maximum-entropy point of $S^{\vec{\tau}}[KB]$ is within $\epsilon$ of some maximum-entropy point in $\mathcal{Q}$.*

## 4.2 Queries for a single individual

We now show how to compute $\Pr_\infty(\varphi|KB)$ for a certain restricted class of first-order formulas $\varphi$ and knowledge bases $KB$. The most significantly restriction is that the query $\varphi$ should be a quantifier-free (first-order) sentence over the vocabulary $\mathcal{P} \cup \{c\}$; thus, it is a query about a single individual, $c$. While this class is rather restrictive, it suffices to express many real-life examples. Moreover, it is significantly richer than the language considered by Paris and Vencovska (1989).

The following definition helps define the class of interest.

**Definition 4.7:** A formula is *essentially propositional* if it is a quantifier-free and proportion-free formula in the language $\mathcal{L}^{\approx}(\{P_1, \ldots, P_k\})$ (so that, in particular, it has no constant symbols) and has only one free variable $x$. ∎

We say that $\varphi(c)$ is a *simple query for $KB$* if:

- $\varphi(x)$ is essentially propositional,

- $KB$ is of the form $\psi(c) \land KB'$, where $\psi(x)$ is essentially propositional and $KB'$ does not mention $c$.

Thus, just as in Theorem 4.1, we suppose that $\psi(c)$ summarizes all that is known about $c$. In this section, we focus on computing the degree of belief $\Pr_\infty(\varphi(c)|KB)$ for a formula $\varphi(c)$ which is a simple query for $KB$.

Note that an essentially propositional formula $\xi(x)$ is equivalent to a disjunction of atoms. For example, over the vocabulary $\{P_1, P_2\}$, the formula $P_1(x) \lor P_2(x)$ is equivalent to $A_1(x) \lor A_2(x) \lor A_3(x)$ (where the atoms are ordered as in Example 3.2). For an essentially propositional formula $\xi$, we take $\mathcal{A}(\xi)$ be the (unique) set of atoms such that $\xi$ is equivalent to $\bigvee_{A_j \in \mathcal{A}(\xi)} A_j(x)$.

If we view a tuple $\vec{u} \in \Delta^K$ as a probability assignment to the atoms, we can extend $\vec{u}$ to a probability assignment on all essentially propositional formulas using this identification of an essentially propositional formula with a set of atoms:

**Definition 4.8:** Let $\xi$ be an essentially propositional formula. We define a function $F_{[\xi]} : \Delta^K \to \mathbb{R}$ as follows:

$$F_{[\xi]}(\vec{u}) = \sum_{A_j \in \mathcal{A}(\xi)} u_j.$$

For essentially propositional formulas $\varphi(x)$ and $\psi(x)$ we define the (partial) function $F_{[\varphi|\psi]} : \Delta^K \to \mathbb{R}$ to be:

$$F_{[\varphi|\psi]}(\vec{u}) = \frac{F_{[\varphi \wedge \psi]}(\vec{u})}{F_{[\psi]}(\vec{u})}.$$

Note that this function is undefined when $F_{[\psi]}(\vec{u}) = 0$. ∎

As the following result shows, if $\varphi$ is a simple query for $KB$ (of the form $\psi(c) \wedge KB'$), then all that matters in computing $\Pr_\infty(\varphi|KB)$ is $F_{[\varphi|\psi]}(\vec{u})$ for tuples $\vec{u}$ of maximum entropy. Thus, in a sense, we are only using $KB'$ to determine the space over which we maximize entropy. Having defined this space, we can focus on $\psi$ and $\varphi$ in determining the degree of belief.

**Theorem 4.9:** *Suppose $\varphi(c)$ is a simple query for $KB$. For all $\vec{\tau}$ sufficiently small, if $\mathcal{Q}$ is the set of maximum-entropy points in $S^{\vec{\tau}}[KB]$ and $F_{[\psi]}(\vec{v}) > 0$ for all $\vec{v} \in \mathcal{Q}$, then for $\lim^* \in \{\limsup, \liminf\}$ we have*

$$\lim_{N \to \infty}{}^* \Pr_N^{\vec{\tau}}(\varphi(c)|KB) \in \left[\inf_{\vec{v} \in \mathcal{Q}} F_{[\varphi|\psi]}(\vec{v}), \sup_{\vec{v} \in \mathcal{Q}} F_{[\varphi|\psi]}(\vec{v})\right].$$

The following is an immediate but important corollary of this theorem. It asserts that, if the space $S^{\vec{\tau}}[KB]$ has a unique maximum-entropy point, then its value uniquely determines the probability $\Pr_\infty^{\vec{\tau}}(\varphi(c)|KB)$.

**Corollary 4.10:** *Suppose $\varphi(c)$ is a simple query for $KB$. For all $\vec{\tau}$ sufficiently small, if $\vec{v}$ is the unique maximum-entropy point in $S^{\vec{\tau}}[KB]$ and $F_{[\psi]}(\vec{v}) > 0$, then*

$$\Pr_\infty^{\vec{\tau}}(\varphi(c)|KB) = F_{[\varphi|\psi]}(\vec{v}).$$

We are interested in $\Pr_\infty(\varphi(c)|KB)$, which means that we are interested in the limit of $\Pr_\infty^{\vec{\tau}}(\varphi(c)|KB)$ as $\vec{\tau} \to \vec{0}$. Suppose $KB$ is essentially positive. Then, by the results of the previous section and the continuity of $F_{[\varphi|\psi]}$, it is enough to look directly at the maximum-entropy point of $S^{\vec{0}}[KB]$. More formally, by combining Theorem 4.9 with Proposition 4.6, we can show:

**Theorem 4.11:** *Suppose $\varphi(c)$ is a simple query for $KB$. If the space $S^{\vec{0}}[KB]$ has a unique maximum-entropy point $\vec{v}$, $KB$ is essentially positive, and $F_{[\psi]}(\vec{v}) > 0$, then*

$$\Pr_\infty(\varphi(c)|KB) = F_{[\varphi|\psi]}(\vec{v}).$$

We believe that this theorem will turn out to cover a lot of cases that occur in practice. As our examples and the discussion in the next section show, we often do get simple queries and knowledge bases that are essentially positive. Concerning the assumption of a unique maximum-entropy point, note that the entropy function is convex and so this assumption is automatically satisfied if $S^{\vec{0}}[KB]$ is a *convex* space. Recall that a space $S$ is convex if for all $\vec{u}, \vec{u}' \in S$, and all $\alpha \in [0, 1]$, it is also the case that $\alpha \vec{u} + (1 - \alpha) \vec{u}' \in S$. The space $S^{\vec{0}}[KB]$ is surely convex if it is defined using a conjunction of linear constraints. While it is clearly possible to create knowledge bases where $S^{\vec{0}}[KB]$ has multiple maximum-entropy points (for example, using disjunctions), we expect that such knowledge bases arise rarely in practical applications. Perhaps the most restrictive assumption made by this theorem is the seemingly innocuous requirement that $F_{[\psi]}(\vec{v}) > 0$. This assumption is obviously necessary for the theorem to hold; without it, the function $F_{[\varphi|\psi]}$ is simply not defined. Unfortunately, we show in Section 4.4 that this requirement is, in fact, a severe one; in particular, it prevents the theorem from being applied to most examples derived from default reasoning, using our statistical interpretation of defaults (Bacchus et al., 1994).

We close this subsection with an example of the theorem in action.

**Example 4.12:** Let the language consist of $\mathcal{P} = \{Hepatitis, Jaundice, BlueEyed\}$ and the constant $Eric$. There are eight atoms in this language. We use $A_{P_1'P_2'P_3'}$ to denote the atom $P_1'(x) \wedge P_2'(x) \wedge P_3'(x)$, where $P_1'$ is either $H$ (denoting $Hepatitis$) or $\overline{H}$ (denoting $\neg Hepatitis$), $P_2'$ is $J$ or $\overline{J}$ (for $Jaundice$ and $\neg Jaundice$, respectively), and $P_3'$ is $B$ or $\overline{B}$ (for $BlueEyed$ and $\neg BlueEyed$, respectively).

Consider the knowledge base $KB_{hep}$:

$$\forall x \, (Hepatitis(x) \Rightarrow Jaundice(x)) \, \wedge$$
$$\|Hepatitis(x)|Jaundice(x)\|_x \approx_1 0.8 \, \wedge$$
$$\||BlueEyed(x)\|_x \approx_2 0.25 \, \wedge$$
$$Jaundice(Eric).$$

If we order the atoms as $A_{HJB}, A_{HJ\overline{B}}, A_{H\overline{J}B}, A_{H\overline{J}\overline{B}}, A_{\overline{H}JB}, A_{\overline{H}J\overline{B}}, A_{\overline{H}\,\overline{J}B}, A_{\overline{H}\,\overline{J}\,\overline{B}}$, then it is not hard to show that $\Gamma(KB_{hep})$ is:

$$
\begin{array}{lcll}
u_3 & = & 0 & \wedge \\
u_4 & = & 0 & \wedge \\
(u_1 + u_2) & \leq & (0.8 + \varepsilon_1)(u_1 + u_2 + u_5 + u_6) & \wedge \\
(u_1 + u_2) & \geq & (0.8 - \varepsilon_1)(u_1 + u_2 + u_5 + u_6) & \wedge \\
(u_1 + u_3 + u_5 + u_7) & \leq & (0.25 + \varepsilon_2) & \wedge \\
(u_1 + u_3 + u_5 + u_7) & \geq & (0.25 - \varepsilon_2) & \wedge \\
(u_1 + u_2 + u_5 + u_6) & > & 0.
\end{array}
$$

To find the space $S^{\vec{0}}[KB_{hep}]$ we simply set $\varepsilon_1 = \varepsilon_2 = 0$. Then it is quite straightforward to find the maximum-entropy point in this space, which, taking $\gamma = 2^{1.6}$, is:

$$(v_1, v_2, v_3, v_4, v_5, v_6, v_7, v_8) = \left( \frac{1}{5 + \gamma}, \frac{3}{5 + \gamma}, 0, 0, \frac{1}{4(5 + \gamma)}, \frac{3}{4(5 + \gamma)}, \frac{\gamma}{4(5 + \gamma)}, \frac{3\gamma}{4(5 + \gamma)} \right).$$

Using $\vec{v}$, we can compute various asymptotic probabilities very easily. For example,

$$
\begin{aligned}
\Pr_\infty(Hepatitis(Eric)|KB_{hep}) &= F_{[Hepatitis|Jaundice]}(\vec{v}) \\
&= \frac{v_1 + v_2}{v_1 + v_2 + v_5 + v_6} \\
&= \frac{\frac{1}{5+\gamma} + \frac{3}{5+\gamma}}{\frac{1}{5+\gamma} + \frac{3}{5+\gamma} + \frac{1}{4(5+\gamma)}, \frac{3}{4(5+\gamma)}} = 0.8,
\end{aligned}
$$

as expected. Similarly, we can show that $\Pr_\infty(BlueEyed(Eric)|KB_{hep}) = 0.25$ and that $\Pr_\infty(BlueEyed(Eric) \wedge Hepatitis(Eric)|KB_{hep}) = 0.2$. Note that the first two answers also

follow from the direct inference principle (Theorem 4.1), which happens to be applicable in this case. The third answer shows that *BlueEyed* and *Hepatitis* are being treated as independent. It is a special case of a more general independence phenomenon that applies to random worlds; see (Bacchus et al., 1994, Theorem 5.27). ∎

## 4.3 Probabilistic propositional logic

In this section we consider two variants of *probabilistic propositional logic*. As the following discussion shows, both can easily be captured by our framework. The embedding we discuss uses simple queries throughout, allowing us to appeal to the results of the previous section.

Nilsson (1986) considered the problem of what could be inferred about the probability of certain propositions given some constraints. For example, we might know that $\Pr(fly|bird) \geq 0.7$ and that $\Pr(yellow) \leq 0.2$, and be interested in $\Pr(fly|bird \wedge yellow)$. Roughly speaking, Nilsson suggests computing this by considering all probability distributions consistent with the constraints, and then computing the range of values given to $\Pr(fly|bird \wedge yellow)$ by these distributions. Formally, suppose our language consists of $k$ primitive proposition, $p_1, \ldots, p_k$. Consider the set $\Omega$ of $K = 2^k$ truth assignments these propositions. We give semantics to probabilistic statements over this language in terms of a probability distribution $\mu$ over the set $\Omega$ (see (Fagin, Halpern, & Megiddo, 1990) for details). Since each truth assignment $\omega \in \Omega$ determines the truth value of every propositional formula $\beta$, we can determine the probability of every such formula:

$$\Pr_\mu(\beta) = \sum_{\omega \models \beta} \mu(\omega).$$

Clearly, we can determine whether a probability distribution $\mu$ satisfies a set $\Lambda$ of probabilistic constraints. The standard notion of probabilistic propositional inference would say that $\Lambda \models \Pr(\beta) \in [\lambda_1, \lambda_2]$ if $\Pr_\mu(\beta)$ is within the range $[\lambda_1, \lambda_2]$ for every distribution $\mu$ that satisfies the constraints in $\Lambda$.

Unfortunately, while this is a very natural definition, the constraints that one can derive from it are typically quite weak. For that reason, Nilsson suggested strengthening this notion of inference by applying the principle of maximum entropy: rather than considering all distributions $\mu$ satisfying $\Lambda$, we consider only the distribution(s) $\mu^*$ that have the greatest entropy among those satisfying the constraints. As we now show, one implication of our results is that the random-worlds method provides a principled motivation for this introduction of maximum entropy to probabilistic propositional reasoning. In fact, the connection between probabilistic propositional reasoning and random worlds should now be fairly clear:

- The primitive propositions $p_1, \ldots, p_k$ correspond to the unary predicates $P_1, \ldots, P_k$.

- A propositional formula $\beta$ over $p_1, \ldots, p_k$ corresponds uniquely to an essentially propositional formula $\xi_\beta$ as follows: we replace each occurrence of the propositional symbol $p_i$ with $P_i(x)$.

- The set $\Lambda$ of probabilistic constraints corresponds to a knowledge base $KB'[\Lambda]$—a constant-free knowledge base containing only proportion expressions. The correspondence is as follows:

- A probability expression of the form $\Pr(\beta)$ appearing in $\Lambda$ is replaced by the proportion expression $\|\xi_\beta(x)\|_x$. Similarly, a conditional probability expression $\Pr(\beta|\beta')$ is replaced by $\|\xi_\beta(x)|\xi_{\beta'}(x)\|_x$.
- Each comparison connective $=$ is replaced by $\approx_i$ for some $i$, and each $\leq$ with $\preceq_i$. (The particular choices for the approximate equality connectives do not matter in this context.)

The other elements that can appear in a proportion formula (such as rational numbers and arithmetical connectives) remain unchanged. For example, the formula $\Pr(fly|bird) \geq 0.7$ would correspond to the proportion formula $\|Fly(x)|Bird(x)\|_x \succeq_i 0.7$.

- There is a one-to-one correspondence between truth assignments and atoms: the truth assignment $\omega$ corresponds to the atom $A = P'_1 \wedge \ldots \wedge P'_k$ where $P'_i$ is $P_i$ if $\omega(p_i) = true$ and $\neg P_i$ otherwise. Let $\omega_1, \ldots, \omega_K$ be the truth assignments corresponding to the atoms $A_1, \ldots, A_K$, respectively.

- There is a one-to-one correspondence between probability distributions over the set $\Omega$ of truth assignments and points in $\Delta^K$. For each point $\vec{u} \in \Delta^K$, let $\mu_{\vec{u}}$ denote the corresponding probability distribution over $\Omega$, where $\mu_{\vec{u}}(\omega_i) = u_i$.

**Remark 4.13:** Clearly, $\omega_j \models \beta$ iff $A_j \in \mathcal{A}(\xi_\beta)$. Therefore, for all $\vec{u}$, we have

$$F_{[\xi_\beta]}(\vec{u}) = \Pr_{\mu_{\vec{u}}}(\beta). \quad \blacksquare$$

The following result demonstrates the tight connection between probabilistic propositional reasoning using maximum entropy and random worlds.

**Theorem 4.14:** *Let $\Lambda$ be a conjunction of constraints of the form $\Pr(\beta|\beta') = \lambda$ or $\Pr(\beta|\beta') \in [\lambda_1, \lambda_2]$. There is a unique probability distribution $\mu^\star$ of maximum entropy satisfying $\Lambda$. Moreover, for all $\beta$ and $\beta'$, if $\Pr_{\mu^\star}(\beta') > 0$, then*

$$\Pr_\infty(\xi_\beta(c)|\xi_{\beta'}(c) \wedge KB'[\Lambda]) = \Pr_{\mu^\star}(\beta|\beta').$$

Theorem 4.14 is an easy corollary of Theorem 4.11. To check that the preconditions of the latter theorem apply, note that the constraints in $\Lambda$ are linear, and so the space $S^{\vec{0}}[KB'[\Lambda]]$ has a unique maximum-entropy point $\vec{v}$. In fact, it is easy to show that $\mu_{\vec{v}}$ is the (unique) maximum-entropy probability distribution over $\Omega$ satisfying the constraints $\Lambda$. In addition, because there are no negated proportion expressions in $\Lambda$, the formula $KB = \xi_{\beta'}(c) \wedge KB'[\Lambda]$ is certainly essentially positive.

Most applications of probabilistic propositional reasoning consider simple constraints of the form used in the theorem, and so such applications can be viewed as very special cases of the random-words approach. In fact, this theorem is essentially a very old one. The connection between counting "worlds" and the entropy maximum in a space defined as a conjunction of linear constraints is very well-known. It has been extensively studied in the field of thermodynamics, starting with the 19th century work of Maxwell and Gibbs. Recently, this type of reasoning has been applied to problems in an AI context by Paris and Vencovska (1989) and Shastri (1989). The work of Paris and Vencovska is particularly relevant because they also realize the necessity of adopting a formal notion of "approximation", although the precise details of their approach differ from ours.

To the best of our knowledge, most of the work on probabilistic propositional reasoning and all formal presentations of the entropy/worlds connection (in particular, those of (Paris & Vencovska, 1989; Shastri, 1989)) have limited themselves to conjunctions of linear constraints. Our more general language gives us a great deal of additional expressive power. For example, it is quite reasonable to want the ability to express that properties are (approximately) statistically independent. For example, we may wish to assert that $Bird(x)$ and $Yellow(x)$ are independent properties by saying $||Bird(x) \wedge Yellow(x)||_x \approx ||Bird(x)||_x \cdot ||Yellow(x)||_x$. Clearly, such constraints are not linear. Nevertheless, our Theorem 4.11 covers such cases and much more.

A version of probabilistic propositional reasoning has also been used to provide probabilistic semantics for default reasoning (Pearl, 1989). Here also, the connection to random worlds is of interest. In particular, it follows from Corollary 4.10 that the recent work of Goldszmidt, Morris, and Pearl (1990) can be embedded in the random-worlds framework. In the rest of this subsection, we explain their approach and the embedding.

Consider a language consisting of propositional formulas over the propositional variables $p_1, \ldots, p_k$, and default rules of the form $B \to C$ (read "$B$'s are typically $C$'s"), where $B$ and $C$ are propositional formulas. A distribution $\mu$ is said to $\epsilon$-satisfy a default rule $B \to C$ if $\mu(C|B) \geq 1 - \epsilon$. In addition to default rules, the framework also permits the use of material implication in a rule, as in $B \Rightarrow C$. A distribution $\mu$ is said to satisfy such a rule if $\mu(C|B) = 1$. A *parameterized probability distribution* (PPD) is a collection $\{\mu_\epsilon\}_{\epsilon > 0}$ of probability distributions over $\Omega$, parameterized by $\epsilon$. A PPD $\{\mu_\epsilon\}_{\epsilon > 0}$ $\epsilon$-satisfies a set $\mathcal{R}$ of rules if for every $\epsilon$, $\mu_\epsilon$ $\epsilon$-satisfies every default rule $r \in \mathcal{R}$ and satisfies every non-default rule $r \in \mathcal{R}$. A set $\mathcal{R}$ of default rules $\epsilon$-entails $B \to C$ if for every PPD that $\epsilon$-satisfies $\mathcal{R}$, $\lim_{\epsilon \to 0} \mu_\epsilon(C|B) = 1$.

As shown in (Geffner & Pearl, 1990), $\epsilon$-entailment possesses a number of reasonable properties typically associated with default reasoning, including a preference for more specific information. However, there are a number of desirable properties that it does not have. Among other things, irrelevant information is not ignored. (See (Bacchus et al., 1994) for an extensive discussion of this issue.)

To obtain additional desirable properties, $\epsilon$-semantics is extended in (Goldszmidt et al., 1990) by an application of the principle of maximum entropy. Instead of considering all possible PPD's, as above, we consider only the PPD $\left\{\mu_{\epsilon, \mathcal{R}}^\star\right\}_{\epsilon > 0}$ such that, for each $\epsilon$, $\mu_{\epsilon, \mathcal{R}}^\star$ has the maximum entropy among distributions that $\epsilon$-satisfy all the rules in $\mathcal{R}$. (See (Goldszmidt et al., 1990) for precise definitions and technical details.) Note that, since the constraints used to define $\mu_{\epsilon, \mathcal{R}}^\star$ are all linear, there is indeed a unique such point of maximum entropy. A rule $B \to C$ is an *ME-plausible consequence* of $\mathcal{R}$ if $\lim_{\epsilon \to 0} \mu_{\epsilon, \mathcal{R}}^\star(C|B) = 1$. The notion of ME-plausible consequence is analyzed in detail in (Goldszmidt et al., 1990), where it is shown to inherit all the nice properties of $\epsilon$-entailment (such as the preference for more specific information), while successfully ignoring irrelevant information. Equally importantly, algorithms are provided for computing the ME-plausible consequences of a set of rules in certain cases.

Our maximum-entropy results can be used to show that the approach of (Goldszmidt et al., 1990) can be embedded in our framework in a straightforward manner. We simply translate a default rule $r$ of the form $B \to C$ into a first-order default rule

$$\theta_r =_{\text{def}} \| \xi_C(x) | \xi_B(x) \|_x \approx_1 1,$$

as in our earlier translation of Nilsson's approach. Note that the formulas that arise under this translation all use the same approximate equality connective $\approx_1$. The reason is that the approach of (Goldszmidt et al., 1990) uses the same $\epsilon$ for all default rules. We can similarly translate a (non-default) rule $r$ of the form $B \Rightarrow C$ into a first-order constraint using universal quantification:

$$\theta_r =_{\text{def}} \forall x \, (\xi_B(x) \Rightarrow \xi_C(x)).$$

Under this translation, we can prove the following theorem.

**Theorem 4.15:** *Let $c$ be a constant symbol. Using the translation described above, for a set $\mathcal{R}$ of defeasible rules, $B \to C$ is an ME-plausible consequence of $\mathcal{R}$ iff*

$$\text{Pr}_\infty \left( \xi_C(c) \,\middle|\, \xi_B(c) \wedge \bigwedge_{r \in \mathcal{R}} \theta_r \right) = 1.$$

In particular, this theorem implies that all the computational techniques and results described in (Goldszmidt et al., 1990) carry over to this special case of the random-worlds method. It also shows that random-world provides a principled justification for the approach (Goldszmidt et al., 1990) present (one which is quite different from the justification given in (Goldszmidt et al., 1990) itself).

## 4.4 Beyond simple queries

In Section 4.2 we restricted attention to simple queries. Our main result, Theorem 4.11, needed other assumptions as well: essential positivity, the existence of a unique maximum-entropy point $\vec{v}$, and the requirement that $F_{[\psi]}(\vec{v}) > 0$. We believe that this theorem is useful in spite of its limitations, as demonstrated by the discussion in Section 4.3. Nevertheless, this result allows us to take advantage of only a small fragment of our rich language. Can we find a more general theorem? After all, the basic concentration result (Theorem 3.13) holds with essentially no restrictions. In this section we show that it is indeed possible to extend Theorem 4.11 significantly. However, there are serious limitations and subtleties. We illustrate these problems by means of examples, and then state an extended result.

Our attempt to address these problems (so far as is possible) leads to a rather complicated final result. In fact, the problems we discuss are as interesting and important as the theorem we actually give: they help us understand more of the limits of maximum entropy. Of course, every issue we discuss in this subsection is relatively minor compared to maximum entropy's main (apparent) restriction, which concerns the use of non-unary predicates. For the reader who is less concerned about the other, lesser, issues we remark that it is possible to skip directly to Section 5.

We first consider the restrictions we placed on the $KB$, and show the difficulties that arise if we drop them. We start with the restriction to a single maximum-entropy point. As the concentration theorem (Theorem 3.13) shows, the entropy of almost every world is near maximum. But it does not follow that all the maximum-entropy points are surrounded by similar numbers of worlds. Thus, in the presence of more than one maximum-entropy point, we face the problem of finding the relative importance, or weighting, of each maximum-entropy point. As the following example illustrates, this weighting is often sensitive to the tolerance values. For this reason, non-unique entropy maxima often lead to nonrobustness.

**Example 4.16:** Suppose $\Phi = \{P, c\}$, and consider the knowledge base

$$KB = (\|P(x)\|_x \preceq_1 0.3) \vee (\|P(x)\|_x \succeq_2 0.7).$$

Assume we want to compute $\Pr_\infty(P(c)|KB)$. In this case, $S^{\vec{\tau}}[KB]$ is

$$\{(u_1, u_2) \in \Delta^2 \ : \ u_1 \leq 0.3 + \tau_1 \text{ or } u_1 \geq 0.7 - \tau_2\},$$

and $S^{\vec{0}}[KB]$ is

$$\{(u_1, u_2) \in \Delta^2 \ : \ u_1 \leq 0.3 \text{ or } u_1 \geq 0.7\}.$$

Note that $S^{\vec{0}}[KB]$ has two maximum-entropy points: $(0.3, 0.7)$ and $(0.7, 0.3)$.

Now consider the maximum-entropy points of $S^{\vec{\tau}}[KB]$ for $\vec{\tau} > \vec{0}$. It is not hard to show that if $\tau_1 > \tau_2$, then this space has a unique maximum-entropy point, $(0.3 + \tau_1, 0.7 - \tau_1)$. In this case, $\Pr_\infty^{\vec{\tau}}(P(c)|KB) = 0.3 + \tau_1$. On the other hand, if $\tau_1 < \tau_2$, then the unique maximum-entropy point of this space is $(0.7 + \tau_2, 0.3 - \tau_2)$, in which case $\Pr_\infty^{\vec{\tau}}(P(c)|KB) = 0.7 + \tau_2$. If $\tau_1 = \tau_2$, then the space $S^{\vec{\tau}}[KB]$ has two maximum-entropy points, and by symmetry we obtain that $\Pr_\infty^{\vec{\tau}}(P(c)|KB) = 0.5$. So, by appropriately choosing a sequence of tolerance vectors converging to $\vec{0}$, we can make the asymptotic value of this fraction either 0.3, 0.5, or 0.7. Thus $\Pr_\infty(P(c)|KB)$ does not exist.

It is not disjunctions *per se* that cause the problem here: if we consider instead the database $KB' = (\|P(x)\|_x \preceq_1 0.3) \vee (\|P(x)\|_x \succeq_2 0.6)$, then there is no difficulty. There is a unique maximum-entropy point of $S^{\vec{0}}[KB']$—$(0.6, 0.4)$—and the asymptotic probability $\Pr_\infty(P(c)|KB') = 0.6$, as we would want.[7] ∎

In light of this example (and many similar ones we can construct), we continue to assume that there is a single maximum-entropy point. As we argued earlier, we expect this to be true in typical practical applications, so the restriction does not seem very serious.

We now turn our attention to the requirement that $F_{[\psi]}(\vec{v}) > 0$. As we have already observed, this seems to be an obvious restriction to make, considering that the function $F_{[\varphi|\psi]}(\vec{v})$ is not defined otherwise. However, this difficulty is actually a manifestation of a much deeper problem. As the following example shows, any approach that just uses the maximum-entropy point of $S^{\vec{0}}[KB]$ will necessarily fail in some cases where $F_{[\psi]}(\vec{v}) = 0$.

**Example 4.17:** Consider the knowledge base

$$KB = (\|Penguin(x)\|_x \approx_1 0) \wedge (\|Fly(x)|Penguin(x)\|_x \approx_2 0) \wedge Penguin(Tweety).$$

---

7. We remark that it is also possible to construct examples of multiple maximum-entropy points by using quadratic constraints rather than disjunction.

Suppose we want to compute $\Pr_\infty(Fly(Tweety)|Penguin(Tweety))$. We can easily conclude from Theorem 4.1 that this degree of belief is 0, as we would expect. However, we cannot reach this conclusion using Theorem 4.11 or anything like it. For consider the maximum-entropy point of $S^{\vec{0}}[KB]$. The coordinates $v_1$, corresponding to $Fly \wedge Penguin$, and $v_2$, corresponding to $\neg Fly \wedge Penguin$, are both 0. Hence, $F_{[Penguin]}(\vec{v}) = 0$, so that Theorem 4.11 does not apply.

But, as we said, the problem is more fundamental. The information we need (that the proportion of flying penguins is zero) is simply not present if all we know is the maximum-entropy point $\vec{v}$. We can obtain the same space $S^{\vec{0}}[KB]$ (and thus the same maximum-entropy point) from quite different knowledge bases. In particular, consider $KB'$ which simply asserts that $(||Penguin(x)||_x \approx_1 0) \wedge Penguin(Tweety)$. This new knowledge base tells us nothing whatsoever about the fraction of flying penguins, and in fact it is easy to show that $\Pr_\infty(Fly(Tweety)|KB') = 0.5$. But of course it is impossible to distinguish this case from the previous one just by looking at $\vec{v}$. It follows that no result in the spirit of Theorem 4.11 (which just uses the value of $\vec{v}$) can be comprehensive. ■

The example shows that the philosophy behind Theorem 4.11 cannot be extended very far, if at all: it is inevitable that there will be problems when $F_{[\psi]}(\vec{v}) = 0$. But it is natural to ask whether there is a different approach altogether in which this restriction can be relaxed. That is, is it possible to construct a technique for computing degrees of belief in those cases where $F_{[\psi]} = 0$? As we mentioned in Section 4.1, we might hope to do this by computing $\Pr_\infty^{\vec{\tau}}(\varphi|KB)$ as a function of $\vec{\tau}$ and then taking the limit as $\vec{\tau}$ goes to 0. In general, this seems very hard. But, interestingly, the computational technique of (Goldszmidt et al., 1990) does use this type of parametric analysis, demonstrating that things might not be so bad for various restricted cases. Another source of hope is to remember that maximum entropy is, for us, merely one tool for computing random-worlds degrees of belief. There may be other approaches that bypass entropy entirely. In particular, some of the theorems we give in (Bacchus et al., 1994) can be seen as doing this; these theorems will often apply even if $F_{[\psi]} = 0$.

Another assumption made throughout Section 4.2 is that the knowledge base has a special form, namely $\psi(c) \wedge KB'$, where $\psi$ is essentially propositional and $KB'$ does not contain any occurrences of $c$. The more general theorem we state later relaxes this somewhat, as follows.

**Definition 4.18:** A knowledge base $KB$ is said to be *separable with respect to query* $\varphi$ if it has the form $\psi \wedge KB'$, where $\psi$ contains neither quantifiers nor proportions, and $KB'$ contains none of the constant symbols appearing in $\varphi$ or in $\psi$.[8] ■

It should be clear that if a query $\varphi(c)$ is simple for $KB$ (as assumed in previous subsection), then the separability condition is satisfied.

As the following example shows, if we do not assume separability, we can easily run into nonrobust behavior:

**Example 4.19:** Consider the following knowledge base $KB$ over the vocabulary $\Phi = \{P, c\}$:

$$(||P(x)||_x \approx_1 0.3 \wedge P(c)) \vee (||P(x)||_x \approx_2 0.3 \wedge \neg P(c)).$$

---

8. Clearly, since our approach is semantic, it also suffices if the knowledge base is equivalent to one of this form.

$KB$ is not separable with respect to the query $P(c)$. The space $S^{\vec{0}}[KB]$ consists of a unique point $(0.3, 0.7)$, which is also the maximum-entropy point. Both disjuncts of $KB$ are consistent with the maximum-entropy point, so we might expect that the presence of the conjuncts $P(c)$ and $\neg P(c)$ in the disjuncts would not affect the degree of belief. That is, if it were possible to ignore or discount the role of the tolerances, we would expect $\mathrm{Pr}_\infty(P(c)|KB) = 0.3$. However, this is not the case. Consider the behavior of $\mathrm{Pr}_\infty^{\vec{\tau}}(P(c)|KB)$ for $\vec{\tau} > \vec{0}$. If $\tau_1 > \tau_2$, then the maximum-entropy point of $S^{\vec{\tau}}[KB]$ is $(0.3 + \tau_1, 0.7 - \tau_1)$. Now, consider some $\epsilon > 0$ sufficiently small so that $\tau_2 + \epsilon < \tau_1$. By Corollary 3.14, we deduce that $\mathrm{Pr}_\infty^{\vec{\tau}}((\|P(x)\|_x > 0.3 + \tau_2) \mid KB) = 1$. Therefore, by Theorem 3.16, $\mathrm{Pr}_\infty^{\vec{\tau}}(P(c)|KB) = \mathrm{Pr}_\infty^{\vec{\tau}}(P(c) \mid KB \wedge (\|P(x)\|_x > 0.3 + \tau_2))$ (assuming the limit exists). But since the newly added expression is inconsistent with the second disjunct, we obtain that $\mathrm{Pr}_\infty^{\vec{\tau}}(P(c)|KB) = \mathrm{Pr}_\infty^{\vec{\tau}}(P(c) \mid P(c) \wedge (\|P(x)\|_x \approx_1 0.3)) = 1$, and not $0.3$. On the other hand, if $\tau_1 < \tau_2$, we get the symmetric behavior, where $\mathrm{Pr}_\infty^{\vec{\tau}}(P(c)|KB) = 0$. Only if $\tau_1 = \tau_2$ do we get the expected value of $0.3$ for $\mathrm{Pr}_\infty^{\vec{\tau}}(P(c)|KB)$. Clearly, by appropriately choosing a sequence of tolerance vectors converging to $\vec{0}$, we can make the asymptotic value of this fraction any of $0$, $0.3$, or $1$, or not exist at all. Again, $\mathrm{Pr}_\infty(P(c)|KB)$ is not robust. ∎

We now turn our attention to restrictions on the query. In Section 4.2, we restricted to queries of the form $\varphi(c)$, where $\varphi(x)$ is essentially propositional. Although we intend to ease this restriction, we do not intend to allow queries that involve statistical information. The following example illustrates the difficulties.

**Example 4.20:** Consider the knowledge base $KB = \|P(x)\|_x \approx_1 0.3$ and the query $\varphi = \|P(x)\|_x \approx_2 0.3$. It is easy to see that the unique maximum-entropy point of $S^{\vec{\tau}}[KB]$ is $(0.3 + \tau_1, 0.7 - \tau_1)$. First suppose $\tau_2 < \tau_1$. From Corollary 3.14, it follows that $\mathrm{Pr}_\infty^{\vec{\tau}}((\|P(x)\|_x > 0.3 + \tau_2) \mid KB) = 1$. Therefore, by Theorem 3.16, $\mathrm{Pr}_\infty^{\vec{\tau}}(\varphi|KB) = \mathrm{Pr}_\infty^{\vec{\tau}}(\varphi \mid KB \wedge (\|P(x)\|_x > 0.3 + \tau_2))$ (assuming the limit exists). The latter expression is clearly $0$. On the other hand, if $\tau_1 < \tau_2$, then $KB[\vec{\tau}] \models \varphi[\vec{\tau}]$, so that $\mathrm{Pr}_\infty^{\vec{\tau}}(\varphi|KB) = 1$. Thus, the limiting behavior of $\mathrm{Pr}_\infty^{\vec{\tau}}(\varphi|KB)$ depends on how $\vec{\tau}$ goes to $\vec{0}$, so that $\mathrm{Pr}_\infty(\varphi|KB)$ is nonrobust. ∎

The real problem here is the semantics of proportion expressions in queries. While the utility of the $\approx$ connective in expressing statistical information in the knowledge base should be fairly uncontroversial, its role in *conclusions* we might draw, such as $\varphi$ in Example 4.20, is much less clear. The formal semantics we have defined requires that we consider all possible tolerances for a proportion expression in $\varphi$, so it is not surprising that nonrobustness is the usual result. One might argue that the tolerances in queries should be allowed to depend more closely on tolerances of expressions in the knowledge base. It is possible to formalize this intuition, as is done in (Koller & Halpern, 1992), to give an alternative semantics for dealing with proportion expressions in queries that often gives more reasonable behavior. Considerations of this alternative semantics would lead us too far afield here; rather, we focus for the rest of the section on first-order queries.

In fact, our goal is to allow arbitrary first-order queries, even those that involve predicates of arbitrary arity and equality (although we still need to restrict the knowledge base to the unary language $\mathcal{L}_1^{\approx}$). However, as the following example shows, quantifiers too can cause problems.

**Example 4.21:** Let $\Phi = \{P, c\}$ and consider $KB_1 = \forall x \, \neg P(x)$, $KB_2 = \|P(x)\|_x \approx_1 0$, and $\varphi = \exists x \, P(x)$. It is easy to see that $S^{\vec{0}}[KB_1] = S^{\vec{0}}[KB_2] = \{(0, 1)\}$, and therefore the unique maximum-entropy point in both is $\vec{v} = (0, 1)$. However, $\mathrm{Pr}_\infty(\varphi | KB_1)$ is clearly 0, whereas $\mathrm{Pr}_\infty(\varphi | KB_2)$ is actually 1. To see the latter fact, observe that the vast majority of models of $KB_2$ around $\vec{v}$ actually satisfy $\exists x\, P(x)$. There is actually only a single world associated with $(0, 1)$ at which $\exists x\, P(x)$ is false. This example is related to Example 4.17, because it illustrates another case in which $S^{\vec{0}}[KB]$ cannot suffice to determine degrees of belief. ∎

In the case of the knowledge base $KB_1$, the maximum-entropy point $(0, 1)$ is quite misleading about the nature of nearby worlds. We must avoid this sort of "discontinuity" when finding the degree of belief of a formula that involves first-order quantifiers. The notion of stability defined below is intended to deal with this problem. To define it, we first need the following notion of a *size description*.

**Definition 4.22:** A *size description (over $\mathcal{P}$)* is a conjunction of $K$ formulas: for each atom $A_j$ over $\mathcal{P}$, it includes exactly one of $\exists x\, A_j(x)$ and $\neg \exists x\, A_j(x)$. For $\vec{u} \in \Delta^K$, the *size description associated with $\vec{u}$*, written $\sigma(\vec{u})$, is that size description which includes $\neg \exists x\, A_i(x)$ if $u_i = 0$ and $\exists x\, A_i(x)$ if $u_i > 0$. ∎

The problems that we want to avoid occur when there is a maximum-entropy point $\vec{v}$ with size description $\sigma(\vec{v})$ such that in a neighborhood of $\vec{v}$, most of the worlds satisfying $KB$ are associated with other size descriptions. Intuitively, the problem with this is that the coordinates of $\vec{v}$ alone give us misleading information about the nature of worlds near $\vec{v}$, and so about degrees of belief.[9] We give a sufficient condition which can be used to avoid this problem in the context of our theorems. This condition is effective and uses machinery (in particular, the ability to find solution spaces) that is needed to use the maximum-entropy approach in any case.

**Definition 4.23:** Let $\vec{v}$ be a maximum-entropy point of $S^{\vec{\tau}}[KB]$. We say that $\vec{v}$ is *safe* (with respect to $KB$ and $\vec{\tau}$) if $\vec{v}$ is not contained in $S^{\vec{\tau}}[KB \wedge \neg \sigma(\vec{v})]$. We say that *$KB$ and $\vec{\tau}$ are stable for $\sigma^\star$* if for every maximum-entropy point $\vec{v} \in S^{\vec{\tau}}[KB]$ we have that $\sigma(\vec{v}) = \sigma^\star$ and that $\vec{v}$ is safe with respect to $KB$ and $\vec{\tau}$. ∎

The next result is the key property of stability that we need.

**Theorem 4.24:** *If $KB$ and $\vec{\tau} > \vec{0}$ are stable for $\sigma^\star$ then $\mathrm{Pr}_\infty^{\vec{\tau}}(\sigma^\star | KB) = 1$.*

Our theorems will use the assumption that there exists some $\sigma^\star$ such that, for all sufficiently small $\vec{\tau}$, $KB$ and $\vec{\tau}$ are stable for $\sigma^\star$. We note that this does not imply that $\sigma^\star$ is necessarily the size description associated with the maximum-entropy point(s) of $S^{\vec{0}}[KB]$.

**Example 4.25:** Consider the knowledge base $KB_2$ in Example 4.21, and recall that $\vec{v} = (0, 1)$ is the maximum-entropy point of $S^{\vec{0}}[KB_2]$. The size description $\sigma(\vec{v})$ is $\neg \exists x\, A_1(x) \wedge \exists x\, A_2(x)$. However the maximum-entropy point of $S^{\vec{\tau}}[KB_2]$ for $\vec{\tau} > 0$ is actually $(\tau_1, 1 - \tau_1)$, so that the appropriate $\sigma^\star$ for such a $\vec{\tau}$ is $\exists x\, A_1(x) \wedge \exists x\, A_2(x)$. ∎

---

9. We actually conjecture that problems of this sort cannot arise in the context of a maximum-entropy point of $S^{\vec{\tau}}[KB]$ for $\vec{\tau} > \vec{0}$. More precisely, for sufficiently small $\vec{\tau}$ and a maximum-entropy point $\vec{v}$ of $S^{\vec{\tau}}[KB]$ with $KB \in \mathcal{L}_1^\approx$, we conjecture that $\mathrm{Pr}_\infty[\mathcal{O}](\sigma(\vec{v}) | KB) = 1$ where $\mathcal{O}$ is an open set that contains $\vec{v}$ but no other maximum-entropy point of $S^{\vec{\tau}}[KB]$. If this is indeed the case, then the machinery of stability that we are about to introduce is unnecessary, since it holds in all cases that we need it. However, we have been unable to prove this.

As we now show, the restrictions outlined above and in Section 4.1 suffice for our next result on computing degrees of belief. In order to state this result, we need one additional concept. Recall that in Section 4.2 we expressed an essentially propositional formula $\varphi(x)$ as a disjunction of atoms. Since we wish to also consider formulas $\varphi$ using more than one constant and non-unary predicates, we need a richer concept than atoms. This is the motivation behind the definition of *complete descriptions*.

**Definition 4.26:** Let $\mathcal{Z}$ be some set of variables and constants. A *complete description D* over $\Phi$ and $\mathcal{Z}$ is an unquantified conjunction of formulas such that:

- For every predicate $R \in \Phi \cup \{=\}$ of arity $r$ and for every $z_{i_1}, \ldots, z_{i_r} \in \mathcal{Z}$, $D$ contains exactly one of $R(z_{i_1}, \ldots, z_{i_r})$ or $\neg R(z_{i_1}, \ldots, z_{i_r})$ as a conjunct.

- $D$ is consistent.[10] ∎

Complete descriptions simply extend the role of atoms in the context of essentially propositional formulas to the more general setting. As in the case of atoms, if we fix some arbitrary ordering of the conjuncts in a complete description, then complete descriptions are mutually exclusive and exhaustive. Clearly, a formula $\xi$ whose free variables and constants are contained in $\mathcal{Z}$, and which is is quantifier- and proportion-free, is equivalent to some disjunction of complete descriptions over $\mathcal{Z}$. For such a formula $\xi$, let $\mathcal{A}(\xi)$ be a set of complete descriptions over $\mathcal{Z}$ such that $\xi$ is equivalent to the disjunction $\bigvee_{D \in \mathcal{A}(\xi)} D$, where $\mathcal{Z}$ is the set of constants and free variables in $\xi$.

For the purposes of the remaining discussion (except within proofs), we are interested only in complete descriptions over an empty set of variables. For a set of constants $\mathcal{Z}$, we can view a description $D$ over $\mathcal{Z}$ as describing the different properties of the constants in $\mathcal{Z}$. In our construction, when considering a $KB$ of the form $\psi \wedge KB'$ which is separable with respect to a query $\varphi$, we define the set $\mathcal{Z}$ to contain precisely those constants in $\varphi$ and in $\psi$. In particular, this means that $KB'$ will mention no constant in $\mathcal{Z}$.

A complete description $D$ over a set of constants $\mathcal{Z}$ can be decomposed into three parts: the *unary part* $D^1$ which consists of those conjuncts of $D$ that involve unary predicates (and thus determines an atom for each of the constant symbols), the *equality part* $D^=$ which consists of those conjuncts of $D$ involving equality (and thus determines which of the constants are equal to each other), and the *non-unary part* $D^{>1}$ which consists of those conjuncts of $D$ involving non-unary predicates (and thus determines the non-unary properties other than equality of the constants). As we suggested, the unary part of such a complete description $D$ extends the notion of "atom" to the case of multiple constants. For this purpose, we also extend $F_{[A]}$ (for an atom $A$) and define $F_{[D]}$ for a description $D$. Intuitively, we are treating each of the individuals as independent, so that the probability that constant $c_1$ satisfies atom $A_{j_1}$ and that constant $c_2$ satisfies $A_{j_2}$ is just the product of the probability that $c_1$ satisfies $A_{j_1}$ and the probability that $c_2$ satisfies $A_{j_2}$.

**Definition 4.27:** For a complete description $D$ without variables whose unary part is equivalent to $A_{j_1}(c_1) \wedge \ldots \wedge A_{j_m}(c_m)$ (for distinct constants $c_1, \ldots, c_m$) and for a point

---

10. Inconsistency is possible because of the use of equality. For example, if $D$ includes $z_1 = z_2$ as well as both $R(z_1, z_3)$ and $\neg R(z_2, z_3)$, it is inconsistent.

$\vec{u} \in \Delta^K$, we define

$$F_{[D]}(\vec{u}) = \prod_{\ell=1}^{m} u_{j_\ell}. \quad \blacksquare$$

Note that $F_{[D]}$ is depends only on $D^1$, the unary part of $D$.

As we mentioned, we can extend our approach to deal with formulas $\varphi$ that also use non-unary predicate symbols. Our computational procedure for such formulas uses the maximum-entropy approach described above combined with the techniques of (Grove et al., 1993b). These latter were used in (Grove et al., 1993b) to compute asymptotic conditional probabilities when conditioning on a first-order knowledge base $KB_{fo}$. The basic idea in that case is as follows: To compute $\Pr_\infty(\varphi | KB_{fo})$, we examine the behavior of $\varphi$ in finite models of $KB_{fo}$. We partition the models of $KB_{fo}$ into a finite collection of classes such that $\varphi$ behaves *uniformly* in each individual class. By this we mean that almost all worlds in the class satisfy $\varphi$ or almost none do; i.e., there is a *0-1 law* for the asymptotic probability of $\varphi$ when we restrict attention to models in a single class. In order to compute $\Pr_\infty(\varphi | KB_{fo})$ we therefore identify the classes, compute the relative weight of each class (which is required because the classes are not necessarily of equal relative size), and then decide for each class whether the asymptotic probability of $\varphi$ is zero or one.

It turns out that much the same ideas continue to work in this framework. In this case, the classes are defined using complete descriptions and the appropriate size description $\sigma^*$. The main difference is that, rather than examining all worlds consistent with the knowledge base, we now concentrate on those worlds in the vicinity of the maximum-entropy points, as outlined in the previous section. It turns out that the restriction to these worlds affects very few aspects of this computational procedure. In fact, the only difference is in computing the relative weight of the different classes. This last step can be done using maximum entropy, using the tools described in Section 4.2.

**Theorem 4.28:** *Let $\varphi$ be a formula in $\mathcal{L}^{\approx}$ and let $KB = \psi \wedge KB'$ be an essentially positive knowledge base in $\mathcal{L}_1^{\approx}$ which is separable with respect to $\varphi$. Let $\mathcal{Z}$ be the set of constants appearing in $\varphi$ or in $\psi$ (so that $KB'$ contains none of the constants in $\mathcal{Z}$) and let $\chi^{\neq}$ be the formula $\bigwedge_{c,c' \in \mathcal{Z}} c \neq c'$. Assume that there exists a size description $\sigma^*$ such that, for all $\vec{\tau} > 0$, $KB$ and $\vec{\tau}$ are stable for $\sigma^*$, and that the space $S^{\vec{0}}[KB]$ has a unique maximum-entropy point $\vec{v}$. Then*

$$\Pr_\infty(\varphi | KB) = \frac{\sum_{D \in \mathcal{A}(\psi \wedge \chi^{\neq})} \Pr_\infty(\varphi | \sigma^* \wedge D) F_{[D]}(\vec{v})}{\sum_{D \in \mathcal{A}(\psi \wedge \chi^{\neq})} F_{[D]}(\vec{v})}$$

*if the denominator is positive.*

Since both $\varphi$ and $\sigma^* \wedge D$ are first-order formulas and $\sigma^* \wedge D$ is precisely of the required form in (Grove et al., 1993b), then $\Pr_\infty(\varphi | \sigma^* \wedge D)$ is either 0 or 1, and we can use the algorithm of (Grove et al., 1993b) to compute this limit, in the time bounds outlined there.

One corollary of the above is that the formula $\chi^{\neq}$ holds with probability 1 given any knowledge base $KB$ of the form we are interested in. This corresponds to a default assumption of *unique names*, a property often considered to be desirable in inductive reasoning systems.

While this theorem does represent a significant generalization of Theorem 4.11, it still has numerous restrictions. There is no question that some of these can be loosened to some extent, although we have not been able to find a clean set of conditions significantly more general than the ones that we have stated. We leave it as an open problem whether such a set of conditions exists. Of course, the most significant restriction we have made is that of allowing only unary predicates in the *KB*. This issue is the subject of the next section.

## 5. Beyond unary predicates

The random-worlds method makes complete sense for the full language $\mathcal{L}^{\approx}$ (and, indeed, for even richer languages). On the other hand, our application of maximum entropy is limited to unary knowledge bases. Is this restriction essential? While we do not have a theorem to this effect (indeed, it is not even clear what the wording of such a theorem would be), we conjecture that it is.

Certainly none of the techniques we have used in this paper can be generalized significantly. One difficulty is that, once we have a binary or higher arity predicate, we see no analogue to the notion of atoms and no canonical form theorem. In Section 3.2 and in the proof of Theorem 3.5, we discuss why it becomes impossible to get rid of nested quantifiers and proportions when we have non-unary predicates. Even considering matters on a more intuitive level, the problems seem formidable. In a unary language, atoms are useful because they are simple descriptions that summarize everything that might be known about a domain element in a model. But consider a language with a single binary predicate $R(x, y)$. Worlds over this language include all finite graphs (where we think of $R(x, y)$ as holding if there is an edge from $x$ to $y$). In this language, there are infinitely many properties that may be true or false about a domain element. For example, the assertions "the node $x$ has $m$ neighbors" are expressible in the language for each $m$. Thus, in order to partition the domain elements according to the properties they satisfy, we would need to define infinitely many partitions. Furthermore, it can be shown that "typically" (i.e., in almost all graphs of sufficiently great size) each node satisfies a different set of first-order properties. Thus, in most graphs, all the nodes are "different", so a partition of domain elements into a finite number of "atoms" makes little sense. It is very hard to see how the basic proof strategy we have used, of summarizing a model by listing the number of elements with various properties, can possibly be useful here.

The difficulty of finding an analogue to entropy in the presence of higher-arity predicates is supported by results from (Grove et al., 1993a). In this paper we have shown that maximum entropy can be a useful tool for computing degrees of belief in certain cases, if the *KB* involves only unary predicates. In (Grove et al., 1993a) we show that there can be *no* general computational technique to compute degrees of belief once we have non-unary predicate symbols in the *KB*. The problem of finding degrees of belief in this case is highly undecidable. This result was proven without statistical assertions in the language, and in fact holds for quite weak sublanguages of first-order logic. (For instance, in a language without equality and with only depth-two quantifier nesting.) So even if there is some generalized version of maximum entropy, it will either be extremely restricted in application or will be useless as a computational tool.

## 6. Conclusion

This paper has had two major thrusts. The first is to establish a connection between maximum entropy and the random-worlds approach for a significant fragment of our language, one far richer than that considered by Paris and Vencovska (1989) or Shastri (1989). The second is to suggest that such a result is unlikely to obtain for the full language.

The fact that we have a connection between maximum entropy and random worlds is significant. For one thing, it allows us to utilize all the tools that have been developed for computing maximum entropy efficiently (see (Goldman, 1987) and the further references therein), and may thus lead to efficient algorithms for computing degrees of belief for a large class of knowledge bases. In addition, maximum entropy is known to have many attractive properties (Jaynes, 1978). Our result shows these properties are shared by the random-worlds approach in the domain where these two approaches agree. Indeed, as shown in (Bacchus et al., 1994), the random-worlds approach has many of these properties for the full (non-unary) language.

On the other hand, a number of properties of maximum entropy, such as its dependence on the choice of language and its inability to handle causal reasoning appropriately, have been severely criticized (Pearl, 1988; Goldszmidt et al., 1990). Not surprisingly, these criticisms apply to random worlds as well. A discussion of these criticisms, and whether they really should be viewed as shortcomings of the random-worlds method, is beyond the scope of this paper; the interested reader should consult (Bacchus et al., 1994, Section 7) for a more thorough discussion of these issues and additional references.

We believe that our observations regarding the limits of the connection between the random-worlds method and maximum entropy are also significant. The question of how widely maximum entropy applies is quite important. Maximum entropy has been gaining prominence as a means of dealing with uncertainty both in AI and other areas. However, the difficulties of using the method once we move to non-unary predicates seem not to have been fully appreciated. In retrospect, this is not that hard to explain; in almost all applications where maximum entropy has been used (and where its application can be best justified in terms of the random-worlds method) the knowledge base is described in terms of unary predicates (or, equivalently, unary functions with a finite range). For example, in physics applications we are interested in such predicates as quantum state (see (Denbigh & Denbigh, 1985)). Similarly, AI applications and expert systems typically use only unary predicates such as symptoms and diseases (Cheeseman, 1983). We suspect that this is not an accident, and that deep problems will arise in more general cases. This poses a challenge to proponents of maximum entropy since, even if one accepts the maximum-entropy principle, the discussion above suggests that it may simply be inapplicable in a large class of interesting examples.

## Appendix A. Proofs for Section 3.2

**Theorem 3.5:** *Every formula in $\mathcal{L}_1^=$ is equivalent to a formula in canonical form. Moreover, there is an effective procedure that, given a formula $\xi \in \mathcal{L}_1^=$ constructs an equivalent formula $\hat{\xi}$ in canonical form.*

**Proof:** We show how to effectively transform $\xi \in \mathcal{L}_1^=$ to an equivalent formula in canonical form. We first rename variables if necessary, so that all variables used in $\xi$ are distinct (i.e., no two quantifiers, including proportion expressions, ever bind the same variable symbol).

We next transform $\xi$ into an equivalent *flat* formula $\xi_f \in \mathcal{L}_1^{\approx}$, where a flat formula is one where no quantifiers (including proportion quantifiers) have within their scope a constant or variable other than the variable(s) the quantifier itself binds. (Note that in this transformation we do not require that $\xi$ be closed. Also, observe that flatness implies that there are no nested quantifiers.)

We define the transformation by induction on the structure of $\xi$. There are three easy steps:

- If $\xi$ is an unquantified formulas, then $\xi_f = \xi$.

- $(\xi' \vee \xi'')_f = \xi'_f \vee \xi''_f$

- $(\neg \xi')_f = \neg(\xi_f)$.

All that remains is to consider quantified formulas of the form $\exists x \, \xi'$, $\|\xi'\|_{\vec{x}}$, or $\|\xi'|\xi''\|_{\vec{x}}$. It turns out that the same transformation works in all three cases. We illustrate the transformation by looking at the case where $\xi$ is of the form $\|\xi'\|_{\vec{x}}$. By the inductive hypothesis, we can assume that $\xi'$ is flat. For the purposes of this proof, we define a *basic* formula to be an atomic formula (i.e., one of the form $P(z)$), a proportion formula, or a quantified formula (i.e., one of the form $\exists x \, \chi$). Let $\chi_1, \ldots, \chi_k$ be all basic subformulas of $\xi'$ that do not mention any variable in $\vec{x}$. Let $z$ be a variable or constant symbol not in $\vec{x}$ that is mentioned in $\xi'$. Clearly $z$ must occur in some basic subformula of $\xi'$, say $\chi'$. By the inductive hypothesis, it is easy to see that $\chi'$ cannot mention any variable in $\vec{x}$ and so, by construction, it is in $\{\chi_1, \ldots, \chi_\ell\}$. In other words, not only do $\{\chi_1, \ldots, \chi_\ell\}$ not mention any variable in $\vec{x}$, but they also contain all occurrences of the other variables and constants. (Notice that this argument fails if the language contains any high-arity predicates, including equality. For then $\xi'$ might include subformulas of the form $R(x, y)$ or $x = y$, which can mix variables outside $\vec{x}$ with those in $\vec{x}$.)

Now, let $B_1, \ldots, B_{2^\ell}$ be all the "atoms" over $\chi_1, \ldots, \chi_\ell$. That is, we consider all formulas $\chi'_1 \wedge \ldots \wedge \chi'_\ell$ where $\chi'_i$ is either $\chi_i$ or $\neg \chi_i$. Now consider the disjunction:

$$\bigvee_{i=1}^{2^\ell} (B_i \wedge \|\xi'\|_{\vec{x}}).$$

This is surely equivalent to $\|\xi'\|_{\vec{x}}$, because some $B_i$ must be true. However, if we assume that a particular $B_i$ is true, we can simplify $\|\xi'\|_{\vec{x}}$ by replacing all the $\chi_i$ subformulas by *true* or *false*, according to $B_i$. (Note that this is allowed only because the $\chi_i$ do not mention any variable in $\vec{x}$.) The result is that we can simplify each disjunct $(B_i \wedge \|\xi'\|_{\vec{x}})$ considerably. In fact, because of our previous observation about $\{\chi_1, \ldots, \chi_\ell\}$, there will be no constants or variables outside $\vec{x}$ left within the proportion quantifier. This completes this step of the induction. Since the other quantifiers can be treated similarly, this proves the flatness result.

69

It now remains to show how a flat formula can be transformed to canonical form. Suppose $\xi \in \mathcal{L}_1^\approx$ is flat. Let $\xi^\star \in \mathcal{L}_1^=$ be the formula equivalent to $\xi$ obtained by using the translation of Section 2.1. Every proportion comparison in $\xi^\star$ is of the form $t \leq t'\varepsilon_i$ where $t$ and $t'$ are polynomials over flat unconditional proportions. In fact, $t'$ is simply a product of flat unconditional proportions (where the empty product is taken to be 1). Note also that since we cleared away conditional proportions by multiplying by $t'$, if $t' = 0$ then so is $t$, and so the formula $t \leq t'\varepsilon_i$ is automatically true. We can therefore replace the comparison by $(t' = 0) \vee (t \leq t'\varepsilon_i \wedge t' > 0)$. Similarly, we can replace a negated comparison by an expression of the form $\neg (t \leq t'\varepsilon_i) \wedge t' > 0$.

The next step is to rewrite all the flat unconditional proportions in terms of atomic proportions. In any such proportion $||\xi'||_{\vec{x}}$, the formula $\xi'$ is a Boolean combination of $P(x_i)$ for predicates $P \in \mathcal{P}$ and $x_i \in \vec{x}$. Thus, the formula $\xi'$ is equivalent to a disjunction $\bigvee_j (A_1^j(x_{i_1}) \wedge \ldots \wedge A_m^j(x_{i_m}))$, where each $A_i^j$ is an atom over $\mathcal{P}$ and $\vec{x} = \{x_{i_1}, \ldots, x_{i_m}\}$. These disjuncts are mutually exclusive and the semantics treats distinct variables as being independent, so

$$||\xi'||_{\vec{x}} = \sum_j \prod_{i=1}^m ||A_i^j(x)||_x.$$

We perform this replacement for each proportion expression. Furthermore, any term $t'$ in an expression of the form $t \leq t'\varepsilon_i$ will be a product of such expressions, and so will be positive.

Next, we must put all pure first-order formulas in the right form. We first rewrite $\xi$ to push all negations inwards as far as possible, so that only atomic subformulas and existential formulas are negated. Next, note that since $\xi$ is flat, each existential subformula must have the form $\exists x\, \xi'$, where $\xi'$ is a quantifier-free formula which mentions no constants and only the variable $x$. Hence, $\xi'$ is a Boolean combination of $P(x)$ for predicates $P \in \mathcal{P}$. Again, the formula $\xi'$ is equivalent to a disjunction of atoms of the form $\bigvee_{A \in \mathcal{A}(\xi)} A(x)$, so $\exists x\, \xi'$ is equivalent to $\bigvee_{A \in \mathcal{A}(\xi)} \exists x\, A(x)$. We replace $\exists x\, \xi'$ by this expression. Finally, we must deal with formulas of the form $P(c)$ or $\neg P(c)$ for $P \in \mathcal{P}$. This is easy: We can again replace a formula $\xi$ of the form $P(c)$ or $\neg P(c)$ by the disjunction $\bigvee_{A \in \mathcal{A}(\xi)} A(c)$.

The penultimate step is to convert $\xi$ into disjunctive normal form. This essentially brings things into canonical form. Note that since we dealt with formulas of the form $\neg P(c)$ in the previous step, we do not have to deal with conjuncts of the form $\neg A_i(c)$.

The final step is to check that we do not have $A_i(c)$ and either $\neg \exists x\, A_i(x)$ or $A_j(c)$ for some $j \neq i$ as conjuncts of some disjunct. If we do, we simply remove that disjunct. ∎

## Appendix B. Proofs for Section 3.3

**Lemma 3.11:** *There exist some function $h : \mathbb{N} \to \mathbb{N}$ and two strictly positive polynomial functions $f, g : \mathbb{N} \to \mathbb{R}$ such that, for $KB \in \mathcal{L}_1^\approx$ and $\vec{u} \in \Delta^K$, if $\#worlds_N^{\vec{r}}[\vec{u}](KB) \neq 0$, then*

$$(h(N)/f(N))e^{NH(\vec{u})} \leq \#worlds_N^{\vec{r}}[\vec{u}](KB) \leq h(N)g(N)e^{NH(\vec{u})}.$$

**Proof:** To choose a world $W \in \mathcal{W}_N$ satisfying $KB$ such that $\pi(W) = \vec{u}$, we must partition the domain among the atoms according to the proportions in $\vec{u}$, and then choose an assignment for the constants in the language subject to the constraints imposed by $KB$. Finally, even though $KB$ mentions only unary predicates, if there are any non-unary predicates in the vocabulary we must choose a denotation for them.

Suppose $\vec{u} = (u_1, \ldots, u_K)$, and let $N_i = u_i N$ for $i = 1, \ldots, K$. The number of partitions of the domain into atoms is $\binom{N}{N_1, \ldots, N_K}$; each such partition completely determines the denotation for the unary predicates. We must also specify the denotations of the constant symbols. There are at most $N^{|\mathcal{C}|}$ ways of choosing these. On the other hand, we know there is at least one model $(W, \vec{\tau})$ of $KB$ such that $\pi(W) = \vec{u}$, so there there at least one choice. In fact, there is at least one world $W' \in \mathcal{W}_N$ such that $(W', \vec{\tau}) \models KB$ for each of the $\binom{N}{N_1, \ldots, N_K}$ ways of partitioning the elements of the domain (and each such world $W'$ is isomorphic to $W$). Finally we must choose the denotation of the non-unary predicates. However, $\vec{u}$ does not constrain this choice and, by assumption, neither does $KB$. Therefore the number of such choices is some function $h(N)$ which is independent of $\vec{u}$.[11] We conclude that:

$$h(N)\binom{N}{N_1, \ldots, N_K} \leq \#worlds_N^{\vec{\tau}}[\vec{u}](KB) \leq h(N)N^{|\mathcal{C}|}\binom{N}{N_1, \ldots, N_K}.$$

It remains to estimate

$$\binom{N}{N_1, \ldots, N_K} = \frac{N!}{N_1! N_2! \ldots N_K!}.$$

To obtain our result, we use Stirling's approximation for the factorials, which says that

$$m! = \sqrt{2\pi m}\, m^m e^{-m}(1 + O(1/m)).$$

It follows that exist constants $L, U > 0$ such that

$$L\, m^m e^{-m} \leq m! \leq U m\, m^m e^{-m}$$

for all $m$. Using these bounds, as well as the fact that $N_i \leq N$, we get:

$$\frac{L}{U^K N^K} \frac{N^N \prod_{i=1}^{K} e^{N_i}}{e^N \prod_{i=1}^{K} N_i^{N_i}} \leq \frac{N!}{N_1! N_2! \ldots N_K!} \leq \frac{UN}{L^K} \frac{N^N \prod_{i=1}^{K} e^{N_i}}{e^N \prod_{i=1}^{K} N_i^{N_i}}.$$

Now, consider the expression common to both bounds:

$$
\begin{aligned}
\frac{N^N \prod_{i=1}^{K} e^{N_i}}{e^N \prod_{i=1}^{K} N_i^{N_i}} &= \frac{N^N}{\prod_{i=1}^{K} N_i^{N_i}} \\
&= \prod_{i=1}^{K} \left(\frac{N}{N_i}\right)^{N_i} \\
&= \prod_{i=1}^{K} e^{N_i \ln(N/N_i)} \\
&= e^{-N \sum_{i=1}^{K} u_i \ln(u_i)} = e^{NH(\vec{u})}.
\end{aligned}
$$

---

11. It is easy to verify that in fact

$$h(N) = \prod_{R \in \Phi - \Psi} 2^{N^{arity(R)}},$$

where $\Psi$ is the unary fragment of $\Phi$ and $arity(R)$ denotes the arity of the predicate symbol $R$.

We obtain that

$$\frac{h(N)L}{U^K N^K} e^{NH(\vec{u})} \leq \# worlds_N^{\vec{\tau}}[\vec{u}](KB) \leq N^{|\mathcal{C}|} h(N) \frac{UN}{L^K} e^{NH(\vec{u})},$$

which is the desired result. ∎

We next want to prove Theorem 3.13. To do this, it is useful to have an alternative representation of the solution space $S^{\vec{\tau}}[KB]$. Towards this end, we have the following definition.

**Definition B.1:** Let $\Pi_N^{\vec{\tau}}[KB] = \{\pi(W) : W \in \mathcal{W}_N; (W, \vec{\tau}) \models KB\}$. Let $\Pi_\infty^{\vec{\tau}}[KB]$ be the limit of these spaces. Formally,

$$\Pi_\infty^{\vec{\tau}}[KB] = \{\vec{u} \ : \ \exists N_0 \ s.t. \ \forall N \geq N_0 \ \exists \vec{u}^N \in \Pi_N^{\vec{\tau}}[KB] \ s.t. \ \lim_{N \to \infty} \vec{u}^N = \vec{u}\}. \quad \blacksquare$$

The following theorem establishes a tight connection between $S^{\vec{\tau}}[KB]$ and $\Pi_\infty^{\vec{\tau}}[KB]$.

**Theorem B.2:**

(a) For all $N$ and $\vec{\tau}$, we have $\Pi_N^{\vec{\tau}}[KB] \subseteq S^{\vec{\tau}}[KB]$.

(b) For all sufficiently small $\vec{\tau}$, we have $\Pi_\infty^{\vec{\tau}}[KB] = S^{\vec{\tau}}[KB]$.

**Proof:** Part (a) is immediate: If $\vec{u} \in \Pi_N^{\vec{\tau}}[KB]$, then $\vec{u} = \pi(W)$ for some $W \in \mathcal{W}_N$ such that $(W, \vec{\tau}) \models KB$. It is almost immediate from the definitions that $\pi(W)$ must satisfy $\Gamma(KB[\vec{\tau}])$, so $\pi(W) \in Sol[\Gamma(KB[\vec{\tau}])]$. The inclusion $\Pi_N^{\vec{\tau}}[KB] \subseteq S^{\vec{\tau}}[KB]$ now follows.

One direction of part (b) follows immediately from part (a). Recall that $\Pi_N^{\vec{\tau}}[KB] \subseteq S^{\vec{\tau}}[KB]$ and that the points in $\Pi_\infty^{\vec{\tau}}[KB]$ are limits of a sequence of points in $\Pi_N^{\vec{\tau}}[KB]$. Since $S^{\vec{\tau}}[KB]$ is closed, it follows that $\Pi_\infty^{\vec{\tau}}[KB] \subseteq S^{\vec{\tau}}[KB]$.

For the opposite inclusion, the general strategy of the proof is to show the following:

(i) If $\vec{\tau}$ is sufficiently small, then for all $\vec{u} \in S^{\vec{\tau}}[KB]$, there is some sequence of points $\left\{ \vec{u}^{N_0}, \vec{u}^{N_0+1}, \vec{u}^{N_0+2}, \vec{u}^{N_0+3}, \ldots \right\} \subset Sol[\Gamma(KB[\vec{\tau}])]$ such that, for all $N \geq N_0$, the coordinates of $\vec{u}^N$ are all integer multiples of $1/N$ and $\lim_{N \to \infty} \vec{u}^N = \vec{u}$.

(ii) if $\vec{w} \in Sol[\Gamma(KB[\vec{\tau}])]$ and all its coordinates are integer multiples of $1/N$, then $\vec{w} \in \Pi_N^{\vec{\tau}}[KB]$.

This clearly suffices to prove that $\vec{u} \in \Pi_\infty^{\vec{\tau}}[KB]$.

We begin with the proof of (ii), which is straightforward. Suppose the point $\vec{w} = (r_1/N, r_2/N, \ldots, r_K/N)$ is in $Sol[\Gamma(KB[\vec{\tau}])]$. We construct a world $W \in \mathcal{W}_N$ such that $\pi(W) = \vec{w}$ as follows. The denotation of atom $A_1$ is the set of elements $\{1, \ldots, r_1\}$, the denotation of atom $A_2$ is the set $\{r_1 + 1, \ldots, r_1 + r_2\}$, and so on. It remains to choose the denotations of the constants (since the denotation of the predicates of arity greater than 1 is irrelevant). Without loss of generality we can assume $KB$ is in canonical form. (If not, we consider $\widehat{KB}$.) Thus, $KB$ is a disjunction of conjunctions, say $\bigvee_j \xi_j$. Since $\vec{w} \in Sol[\Gamma(KB[\vec{\tau}])]$, we must have $\vec{w} \in Sol[\Gamma(\xi_j[\vec{\tau}])]$ for some $j$. We use $\xi_j$ to define the properties of the constants. If $\xi_j$ contains $A_i(c)$ for some atom $A_i$, then we make $c$ satisfy $A_i$. Note that, by Definition 3.6, if $\xi_j$ has such a conjunct then $u_i > 0$. If $\xi_j$ contains no atomic conjunct mentioning the constant $c$, then we make $c$ satisfy $A_i$ for some arbitrary atom with $u_i > 0$. It should now be clear that $(W, \vec{\tau})$ satisfies $\xi_j$, and so satisfies $KB$. Note that in this construction it is important that we started with $\vec{w}$ in $Sol[\Gamma(KB[\vec{\tau}])]$, rather than just in the closure space $S^{\vec{\tau}}[KB]$; otherwise, the point would not necessarily satisfy $\Gamma(KB[\vec{\tau}])$.

We now consider condition (i). This is surprisingly difficult to prove; the proof involves techniques from algebraic geometry. Our job would be relatively easy if $Sol[\Gamma(KB[\vec{\tau}])]$ were an open set. Unfortunately, it is not. On the other hand, it would behave essentially like an open set if we could replace the occurrences of $\leq$ in $\Gamma(KB[\vec{\tau}])$ by $<$. It turns out that, for our purposes here, this replacement is possible.

Let $\Gamma^<(KB[\vec{\tau}])$ be the same as $\Gamma(KB[\vec{\tau}])$ except that every (unnegated) conjunct of the form $(t \leq \tau_i t')$ is replaced by $(t < \tau_i t')$. (Notice that this is essentially the opposite transformation to the one used when defining essential positivity in Definition 4.4.) Finally, let $S^{<\vec{\tau}}[KB]$ be $\overline{Sol[\Gamma^<(KB[\vec{\tau}])]}$. It turns out that, for all sufficiently small $\vec{\tau}$, $S^{<\vec{\tau}}[KB] = S^{\vec{\tau}}[KB]$. This result, which we label as Lemma B.5, will be stated and proved later. For now we use the lemma to continue the proof of the main result.

Consider some $\vec{u} \in S^{\vec{\tau}}[KB]$. It suffices to show that for all $\delta > 0$ there exists $N_0$ such that for all $N > N_0$, there exists a point $\vec{u}^N \in Sol[\Gamma^<(KB[\vec{\tau}])]$ such that all the coordinates of $\vec{u}^N$ are integer multiples of $1/N$ and such that $|\vec{u} - \vec{u}^N| < \delta$. (For then we can take smaller and smaller $\delta$'s to create a sequence $\vec{u}^N$ converging to $\vec{u}$.) Hence, let $\delta > 0$. By Lemma B.5, we can find some $\vec{u}' \in Sol[\Gamma^<(KB[\vec{\tau}])]$ such that $|\vec{u} - \vec{u}'| < \delta/2$. By definition, every conjunct in $\Gamma^<(KB[\vec{\tau}])$ is of the form $q'(\vec{w}) = 0$, $q'(\vec{w}) > 0$, $q(\vec{w}) < \tau_i q'(\vec{w})$, or $q(\vec{w}) > \tau_i q'(\vec{w})$, where $q'$ is a positive polynomial. Ignore for the moment the constraints of the form $q'(\vec{w}) = 0$, and consider the remaining constraints that $\vec{u}'$ satisfies. These constraints all involve strict inequalities, and the functions involved ($q$ and $q'$) are continuous. Thus, there exists some $\delta' > 0$ such that for all $\vec{w}$ for which $|\vec{u}' - \vec{w}| < \delta'$, these constraints are also satisfied by $\vec{w}$. Now consider a conjunct of the form $q'(\vec{w}) = 0$ that is satisfied by $\vec{u}'$. Since $q'$ is positive, this happens if and only if the following condition holds: for every coordinate $w_i$ that actually appears in $q'$, we have $u_i' = 0$. In particular, if $\vec{w}$ and $\vec{u}'$ have the same coordinates with value 0, then $q'(\vec{w}) = 0$. It follows that for all $\vec{w}$, if $|\vec{u}' - \vec{w}| < \delta'$ and $\vec{u}'$ and $\vec{w}$ have the same coordinates with value 0, then $\vec{w}$ also satisfies $\Gamma^<(KB[\vec{\tau}])$.

We now construct $\vec{u}^N$ that satisfies the requirements. Let $i^\star$ be the index of that component of $\vec{u}'$ with the largest value. We define $\vec{u}^N$ by considering each of its components $u_i^N$, for $1 \leq i \leq K$:

$$u_i^N = \begin{cases} 0 & u_i' = 0 \\ \lceil N u_i' \rceil / N & i \neq i^\star \text{ and } u_i' > 0 \\ u_i^N - \sum_{j \neq i^\star}(u_j^N - u_j') & i = i^\star. \end{cases}$$

It is easy to verify that the components of $\vec{u}^N$ sum to 1. All the components in $\vec{u}'$, other than the $i^\star$'th, are increased by at most $1/N$. The component $u_{i^\star}^N$ is decreased by at most $K/N$. We will show that $\vec{u}^N$ has the right properties for all $N > N_0$, where $N_0$ is such that $1/N_0 < \min(u_{i^\star}, \delta/2, \delta')/2K$. The fact that $K/N_0 < u_{i^\star}$ guarantees that $\vec{u}^N$ is in $\Delta^K$ for all $N > N_0$. The fact that $2K/N_0 < \delta/2$ guarantees that $\vec{u}^N$ is within $\delta/2$ of $\vec{u}'$, and hence within $\delta$ of $\vec{u}$. Since $2K/N_0 < \delta'$, it follows that $|\vec{u}' - \vec{u}^N| < \delta'$. Since $\vec{u}^N$ is constructed to have exactly the same 0 coordinates as $\vec{u}'$, we conclude that $\vec{u}^N \in Sol[\Gamma^<(KB[\vec{\tau}])]$, as required. Condition (i), and hence the entire theorem, now follows. ∎

It now remains to prove Lemma B.5, which was used in the proof just given. As we hinted earlier, this requires tools from algebraic geometry. We base our definitions on the presentation in (Bochnak, Coste, & Roy, 1987). A subset $A$ of $\mathbb{R}^\ell$ is said to be *semi-algebraic* if it is definable in the language of real-closed fields. That is, $A$ is semi-algebraic if there is a first-order formula $\varphi(x_1, \ldots, x_\ell)$ whose free variables are $x_1, \ldots, x_\ell$ and whose only non-logical symbols are $0, 1, +, \times, <$ and $=$, such that $\mathbb{R} \models \varphi(u_1, \ldots, u_\ell)$ iff $(u_1, \ldots, u_\ell) \in A$.[12] A function $f : X \to Y$, where $X \subseteq \mathbb{R}^h$ and $Y \subseteq \mathbb{R}^\ell$, is said to be semi-algebraic if its graph (i.e., $\{(\vec{u}, \vec{w}) : f(\vec{u}) = \vec{w}\}$) is semi-algebraic. The main tool we use is the following *Curve Selection Lemma* (see (Bochnak et al., 1987, p. 34)):

**Lemma B.3:** *Suppose that $A$ is a semi-algebraic set in $\mathbb{R}^\ell$ and $\vec{u} \in \overline{A}$. Then there exists a continuous, semi-algebraic function $f : [0,1] \to \mathbb{R}^\ell$ such that $f(0) = \vec{u}$ and $f(t) \in A$ for all $t \in (0,1]$.*

Our first use of the Curve Selection Lemma is in the following, which says that, in a certain sense, semi-algebraic functions behave "nicely" near limits. The type of phenomenon we wish to avoid is illustrated by $x \sin \frac{1}{x}$ which is continuous at 0, but has infinitely many local maxima and minima near 0.

**Proposition B.4:** *Suppose that $g : [0,1] \to \mathbb{R}$ is a continuous, semi-algebraic function such that $g(u) > 0$ if $u > 0$ and $g(0) = 0$. Then there exists some $\epsilon > 0$ such that $g$ is strictly increasing in the interval $[0, \epsilon]$.*

**Proof:** Suppose, by way of contradiction, that $g$ satisfies the hypotheses of the proposition but there is no $\epsilon$ such that $g$ is increasing in the interval $[0, \epsilon]$. We define a point $u$ in $[0,1]$ to be *bad* if for some $u' \in [0, u)$ we have $g(u') \geq g(u)$. Let $A$ be the set of all the bad points. Since $g$ is semi-algebraic so is $A$, since $u' \in A$ iff

$$\exists u' ((0 \leq u' < u) \wedge (g(u) \leq g(u'))).$$

Since, by assumption, $g$ is not increasing in any interval $[0, \epsilon]$, we can find bad points arbitrarily close to 0 and so $0 \in \overline{A}$. By the Curve Selection Lemma, there is a continuous semi-algebraic curve $f : [0,1] \to \mathbb{R}$ such that $f(0) = 0$ and $f(t) \in A$ for all $t \in (0,1]$. Because of the continuity of $f$, the range of $f$, i.e., $f([0,1])$, is $[0, r]$ for some $r \in [0,1]$. By the definition of $f$, $(0, r] \subseteq A$. Since $0 \notin A$, it follows that $f(1) \neq 0$; therefore $r > 0$ and so, by assumption, $g(r) > 0$. Since $g$ is a continuous function, it achieves a maximum $v > 0$ over the range $[0, r]$. Consider the minimum point in the interval where this maximum is achieved. More precisely, let $u$ be the infimum of the set $\{u' \in [0, r] : g(u') = v\}$; clearly, $g(u) = v$. Since $v > 0$ we obtain that $u > 0$ and therefore $u \in A$. Thus, $u$ is bad. But that means that there is a point $u' < u$ for which $g(u') \geq g(u)$, which contradicts the choice of $v$ and $u$. ∎

We can now prove Lemma B.5. Recall, the result we need is as follows.

---

12. In (Bochnak et al., 1987), a set is taken to be semi-algebraic if it is definable by a quantifier-free formula in the language of real closed fields. However, as observed in (Bochnak et al., 1987), since the theory of real closed fields admits elimination of quantifiers (Tarski, 1951), the two definitions are equivalent.

**Lemma B.5:** *For all sufficiently small $\vec{\tau}$, $S^{<\vec{\tau}}[KB] = S^{\vec{\tau}}[KB]$.*

**Proof:** Clearly $S^{<\vec{\tau}}[KB] \subseteq S^{\vec{\tau}}[KB]$. To prove the reverse inclusion we consider $\widehat{KB}$, a canonical form equivalent of $KB$. We consider each disjunct of $\widehat{KB}$ separately. Let $\xi$ be a conjunction that is one of the disjuncts in $\widehat{KB}$. It clearly suffices to show that $Sol[\Gamma(\xi[\vec{\tau}])] \subseteq S^{<\vec{\tau}}[\xi] = \overline{Sol[\Gamma^{<}(\xi[\vec{\tau}])]}$. Assume, by way of contradiction, that for arbitrarily small $\vec{\tau}$, there exists some $\vec{u} \in Sol[\Gamma(\xi[\vec{\tau}])]$ which is "separated" from the set $Sol[\Gamma^{<}(\xi[\vec{\tau}])]$, i.e., is not in its closure. More formally, we say that $\vec{u}$ is $\delta$-*separated* from $Sol[\Gamma^{<}(\xi[\vec{\tau}])]$ if there is no $\vec{u}' \in Sol[\Gamma^{<}(\xi[\vec{\tau}])]$ such that $|\vec{u} - \vec{u}'| < \delta$.

We now consider those $\vec{\tau}$ and those points in $Sol[\Gamma(\xi[\vec{\tau}])]$ that are separated from $Sol[\Gamma^{<}(\xi[\vec{\tau}])]$:[13]

$$A = \{(\vec{\tau}, \vec{u}, \delta) \; : \; \vec{\tau} > \vec{0}, \; \delta > 0, \; \vec{u} \in Sol[\Gamma(\xi[\vec{\tau}])] \text{ is } \delta\text{-separated from } Sol[\Gamma^{<}(\xi[\vec{\tau}])]\}.$$

Clearly $A$ is semi-algebraic. By assumption, there are points in $A$ for arbitrarily small tolerance vectors $\vec{\tau}$. Since $A$ is a bounded subset of $\mathbb{R}^{m+K+1}$ (where $m$ is the number of tolerance values in $\vec{\tau}$), we can use the Bolzano–Weierstrass Theorem to conclude that this set of points has an accumulation point whose first component is $\vec{0}$. Thus, there is a point $(\vec{0}, \vec{w}, \delta')$ in $\overline{A}$. By the Curve Selection Lemma, there is a continuous semi-algebraic function $f : [0, 1] \to \mathbb{R}^{m+K+1}$ such that $f(0) = (\vec{0}, \vec{w}, \delta')$ and $f(t) \in A$ for $t \in (0, 1]$.

Since $f$ is semi-algebraic, it is semi-algebraic in each of its coordinates. By Lemma B.4, there is some $v > 0$ such that $f$ is strictly increasing in each of its first $m$ coordinates over the domain $[0, v]$. Suppose that $f(v) = (\vec{\tau}, \vec{u}, \delta)$. Now, consider the constraints in $\Gamma(\xi[\vec{\tau}])$ that have the form $q(\vec{w}) > \tau_j q'(\vec{w})$. These constraints are all satisfied by $\vec{u}$ and they all involve strong inequalities. By the continuity of the polynomials $q$ and $q'$, there exists some $\epsilon > 0$ such that, for all $\vec{u}'$ such that $|\vec{u} - \vec{u}'| < \epsilon$, $\vec{u}'$ also satisfies these constraints.

Now, by the continuity of $f$, there exists a point $v' \in (0, v)$ sufficiently close to $v$ such that if $f(v') = (\vec{\tau}', \vec{u}', \delta')$, then $|\vec{u} - \vec{u}'| < \min(\delta, \epsilon)$. Since $f(v) = (\vec{\tau}, \vec{u}, \delta) \in A$ and $|\vec{u} - \vec{u}'| < \delta$, it follows that $\vec{u}' \notin Sol[\Gamma^{<}(\xi[\vec{\tau}])]$. We conclude the proof by showing that this is impossible. That is, we show that $\vec{u}' \in Sol[\Gamma^{<}(\xi[\vec{\tau}])]$. The constraints appearing in $\Gamma^{<}(\xi[\vec{\tau}])$ can be of the following forms: $q'(\vec{w}) = 0$, $q'(\vec{w}) > 0$, $q(\vec{w}) < \tau_j q'(\vec{w})$, or $q(\vec{w}) > \tau_j q'(\vec{w})$, where $q'$ is a positive polynomial. Since $f(v') \in A$, we know that $\vec{u}' \in Sol[\Gamma(\xi[\vec{\tau}'])]$. The constraints of the form $q'(\vec{w}) = 0$ and $q'(\vec{w}) > 0$ are identical in $\Gamma(\xi[\vec{\tau}'])$ and in $\Gamma^{<}(\xi[\vec{\tau}])$, and are therefore satisfied by $\vec{u}'$. Since $|\vec{u}' - \vec{u}| < \epsilon$, our discussion in the previous paragraph implies that the constraints of the form $q(\vec{w}) > \tau_j q'(\vec{w})$ are also satisfied by $\vec{u}'$. Finally, consider a constraint of the form $q(\vec{w}) < \tau_j q'(\vec{w})$. The corresponding constraint in $\Gamma(\xi[\vec{\tau}'])$ is $q(\vec{w}) \leq \tau'_j q'(\vec{w})$. Since $\vec{u}'$ satisfies this latter constraint, we know that $q(\vec{u}') \leq \tau'_j q'(\vec{u}')$. But now, recall that we proved that $f$ is increasing over $[0, v]$ in the first $m$ coordinates. In particular, $\tau'_j < \tau_j$. By the definition of canonical form, $q'(\vec{u}') > 0$, so that we conclude $q(\vec{u}') \leq \tau'_j q'(\vec{u}') < \tau_j q'(\vec{u}')$. Hence the constraints of this type are also satisfied by $\vec{u}'$. This concludes the proof that $\vec{u}' \in Sol[\Gamma^{<}(KB[\vec{\tau}])]$, thus deriving a contradiction and proving the result. ∎

We are finally ready to prove Theorem 3.13.

---

13. We consider only those components in the infinite vector $\vec{\tau}$ that actually appear in $Sol[\Gamma(\xi[\vec{\tau}])]$.

**Theorem 3.13:** *For all sufficiently small $\vec{\tau}$, the following is true. Let $\mathcal{Q}$ be the points with greatest entropy in $S^{\vec{\tau}}[KB]$ and let $\mathcal{O} \subseteq \mathbb{R}^K$ be any open set containing $\mathcal{Q}$. Then for all $\theta \in \mathcal{L}^\approx$ and for $\lim^* \in \{\limsup, \liminf\}$ we have*

$$\lim_{N \to \infty}{}^* \Pr_N^{\vec{\tau}}(\theta | KB) = \lim_{N \to \infty}{}^* \frac{\# worlds_N^{\vec{\tau}}[\mathcal{O}](\theta \wedge KB)}{\# worlds_N^{\vec{\tau}}[\mathcal{O}](KB)}.$$

**Proof:** Let $\vec{\tau}$ be small enough so that Theorem B.2 applies and let $\mathcal{Q}$ and $\mathcal{O}$ be as in the statement of the theorem. It clearly suffices to show that the set $\mathcal{O}$ contains almost all of the worlds that satisfy $KB$. More precisely, the fraction of such worlds that are in $\mathcal{O}$ tends to 1 as $N \to \infty$.

Let $\rho$ be the entropy of the points in $\mathcal{Q}$. We begin the proof by showing the existence of $\rho_L < \rho_U \ (< \rho)$ such that (for sufficiently large $N$) (a) every point $\vec{u} \in \Pi_N^{\vec{\tau}}[KB]$ where $\vec{u} \notin \mathcal{O}$ has entropy at most $\rho_L$ and (b) there is at least one point $\vec{u} \in \Pi_N^{\vec{\tau}}[KB]$ with $\vec{u} \in \mathcal{O}$ and entropy at least $\rho_U$.

For part (a), consider the space $S^{\vec{\tau}}[KB] - \mathcal{O}$. Since this space is closed, the entropy function takes on a maximum value in this space; let this be $\rho_L$. Since this space does not include any point with entropy $\rho$ (these are all in $\mathcal{Q} \subseteq \mathcal{O}$), we must have $\rho_L < \rho$. By Theorem B.2, $\Pi_N^{\vec{\tau}}[KB] \subseteq S^{\vec{\tau}}[KB]$. Therefore, for any $N$, the entropy of any point in $\Pi_N^{\vec{\tau}}[KB] - \mathcal{O}$ is at most $\rho_L$.

For part (b), let $\rho_U$ be some value in the interval $(\rho_L, \rho)$ (for example $(\rho_L + \rho)/2$) and let $\vec{v}$ be any point in $\mathcal{Q}$. By the continuity of the entropy function, there exists some $\delta > 0$ such that for all $\vec{u}$ with $|\vec{u} - \vec{v}| < \delta$, we have $H(\vec{u}) \geq \rho_U$. Because $\mathcal{O}$ is open we can, by considering a smaller $\delta$ if necessary, assume that $|\vec{u} - \vec{v}| < \delta$ implies $\vec{u} \in \mathcal{O}$. By the second part of Theorem B.2, there is a sequence of points $\vec{u}^N \in \Pi_N^{\vec{\tau}}[KB]$ such that $\lim_{N \to \infty} \vec{u}^N = \vec{v}$. In particular, for $N$ large enough we have $|\vec{u}^N - \vec{v}| < \delta$, so that $H(\vec{u}^N) > \rho_U$, proving part (b).

To complete the proof, we use Lemma 3.11 to conclude that for all $N$,

$$\# worlds_N^{\vec{\tau}}(KB) \geq \# worlds_N^{\vec{\tau}}[\vec{u}^N](KB) \geq (h(N)/f(N))e^{NH(\vec{u}^N)} \geq (h(N)/f(N))e^{N\rho_U}.$$

On the other hand,

$$\begin{aligned} \# worlds_N^{\vec{\tau}}[\Delta^K - \mathcal{O}](KB) &\leq \sum_{\vec{u} \in \Pi_N^{\vec{\tau}}[KB] - \mathcal{O}} \# worlds_N^{\vec{\tau}}[\vec{u}](KB) \\ &\leq |\{\vec{w} \in \Pi_N^{\vec{\tau}}[KB] : \vec{w} \notin \mathcal{O}\}| \, h(N)g(N)e^{N\rho_L} \\ &\leq (N+1)^K h(N)g(N)e^{N\rho_L}. \end{aligned}$$

Therefore the fraction of models of $KB$ which are outside $\mathcal{O}$ is at most

$$\frac{(N+1)^K h(N)f(N)g(N)e^{N\rho_L}}{h(N)e^{N\rho_U}} = \frac{(N+1)^K f(N)g(N)}{e^{N(\rho_U - \rho_L)}}.$$

Since $(N+1)^k f(N)g(N)$ is a polynomial in $N$, this fraction tends to 0 as $N$ grows large. The result follows. ∎

## Appendix C. Proofs for Section 4

**Proposition 4.6:** *Assume that $KB$ is essentially positive and let $\mathcal{Q}$ be the set of maximum-entropy points of $S^{\vec{0}}[KB]$ (and thus also of $S^{\leq \vec{0}}[KB]$). Then for all $\epsilon > 0$ and all sufficiently small tolerance vectors $\vec{\tau}$ (where "sufficiently small" may depend on $\epsilon$), every maximum-entropy point of $S^{\vec{\tau}}[KB]$ is within $\epsilon$ of some maximum entropy-point in $\mathcal{Q}$.*

**Proof:** Fix $\epsilon > 0$. By way of contradiction, assume that that there is some sequence of tolerance vectors $\vec{\tau}^m$, $m = 1, 2, \ldots$, that converges to $\vec{0}$, and for each $m$ a maximum-entropy point $\vec{u}^m$ of $S^{\vec{\tau}^m}[KB]$ such that for all $m$, $\vec{u}^m$ is at least $\epsilon$ away from $\mathcal{Q}$. Since the space $\Delta^K$ is compact, we can assume without loss of generality that this sequence converges to some point $\vec{u}$. Recall that $\Gamma(KB)$ is a finite combination (using "and" and "or") of constraints, where every such constraint is of the form $q'(\vec{w}) = 0$, $q'(\vec{w}) > 0$, $q(\vec{w}) \leq \varepsilon_j q'(\vec{w})$, or $q(\vec{w}) > \varepsilon_j q'(\vec{w})$, such that $q'$ is a positive polynomial. Since the overall number of constraints is finite we can assume, again without loss of generality, that all the $\vec{u}^m$'s satisfy precisely the same constraints. We claim that the corresponding conjuncts in $\Gamma^{\leq}(KB[\vec{0}])$ are satisfied by $\vec{u}$. For a conjunct of the form $q'(\vec{w}) = 0$ note that, if $q'(\vec{u}^m) = 0$ for all $m$, then this also holds at the limit, so that $q(\vec{u}) = 0$. A conjunct of the form $q'(\vec{w}) > 0$ translates into $q'(\vec{w}) \geq 0$ in $\Gamma^{\leq}(KB[\vec{0}])$; such conjuncts are trivially satisfied by any point in $\Delta^K$. If a conjunct of the form $q(\vec{w}) \leq \varepsilon_j q'(\vec{w})$ is satisfied for all $\vec{u}^m$ and $\vec{\tau}^m$, then at the limit we have $q(\vec{u}) \leq 0$, which is precisely the corresponding conjunct in $\Gamma^{\leq}(KB[\vec{0}])$. Finally, for a conjunct of the form $q(\vec{w}) > \varepsilon_j q'(\vec{w})$, if $q(\vec{u}^m) > \tau_j^m q'(\vec{u}^m)$ for all $m$, then at the limit we have $q(\vec{u}) \geq 0$, which again is the corresponding conjunct in $\Gamma^{\leq}(KB[\vec{0}])$. It follows that $\vec{u}$ is in $S^{\leq \vec{0}}[KB]$.

By assumption, all points $\vec{u}^m$ are at least $\epsilon$ away from $\mathcal{Q}$. Hence, $\vec{u}$ cannot be in $\mathcal{Q}$. If we let $\rho$ represent the entropy of the points in $\mathcal{Q}$, since $\mathcal{Q}$ is the set of all maximum-entropy points in $S^{\leq \vec{0}}[KB]$, it follows that $H(\vec{u}) < \rho$. Choose $\rho_L$ and $\rho_U$ such that $H(\vec{u}) < \rho_L < \rho_U < \rho$. Since the entropy function is continuous, we know that for sufficiently large $m$, $H(\vec{u}^m) \leq \rho_L$. Since $\vec{u}^m$ is a maximum-entropy point of $S^{\vec{\tau}^m}[KB]$, it follows that the entropy achieved in this space for sufficiently large $m$ is at most $\rho_L$. We derive a contradiction by showing that for sufficiently large $m$, there is some point in $Sol[\Gamma(KB[\vec{\tau}^m])]$ with entropy at least $\rho_U$. The argument is as follows. Let $\vec{v}$ be some point in $\mathcal{Q}$. Since $\vec{v}$ is a maximum-entropy point of $S^{\vec{0}}[KB]$, there are points in $Sol[\Gamma(KB[\vec{0}])]$ arbitrarily close to $\vec{v}$. In particular, there is some point $\vec{u}' \in Sol[\Gamma(KB[\vec{0}])]$ whose entropy is at least $\rho_U$. As we now show, this point is also in $Sol[\Gamma(KB[\vec{\tau}])]$ for all sufficiently small $\vec{\tau}$. Again, consider all the conjuncts in $\Gamma(KB[\vec{0}])$ satisfied by $\vec{u}'$ and the corresponding conjuncts in $\Gamma(KB[\vec{\tau}])$. Conjuncts of the form $q'(\vec{w}) = 0$ and $q'(\vec{w}) > 0$ in $\Gamma(KB[\vec{0}])$ remain unchanged in $\Gamma(KB[\vec{\tau}])$. Conjuncts of the form $q(\vec{w}) \leq \tau_j q'(\vec{w})$ in $\Gamma(KB[\vec{\tau}])$ are certainly satisfied by $\vec{u}'$, since the corresponding conjunct in $\Gamma(KB[\vec{0}])$, namely $q(\vec{w}) \leq 0$, is satisfied by $\vec{u}'$, so that $q(\vec{u}') \leq 0 \leq \tau_j q'(\vec{u}')$ (recall that $q'$ is a positive polynomial). Finally, consider a conjunct in $\Gamma(KB[\vec{\tau}])$ of the form $q(\vec{w}) > \tau_j q'(\vec{w})$. The corresponding conjunct in $\Gamma(KB[\vec{0}])$ is $q(\vec{w}) > 0$. Suppose $q(\vec{u}') = \delta > 0$. Since the value of $q'$ is bounded over the compact space $\Delta^K$, it follows that for all sufficiently small $\tau_j$, $\tau_j q'(\vec{u}') < \delta$. Thus, $q(\vec{u}') > \tau_j q'(\vec{u}')$ for all sufficiently small $\tau_j$, as required. It follows that $\vec{u}'$ is in $Sol[\Gamma(KB[\vec{\tau}])]$ for all sufficiently small $\vec{\tau}$ and, in particular, in $Sol[\Gamma(KB[\vec{\tau}^m])]$ for all sufficiently large $m$. But $H(\vec{u}') \geq \rho_U$, whereas we showed that the maximum entropy achieved in $S^{\vec{\tau}^m}[KB]$ is at most $\rho_L < \rho_U$.

This contradiction proves that our assumption was false, so that the conclusion of the proposition necessarily holds. ∎

**Theorem 4.9:** *Suppose $\varphi(c)$ is a simple query for KB. For all $\vec{\tau}$ sufficiently small, if $\mathcal{Q}$ is the set of maximum-entropy points in $S^{\vec{\tau}}[KB]$ and $F_{[\psi]}(\vec{v}) > 0$ for all $\vec{v} \in \mathcal{Q}$, then for $\lim^{\star} \in \{\limsup, \liminf\}$ we have*

$$\lim_{N \to \infty}{}^{\star} \Pr_N^{\vec{\tau}}(\varphi(c)|KB) \in \left[ \inf_{\vec{v} \in \mathcal{Q}} F_{[\varphi|\psi]}(\vec{v}), \sup_{\vec{v} \in \mathcal{Q}} F_{[\varphi|\psi]}(\vec{v}) \right].$$

**Proof:** Let $W \in \mathcal{W}^{\star}$, and let $\vec{u} = \pi(W)$. The value of the proportion expression $||\psi(x)||_x$ at $W$ is clearly

$$\sum_{A_j \in \mathcal{A}(\psi)} ||A_j(x)||_x = \sum_{A_j \in \mathcal{A}(\psi)} u_j = F_{[\psi]}(\vec{u}).$$

If $F_{[\psi]}(\vec{u}) > 0$, then by the same reasoning we conclude that the value of $||\varphi(x)|\psi(x)||_x$ at $W$ is equal to $F_{[\varphi|\psi]}(\vec{u})$.

Now, let $\lambda_L$ and $\lambda_R$ be $\inf_{\vec{v} \in \mathcal{Q}} F_{[\varphi|\psi]}(\vec{v})$ and $\sup_{\vec{v} \in \mathcal{Q}} F_{[\varphi|\psi]}(\vec{v})$ respectively; by our assumption, $F_{[\varphi|\psi]}(\vec{v})$ is well-defined for all $\vec{v} \in \mathcal{Q}$. Since the denominator is not 0, $F_{[\varphi|\psi]}$ is a continuous function at each maximum-entropy point. Thus, since $F_{[\varphi|\psi]}(\vec{v}) \in [\lambda_L, \lambda_R]$ for all maximum-entropy points, the value of $F_{[\varphi|\psi]}(\vec{u})$ for $\vec{u}$ "close" to some $\vec{v} \in \mathcal{Q}$, will either be in the range $[\lambda_L, \lambda_U]$ or very close to it. More precisely, choose any $\epsilon > 0$, and define $\theta[\epsilon]$ to be the formula

$$||\varphi(x)|\psi(x)||_x \in [\lambda_L - \epsilon, \lambda_U + \epsilon].$$

Since $\epsilon > 0$, it is clear that there is some sufficiently small open set $\mathcal{O}$ around $\mathcal{Q}$ such that this proportion expression is well-defined and within these bounds at all worlds in $\mathcal{O}$. Thus, by Corollary 3.14, $\Pr_\infty^{\vec{\tau}}(\theta[\epsilon]||KB) = 1$. Using Theorem 3.16, we obtain that for $\lim^{\star}$ as above,

$$\lim_{N \to \infty}{}^{\star} \Pr_N^{\vec{\tau}}(\varphi(c)|KB) = \lim_{N \to \infty}{}^{\star} \Pr_N^{\vec{\tau}}(\varphi(c)|KB \wedge \theta[\epsilon]).$$

But now we can use the direct inference technique outlined earlier. We are interested in the probability of $\varphi(c)$, where the only information we have about $c$ in the knowledge base is $\psi(c)$ and where we have statistics for $||\varphi(x)|\psi(x)||_x$. These are precisely the conditions under which Theorem 4.1 applies. We conclude that

$$\lim_{N \to \infty}{}^{\star} \Pr_N^{\vec{\tau}}(\varphi(c)|KB) \in [\lambda_L - \epsilon, \lambda_U + \epsilon].$$

Since this holds for all $\epsilon > 0$, it is necessarily the case that

$$\lim_{N \to \infty}{}^{\star} \Pr_N^{\vec{\tau}}(\varphi(c)|KB) \in [\lambda_L, \lambda_U],$$

as required. ∎

**Theorem 4.11:** *Suppose $\varphi(c)$ is a simple query for KB. If the space $S^{\vec{0}}[KB]$ has a unique maximum-entropy point $\vec{v}$, KB is essentially positive, and $F_{[\psi]}(\vec{v}) > 0$, then*

$$\Pr_\infty(\varphi(c)|KB) = F_{[\varphi|\psi]}(\vec{v}).$$

**Proof:** Note that the fact that $S^{\vec{0}}[KB]$ has a unique maximum-entropy point does not guarantee that this is also the case for $S^{\vec{\tau}}[KB]$. However, Proposition 4.6 implies that the maximum-entropy points of the latter space are necessarily close to $\vec{v}$. More precisely, if we choose some $\epsilon > 0$, we conclude that for all sufficiently small $\vec{\tau}$, all the maximum-entropy points of $S^{\vec{\tau}}[KB]$ will be within $\epsilon$ of $\vec{v}$. Now, pick some arbitrary $\delta > 0$. Since $F_{[\psi]}(\vec{v}) > 0$, it follows that $F_{[\varphi|\psi]}$ is continuous at $\vec{v}$. Therefore, there exists some $\epsilon > 0$ such that if $\vec{u}$ is within $\epsilon$ of $\vec{v}$, $F_{[\varphi|\psi]}(\vec{u})$ is within $\delta$ of $F_{[\varphi|\psi]}(\vec{v})$. In particular, this is the case for all maximum-entropy points of $S^{\vec{\tau}}[KB]$ for all sufficiently small $\vec{\tau}$. This allows us to apply Theorem 4.9 and conclude that for all sufficiently small $\vec{\tau}$ and for $\lim^* \in \{\limsup, \liminf\}$, $\lim^*_{N \to \infty} \Pr^{\vec{\tau}}_N(\varphi(c)|KB)$ is within $\delta$ of $F_{[\varphi|\psi]}(\vec{v})$. Hence, this is also the case for $\lim_{\vec{\tau} \to \vec{0}} \lim^*_{N \to \infty} \Pr^{\vec{\tau}}_N(\varphi(c)|KB)$. Since this holds for all $\delta > 0$, it follows that

$$\lim_{\vec{\tau} \to \vec{0}} \liminf_{N \to \infty} \Pr^{\vec{\tau}}_N(\varphi(c)|KB) = \lim_{\vec{\tau} \to \vec{0}} \limsup_{N \to \infty} \Pr^{\vec{\tau}}_N(\varphi(c)|KB) = F_{[\varphi|\psi]}(\vec{v}).$$

Thus, by definition, $\Pr_\infty(\varphi(c)|KB) = F_{[\varphi|\psi]}(\vec{v})$. ∎

**Theorem 4.14:** *Let $\Lambda$ be a conjunction of constraints of the form $\Pr(\beta|\beta') = \lambda$ or $\Pr(\beta|\beta') \in [\lambda_1, \lambda_2]$. There is a unique probability distribution $\mu^*$ of maximum entropy satisfying $\Lambda$. Moreover, for all $\beta$ and $\beta'$, if $\Pr_{\mu^*}(\beta') > 0$, then*

$$\Pr_\infty(\xi_\beta(c)|\xi_{\beta'}(c) \wedge KB'[\Lambda]) = \Pr_{\mu^*}(\beta|\beta').$$

**Proof:** Clearly, the formulas $\varphi(x) = \xi_\beta(x)$ and $\psi(x) = \xi_{\beta'}(x)$ are essentially propositional. The knowledge base $KB'[\Lambda]$ is in the form of a conjunction of simple proportion formulas, none of which are negated. As a result, the set of constraints associated with $KB = \psi(c) \wedge KB'[\Lambda]$ also has a simple form. $KB'[\Lambda]$ generates a conjunction of constraints which can be taken as having the form $q(\vec{w}) \le \varepsilon_j q'(\vec{w})$. On the other hand, $\psi(c)$ generates some Boolean combination of constraints all of which have the form $w_j > 0$. We begin by considering the set $S^{\le \vec{0}}[KB]$ (rather than $S^{\vec{0}}[KB]$), so we can ignore the latter constraints for now.

$S^{\le \vec{0}}[KB]$ is defined by a conjunction of linear constraints which (as discussed earlier) implies that it is convex, and thus has a unique maximum-entropy point, say $\vec{v}$. Let $\mu^* = \mu_{\vec{v}}$ be the distribution over $\Omega$ corresponding to $\vec{v}$. It is clear that the constraints of $\Gamma^\le(KB[\vec{0}])$ on the points of $\Delta^K$ are precisely the same ones as those of $\Lambda$. Therefore, $\mu^*$ is the unique maximum-entropy distribution satisfying the constraints of $\Lambda$. By Remark 4.13, it follows that $F_{[\xi_{\beta'}]}(\vec{v}) = \mu^*(\beta')$. Since we have assumed that $\mu^*(\beta') > 0$, we are are almost in a position to use Theorem 4.11. It remains to prove essential positivity.

Recall that the difference between $\Gamma^\le(KB[\vec{0}])$ and $\Gamma(KB[\vec{0}])$ is that the latter may have some conjuncts of the form $w_j > 0$. Checking definitions 3.4 and 3.6 we see that such terms can appear only due to $\xi_{\beta'}(c)$ and, in fact, together they assert that $F_{[\xi_{\beta'}]}(\vec{w}) > 0$. But we have assumed that $F_{[\xi_{\beta'}]}(\vec{v}) > 0$ and so $\vec{v}$ is a maximum-entropy point of $S^{\vec{0}}[KB]$ as well. Thus, essential positivity holds and so, by Theorem 4.11,

$$\Pr_\infty(\varphi(c)|\psi(c) \wedge KB'[\Lambda]) = F_{[\varphi|\psi]}(\mu^*) = \Pr_{\mu^*}(\beta|\beta')$$

as required. ∎

**Theorem 4.15:** *Let $c$ be a constant symbol. Using the translation described in Section 4.3, for a set $\mathcal{R}$ of defeasible rules, $B \rightarrow C$ is an ME-plausible consequence of $\mathcal{R}$ iff*

$$\mathrm{Pr}_\infty\left(\xi_C(c)\,\middle|\,\xi_B(c) \wedge \bigwedge_{r \in \mathcal{R}} \theta_r\right) = 1.$$

**Proof:** Let $KB'$ denote $\bigwedge_{r \in \mathcal{R}} \theta_r$. For all sufficiently small $\vec{\tau}$ and for $\epsilon = \tau_1$, let $\mu^*$ denote $\mu^*_{\epsilon,\mathcal{R}}$. It clearly suffices to prove that

$$\mathrm{Pr}^{\vec{\tau}}_\infty(\xi_C(c)|\xi_B(c) \wedge KB') = \mathrm{Pr}_{\mu^*}(C|B),$$

where by equality we also mean that one side is defined iff the other is also defined. It is easy to verify that a point $\vec{u}$ in $\Delta^K$ satisfies $\Gamma(KB'[\vec{\tau}])$ iff the corresponding distribution $\mu$ $\epsilon$-satisfies $\mathcal{R}$. Therefore, the maximum-entropy point $\vec{v}$ of $S^{\vec{\tau}}[KB']$ (which is unique, by linearity) corresponds precisely to $\mu^*$. Now, there are two cases: either $\mu^*(B) > 0$ or $\mu^*(B) = 0$. In the first case, by Remark 4.13, $\mathrm{Pr}_{\mu^*}(\xi_B(c)) = F_{[\xi_B(c)]}(\vec{v})$, so the latter is also positive. This also implies that $\vec{v}$ is consistent with the constraints $\Gamma(\psi(c))$ entailed by $\psi(c) = \xi_B(c)$, so that $\vec{v}$ is also the unique maximum-entropy point of $S^{\vec{\tau}}[KB]$ (where $KB = \xi_B(c) \wedge KB'$). We can therefore use Corollary 4.10 and Remark 4.13 to conclude that $\mathrm{Pr}^{\vec{\tau}}_\infty(\xi_C(c)|KB) = F_{[\xi_C(c)|\xi_B(c)]}(\vec{v}) = \mathrm{Pr}_{\mu^*}(C|B)$ and that all three terms are well-defined. Assume, on the other hand, that $\mu^*(B) = 0$, so that $\mathrm{Pr}_{\mu^*}(C|B)$ is not well-defined. In this case, we can use a known result (see (Paris & Vencovska, 1989)) for the maximum-entropy point over a space defined by linear constraints, and conclude that for all $\mu$ satisfying $\mathcal{R}$, necessarily $\mu(B) = 0$. Using the connection between distributions $\mu$ satisfying $\mathcal{R}$ and points $\vec{u}$ in $S^{\vec{\tau}}[KB']$, we conclude that this is also the case for all $\vec{u} \in S^{\vec{\tau}}[KB']$. By part (a) of Theorem B.2, this means that in any world satisfying $KB'$, the proportion $||\xi_B(x)||_x$ is necessarily 0. Thus, $KB'$ is inconsistent with $\xi_B(c)$, and $\mathrm{Pr}^{\vec{\tau}}_\infty(\xi_C(c)|\xi_B(c) \wedge KB')$ is also not well-defined. ■

## Appendix D. Proofs for Section 4.4

**Theorem 4.24:** *If $KB$ and $\vec{\tau} > \vec{0}$ are stable for $\sigma^*$ then $\mathrm{Pr}^{\vec{\tau}}_\infty(\sigma^*|KB) = 1$.*

**Proof:** By Theorem 3.14, it suffices to show that there is some open neighborhood containing $\mathcal{Q}$, the maximum-entropy points of $S^{\vec{\tau}}[KB]$, such that every world $W$ of $KB$ in this neighborhood has $\sigma(W) = \sigma^*$. So suppose this is not the case. Then there is some sequence of worlds $W_1, W_2, \ldots$ such that $(W_i, \vec{\tau}) \models KB \wedge \neg\sigma^*$ and $\lim_{i \to \infty} \min_{\vec{v} \in \mathcal{Q}} |\pi(W_i) - \vec{v}| = 0$. Since $\Delta^K$ is compact the sequence $\pi(W_1), \pi(W_2), \ldots$ must have at least one accumulation point, say $\vec{u}$. This point must be in the closure of the set $\mathcal{Q}$. But, in fact, $\mathcal{Q}$ is a closed set (because entropy is a continuous function) and so $\vec{u} \in \mathcal{Q}$. By part (a) of Theorem B.2, $\pi(W_i) \in S^{\vec{\tau}}[KB \wedge \neg\sigma^*]$ for every $i$ and so, since this space is closed, $\vec{u} \in S^{\vec{\tau}}[KB \wedge \neg\sigma^*]$ as well. But this means that $\vec{u}$ is an unsafe maximum-entropy point, contrary to the definition and assumption of stability. ■

In the remainder of this section we prove Theorem 4.28. For this purpose, fix $KB = \psi \wedge KB'$, $\varphi$, and $\sigma^*$ to be as in the statement of this theorem, and let $\vec{v}$ be the unique maximum-entropy point of $S^{\vec{0}}[KB]$.

Let $\mathcal{Z} = \{c_1, \ldots, c_m\}$ be the set of constant symbols appearing in $\psi$ and in $\varphi$. Due to the separability assumption, $KB'$ contains none of the constant symbols in $\mathcal{Z}$. Let $\chi^{\neq}$ be the formula $\bigwedge_{i \neq j} c_i \neq c_j$. We first prove that $\chi^{\neq}$ has probability 1 given $KB'$.

**Lemma D.1:** *For $\chi^{\neq}$ and $KB'$ as above, $\mathrm{Pr}_{\infty}(\chi^{\neq} | KB') = 1$.*

**Proof:** We actually show that $\mathrm{Pr}_{\infty}(\neg\chi^{\neq} | KB') = 0$. Let $c$ and $c'$ be two constant symbols in $\{c_1, \ldots, c_m\}$ and consider $\mathrm{Pr}_{\infty}(c = c' | KB')$. We again use the direct inference technique. Note that for any world of size $N$ the proportion expression $||x = x'||_{x,x'}$ denotes exactly $1/N$. It is thus easy to see that $\mathrm{Pr}_{\infty}(||x = x'||_{x,x'} \approx_i 0 | KB') = 1$ (for any choice of $i$). Thus, by Theorem 3.16, $\mathrm{Pr}_{\infty}(c = c' | KB') = \mathrm{Pr}_{\infty}(c = c' | KB' \wedge ||x = x'||_{x,x'} \approx_i 0)$. But since $c$ and $c'$ appear nowhere in $KB'$ we can use Theorem 4.1 to conclude that $\mathrm{Pr}_{\infty}(c = c' | KB') = 0$. It is straightforward to verify that, since $\neg\chi^{\neq}$ is equivalent to a finite disjunction, each disjunct of which implies $c = c'$ for at least one pair of constants $c$ and $c'$, we must have $\mathrm{Pr}_{\infty}(\neg\chi^{\neq} | KB') = 0$. ∎

As we stated in Section 4.4, our general technique for computing the probability of an arbitrary formula $\varphi$ is to partition the worlds into a finite collection of classes such that $\varphi$ behaves uniformly over each class and then to compute the relative weights of the classes. As we show later, the classes are essentially defined using complete descriptions. Their relative weight corresponds to the probabilities of the different complete descriptions given $KB$.

**Proposition D.2:** *Let $KB = KB' \wedge \psi$ and $\vec{v}$ be as above. Assume that $\mathrm{Pr}_{\infty}(\psi | KB') > 0$. Let $D$ be a complete description over $\mathcal{Z}$ that is consistent with $\psi$.*

(a) *If $D$ is inconsistent with $\chi^{\neq}$, then $\mathrm{Pr}_{\infty}(D | KB) = 0$.*

(b) *If $D$ is consistent with $\chi^{\neq}$, then*

$$\mathrm{Pr}_{\infty}(D | KB) = \frac{F_{[D]}(\vec{v})}{\sum_{D' \in \mathcal{A}(\psi \wedge \chi^{\neq})} F_{[D']}(\vec{v})}.$$

**Proof:** First, observe that if all limits exist and the denominator is nonzero, then

$$\mathrm{Pr}_{\infty}(\neg\chi^{\neq} | \psi \wedge KB') = \frac{\mathrm{Pr}_{\infty}(\neg\chi^{\neq} \wedge \psi | KB')}{\mathrm{Pr}_{\infty}(\psi | KB')}.$$

By hypothesis, the denominator is indeed nonzero. Furthermore, by Lemma D.1, $\mathrm{Pr}_{\infty}(\neg\chi^{\neq} \wedge \psi | KB') \leq \mathrm{Pr}_{\infty}(\neg\chi^{\neq} | KB') = 0$. Hence $\mathrm{Pr}_{\infty}(\chi^{\neq} | KB) = \mathrm{Pr}_{\infty}(\chi^{\neq} | KB' \wedge \psi) = 1$. We can therefore use Theorem 3.16 to conclude that

$$\mathrm{Pr}_{\infty}(D | KB) = \mathrm{Pr}_{\infty}(D | KB \wedge \chi^{\neq}).$$

Part (a) of the proposition follows immediately.

To prove part (b), recall that $\psi$ is equivalent to the disjunction $\bigvee_{E \in \mathcal{A}(\psi)} E$. By simple probabilistic reasoning, the assumption that $\mathrm{Pr}_{\infty}(\psi | KB') > 0$, and part (a), we conclude that

$$\mathrm{Pr}_{\infty}(D | \psi \wedge KB') = \frac{\mathrm{Pr}_{\infty}(D \wedge \psi | KB')}{\mathrm{Pr}_{\infty}(\psi | KB')} = \frac{\mathrm{Pr}_{\infty}(D \wedge \psi | KB')}{\sum_{E \in \mathcal{A}(\psi \wedge \chi^{\neq})} \mathrm{Pr}_{\infty}(E | KB')}.$$

By assumption, $D$ is consistent with $\chi^{\neq}$ and is in $\mathcal{A}(\psi)$. Since $D$ is a complete description, we must have that $D \Rightarrow \psi$ is valid. Thus, the numerator on the right-hand side of this equation is simply $\Pr_\infty(D|KB')$. Hence, the problem of computing $\Pr_\infty(D|KB)$ reduces to a series of computations of the form $\Pr_\infty(E|KB')$ for various complete descriptions $E$.

Fix any such description $E$. Recall that $E$ can be decomposed into three parts: the unary part $E^1$, the non-unary part $E^{>1}$, and the equality part $E^=$. Since $E$ is in $\mathcal{A}(\chi^{\neq})$, we conclude that $\chi^{\neq}$ is equivalent to $E^=$. Using Theorem 3.16 twice and some probabilistic reasoning, we get:

$$\begin{aligned}
\Pr_\infty(E^{>1} \wedge E^1 \wedge E^=|KB') &= \Pr_\infty(E^{>1} \wedge E^1 \wedge E^=|KB' \wedge \chi^{\neq}) \\
&= \Pr_\infty(E^{>1} \wedge E^1|KB' \wedge \chi^{\neq}) \\
&= \Pr_\infty(E^{>1}|KB' \wedge \chi^{\neq} \wedge E^1) \cdot \Pr_\infty(E^1|KB' \wedge \chi^{\neq}) \\
&= \Pr_\infty(E^{>1}|KB' \wedge \chi^{\neq} \wedge E^1) \cdot \Pr_\infty(E^1|KB').
\end{aligned}$$

In order to simplify the first expression, recall that none of the predicate symbols in $E^{>1}$ occur anywhere in $KB' \wedge \chi^{\neq} \wedge E^1$. Therefore, the probability of $E^{>1}$ given $KB' \wedge \chi^{\neq}$ is equal to the probability that the elements denoting the $|\mathcal{Z}|$ (different) constants satisfy some particular configuration of non-unary properties. It should be clear that, by symmetry, all such configurations are equally likely. Therefore, the probability of any one of them is a constant, equal to 1 over the total number of configurations.[14] Let $\rho$ denote the constant which is equal to $\Pr_\infty(E^{>1}|KB' \wedge \chi^{\neq} \wedge E^1)$ for all $E$.

The last step is to show that, if $E^1$ is equivalent to $\bigwedge_{j=1}^m A_{i_j}(c_j)$, then $\Pr_\infty(E^1|KB') = F_{[D]}(\vec{v})$:

$$\begin{aligned}
\Pr_\infty(\bigwedge_{j=1}^m A_{i_j}(c_j)|KB') &= \Pr_\infty(A_{i_1}(c_1)|\bigwedge_{j=2}^m A_{i_j}(c_j) \wedge KB') \cdot \Pr_\infty(A_{i_2}(c_2)|\bigwedge_{j=3}^m A_{i_j}(c_j) \wedge KB') \\
&\quad \cdot \ldots \cdot \Pr_\infty(A_{i_{m-1}}(c_{m-1})|A_{i_m}(c_m) \wedge KB') \cdot \Pr_\infty(A_{i_m}(c_m)|KB') \\
&= v_{i_1} \cdot \ldots \cdot v_{i_m} \text{ (using Theorem 4.11; see below)} \\
&= F_{[D]}(\vec{v}).
\end{aligned}$$

The first step is simply probabilistic reasoning. The second step uses $m$ applications of Theorem 4.11. It is easy to see that $A_{i_j}(c_j)$ is a simple query for $A_{i_{j+1}}(c_{j+1}) \wedge \ldots \wedge A_{i_m}(c_m) \wedge KB'$. We would like to show that

$$\Pr_\infty(A_{i_j}(c_j)|\bigwedge_{\ell=j+1}^m A_{i_\ell}(c_\ell) \wedge KB') = \Pr_\infty(A_{i_j}(c_j)|KB') = v_{i_j},$$

where Theorem 4.11 justifies the last equality. To prove the first equality, we show that for all $j$, the spaces $S^{\vec{0}}[KB']$ and $S^{\vec{0}}[\bigwedge_{\ell=j+1}^m A_{i_\ell}(c_j) \wedge KB']$ have the same maximum-entropy point, namely $\vec{v}$. This is proved by backwards induction; the $j = m$ case is trivially true. The difference between the $(j-1)$st and $j$th case is the added conjunct $A_{i_j}(c_j)$, which amounts to adding the new constraint $w_{i_j} > 0$. There are two possibilities. First, if $v_{i_j} > 0$, then $\vec{v}$ satisfies this new constraint anyway and so remains the maximum-entropy point, completing this step of the induction. If $v_{i_j} = 0$ this is not the case, and indeed, the property we are trying to prove can be false (for $j < m$). But this does not matter, because we then know that $\Pr_\infty(A_{i_j}(c_j) \mid \bigwedge_{\ell=j+1}^m A_{i_\ell}(c_\ell) \wedge KB') = \Pr_\infty(A_{i_j}(c_j) \mid KB') = v_{i_j} = 0$. Since both of the products in question include a 0 factor, it is irrelevant as to whether the other terms agree.

---

14. Although we do not need the value of this constant in our calculations below, it is in fact easy to verify that its value is $\prod_{R \in (\mathbf{\Psi} - \mathbf{\Psi})} 2^{m^{arity(R)}}$, where $m = |\mathcal{Z}|$.

We can now put everything together to conclude that

$$\Pr_\infty(D \mid KB) = \frac{\Pr_\infty(D \mid KB')}{\sum_{E \in \mathcal{A}(\psi \wedge \chi^{\neq})} \Pr_\infty(E \mid KB')} = \frac{F_{[D]}(\vec{v})}{\sum_{E \in \mathcal{A}(\psi \wedge \chi^{\neq})} F_{[E]}(\vec{v})},$$

proving part (b). ∎

We now address the issue of computing $\Pr_\infty(\varphi \mid KB)$ for an arbitrary formula $\varphi$. To do that, we must first investigate the behavior of $\Pr_\infty^{\vec{\tau}}(\varphi \mid KB)$ for small $\vec{\tau}$. Fix some sufficiently small $\vec{\tau} > 0$, and let $\mathcal{Q}$ be the set of maximum-entropy points of $S^{\vec{\tau}}[KB]$. Assume $KB$ and $\vec{\tau}$ are stable for $\sigma^\star$. By definition, this means that for every $\vec{v} \in \mathcal{Q}$, we have $\sigma(\vec{v}) = \sigma^\star$. Let $I$ be the set of $i$'s for which $\sigma^\star$ contains the conjunct $\exists x A_i(x)$. Since $\sigma(\vec{v}) = \sigma^\star$ for all $\vec{v}$, we must have that $v_i > 0$ for all $i \in I$. Since $\mathcal{Q}$ is a closed set, this implies that there exists some $\epsilon > 0$ such that for all $\vec{v} \in \mathcal{Q}$ and for all $i \in I$, we have $v_i > \epsilon$. Let $\theta[\epsilon]$ be the formula

$$\bigwedge_{i \in I} \|A_i(x)\|_x > \epsilon.$$

The following proposition is now easy to prove:

**Proposition D.3:** *Suppose that $KB$ and $\vec{\tau}$ are stable for $\sigma^\star$ and that $\mathcal{Q}$, $i$, $\theta[\epsilon]$, and $\chi^{\neq}$ are as above. Then*

$$\Pr_\infty^{\vec{\tau}}(\varphi \mid KB) = \sum_{D \in \mathcal{A}(\psi)} \Pr_\infty^{\vec{\tau}}(\varphi \mid KB' \wedge \theta[\epsilon] \wedge \sigma^\star \wedge D) \cdot \Pr_\infty^{\vec{\tau}}(D \mid KB).$$

**Proof:** Clearly, $\theta[\epsilon]$ satisfies the conditions of Corollary 3.14, allowing us to conclude that $\Pr_\infty^{\vec{\tau}}(\theta[\epsilon] \mid KB) = 1$. Similarly, by Theorem 4.24 and the assumptions of Theorem 4.28, we can conclude that $\Pr_\infty^{\vec{\tau}}(\sigma^\star \mid KB) = 1$. Since the conjunction of two assertions that have probability 1 also has probability 1, we can use Theorem 3.16 to conclude that $\Pr_\infty^{\vec{\tau}}(\varphi \mid KB) = \Pr_\infty^{\vec{\tau}}(\varphi \mid KB \wedge \theta[\epsilon] \wedge \sigma^\star)$.

Now, recall that $\psi$ is equivalent to the disjunction $\bigvee_{D \in \mathcal{A}(\psi)} D$. By straightforward probabilistic reasoning, we can therefore conclude that

$$\Pr_\infty^{\vec{\tau}}(\varphi \mid KB \wedge \theta[\epsilon] \wedge \sigma^\star) = \sum_{D \in \mathcal{A}(\psi)} \Pr_\infty^{\vec{\tau}}(\varphi \mid KB \wedge \theta[\epsilon] \wedge \sigma^\star \wedge D) \cdot \Pr_\infty^{\vec{\tau}}(D \mid KB \wedge \theta[\epsilon] \wedge \sigma^\star).$$

By Theorem 3.16 again, $\Pr_\infty^{\vec{\tau}}(D \mid KB \wedge \theta[\epsilon] \wedge \sigma^\star) = \Pr_\infty^{\vec{\tau}}(D \mid KB)$. The desired expression now follows. ∎

We now simplify the expression $\Pr_\infty^{\vec{\tau}}(\varphi \mid KB \wedge \theta[\epsilon] \wedge \sigma^\star \wedge D)$.

**Proposition D.4:** *For $\varphi$, $KB$, $\sigma^\bullet$, $D$, and $\theta[\epsilon]$ as above, if $\mathrm{Pr}^{\vec{\tau}}_\infty(D|KB) > 0$, then*

$$\mathrm{Pr}^{\vec{\tau}}_\infty(\varphi|KB \wedge \theta[\epsilon] \wedge \sigma^\bullet \wedge D) = \mathrm{Pr}_\infty(\varphi|\sigma^\bullet \wedge D),$$

*and its value is either 0 or 1. Note that since the latter probability only refers to first-order formulas, it is independent of the tolerance values.*

**Proof:** That the right-hand side is either 0 or 1 is proved in (Grove et al., 1993b), where it is shown that the asymptotic probability of any pure first-order sentence when conditioned on knowledge of the form $\sigma^\bullet \wedge D$ (which is, essentially, what was called a *model description* in (Grove et al., 1993b)) is either 0 or 1. Very similar techniques can be used to show that the left-hand side is also either 0 or 1, and that the conjuncts $KB \wedge \theta[\epsilon]$ do not affect this limit (so that the left-hand side and the right-hand side are in fact equal). We briefly sketch the relevant details here, referring the reader to (Grove et al., 1993b) for full details.

The idea (which actually goes back to Fagin (1976)) is to associate with a model description such as $\sigma^\bullet \wedge D$ a theory $T$ which essentially consists of *extension axioms*. Intuitively, an extension axiom says that any finite substructure of the model defined by a complete description $D'$ can be extended in all possible ways definable by another description $D''$. We say that a description $D''$ *extends* a description $D'$ if all conjuncts of $D'$ are also conjuncts in $D''$. An extension axiom has the form $\forall x_1, \ldots, x_j \, (D' \Rightarrow \exists x_{j+1} \, D'')$, where $D'$ is a complete description over $\mathcal{X} = \{x_1, \ldots, x_j\}$ and $D''$ is a complete description over $\mathcal{X} \cup \{x_{j+1}\}$, such that $D''$ extends $D'$, both $D'$ and $D''$ extend $D$, and both are consistent with $\sigma^\bullet$. It is then shown that (a) $T$ is complete (so that for each formula $\xi$, either $T \models \xi$ or $T \models \neg\xi$) and (b) if $\xi \in T$ then $\mathrm{Pr}_\infty(\xi|\sigma^\bullet \wedge D) = 1$. From (b) it easily follows that if $T \models \xi$, then $\mathrm{Pr}_\infty(\xi|\sigma^\bullet \wedge D)$ is also 1. Using (a), the desired 0-1 law follows. The only difference from the proof in (Grove et al., 1993b) is that we need to show that (b) holds even when we condition on $KB \wedge \theta[\epsilon] \wedge \sigma^\bullet \wedge D$, instead of just on $\sigma^\bullet \wedge D$.

So suppose $\xi$ is the extension axiom $\forall x_1, \ldots, x_j \, (D' \Rightarrow \exists x_{j+1} \, D'')$. We must show that $\mathrm{Pr}_\infty(\xi|KB \wedge \theta[\epsilon] \wedge \sigma^\bullet \wedge D) = 1$. We first want to show that the right-hand side of the conditional is consistent. As observed in the previous proof, it follows from Theorem 3.16 that $\mathrm{Pr}_\infty(D|KB) = \mathrm{Pr}^{\vec{\tau}}_\infty(\varphi|KB \wedge \theta[\epsilon] \wedge \sigma^\bullet)$. Since we are assuming that $\mathrm{Pr}_\infty(D|KB) > 0$, it follows that $\mathrm{Pr}_\infty(KB \wedge \theta[\epsilon] \wedge \sigma^\bullet \wedge D) > 0$, and hence $KB \wedge \theta[\epsilon] \wedge \sigma^\bullet \wedge D$ must be consistent.

Fix a domain size $N$ and consider the set of worlds satisfying $KB \wedge \theta[\epsilon] \wedge \sigma^\bullet \wedge D$. Now consider some particular $j$ domain elements, say $d_1, \ldots, d_j$, that satisfy $D'$. Observe that, since $D'$ extends $D$, the denotations of the constants are all among $d_1, \ldots, d_j$. For a given $d \notin \{d_1, \ldots, d_j\}$, let $B(d)$ denote the event that $d_1, \ldots, d_j, d$ satisfy $D''$, given that $d_1, \ldots, d_j$ satisfy $D'$. What is the probability of $B(d)$ given $KB \wedge \theta[\epsilon] \wedge \sigma^\bullet \wedge D$? First, note that since $d$ does not denote any constant, it cannot be mentioned in any way in the knowledge base. Thus, this probability is the same for all $d$. The description $D''$ determines two types of properties for $x_{j+1}$. The unary properties of $x_{j+1}$ itself—i.e., the atom $A_i$ to which $x_{j+1}$ must belong—and the relations between $x_{j+1}$ and the remaining variables $x_1, \ldots, x_j$ using the non-unary predicate symbols. Since $D''$ is consistent with $\sigma^\bullet$, the description $\sigma^\bullet$ must contain a conjunct $\exists x \, A_i(x)$ if $D''$ implies $A_i(x_{j+1})$. By definition, $\theta[\epsilon]$ must therefore contain the conjunct $||A_i(x)||_x > \epsilon$. Hence, the probability of picking $d$ in $A_i$ is at least $\epsilon$. For any sufficiently large $N$, the probability of picking $d$ in $A_i$ which is different from $d_1, \ldots, d_j$ (as required by the definition of the extension axiom) is at least $\epsilon/2 > 0$. The probability that $d_1, \ldots, d_j, d$ also satisfy the remaining conjuncts of $D''$, given that $d$ is in atom $A_i$ and $d_1, \ldots, d_j$ satisfy $D'$, is very small but bounded away from 0. (For this to hold, we need the assumption that the non-unary predicates are not mentioned in the $KB$.) This is the case because the total number of possible ways to choose the properties of $d$ (as they relate to $d_1, \ldots, d_j$) is independent of $N$. We can therefore conclude that the probability of $B(d)$ (for sufficiently large $N$), given that $d_1, \ldots, d_j$ satisfy $D'$, is bounded away from 0 by some $\lambda$ independent of $N$. Since the properties of an element $d$ and its relation to $d_1, \ldots, d_j$ can be chosen independently of the properties of a different element $d'$, the different events $B(d), B(d'), \ldots$ are all independent. Therefore, the probability that there is no domain element at all that, together with $d_1, \ldots, d_j$, satisfies $D''$ is at most $(1 - \lambda)^{N-j}$. This bounds the probability of the extension axiom being false, relative to fixed $d_1, \ldots, d_j$. There are $\binom{N}{j}$ ways of these choosing $j$ elements, so the probability of the axiom being false anywhere in a model is at most $\binom{N}{j}(1-\lambda)^{N-j}$. This tends to 0 as $N$ goes to infinity. Therefore, the extension axiom $\forall x_1, \ldots, x_j (D' \Rightarrow \exists x_{j+1} D'')$ has asymptotic probability 1 given $KB \wedge \theta[\epsilon] \wedge \sigma^\star \wedge D$, as desired. ∎

Finally, we are in a position to prove Theorem 4.28.

**Theorem 4.28:** Let $\varphi$ be a formula in $\mathcal{L}^{\approx}$ and let $KB = KB' \wedge \psi$ be an essentially positive knowledge base in $\mathcal{L}_1^{\approx}$ which is separable with respect to $\varphi$. Let $\mathcal{Z}$ be the set of constants appearing in $\varphi$ or in $\psi$ (so that $KB'$ contains none of the constants in $\mathcal{Z}$) and let $\chi^{\neq}$ be the formula $\bigwedge_{c,c' \in \mathcal{Z}} c \neq c'$. Assume that there exists a size description $\sigma^\star$ such that, for all $\vec{\tau} > 0$, $KB$ and $\vec{\tau}$ are stable for $\sigma^\star$, and that the space $S^{\vec{0}}[KB]$ has a unique maximum-entropy point $\vec{v}$. Then

$$\Pr_\infty(\varphi | KB) = \frac{\sum_{D \in \mathcal{A}(\psi \wedge \chi^{\neq})} \Pr_\infty(\varphi | \sigma^\star \wedge D) F_{[D]}(\vec{v})}{\sum_{D \in \mathcal{A}(\psi \wedge \chi^{\neq})} F_{[D]}(\vec{v})}$$

if the denominator is positive.

**Proof:** Assume without loss of generality that $\psi$ mentions all the constant symbols in $\varphi$, so that $\mathcal{A}(\psi \wedge \chi^{\neq}) \subseteq \mathcal{A}(\psi)$. By Proposition D.3,

$$\Pr_\infty^{\vec{\tau}}(\varphi | KB) = \sum_{D \in \mathcal{A}(\psi)} \Pr_\infty^{\vec{\tau}}(\varphi | KB \wedge \theta[\epsilon] \wedge \sigma^\star \wedge D) \cdot \Pr_\infty^{\vec{\tau}}(D | KB).$$

Note that we cannot easily take limits of $\Pr_\infty^{\vec{\tau}}(\varphi | KB \wedge \theta[\epsilon] \wedge \sigma^\star \wedge D)$ as $\vec{\tau}$ goes to $\vec{0}$, because this expression depends on $\theta[\epsilon]$ and the value of $\epsilon$ used depends on the choice of $\vec{\tau}$. However, applying Proposition D.4, we get

$$\Pr_\infty^{\vec{\tau}}(\varphi | KB) = \sum_{D \in \mathcal{A}(\psi)} \Pr_\infty(\varphi | \sigma^\star \wedge D) \cdot \Pr_\infty^{\vec{\tau}}(D | KB).$$

We can now take the limit as $\vec{\tau}$ goes to $\vec{0}$. To do this, we use Proposition D.2. The hypotheses of the theorem imply that $\Pr_\infty(\psi | KB') > 0$ (for otherwise, the denominator $\sum_{D \in \mathcal{A}(\psi \wedge \chi^{\neq})} F_{[D]}(\vec{v})$ would be zero). Part (a) of the proposition tells us we can ignore those complete descriptions that are inconsistent with $\chi^{\neq}$. We can now apply part (b) to get the desired result. ∎

## Acknowledgements

We are very grateful to Professor Gregory Brumfiel, of the Department of Mathematics at Stanford University, for his invaluable help with the proof of Proposition B.5. We would like to thank Fahiem Bacchus, with whom we started working on this general area of research, and Moshe Vardi for useful comments on a previous draft of this paper. A preliminary version of this paper appeared in *Proc. 7th IEEE Symposium on Logic in Computer Science.* Some of this research was performed while Adam Grove and Daphne Koller were at Stanford University and at the IBM Almaden Research Center. This research was sponsored in part by the Air Force Office of Scientific Research (AFSC), under Contract F49620-91-C-0080, by an IBM Graduate Fellowship, and by a University of California President's Postdoctoral Fellowship.